\documentclass[review]{elsarticle}

\usepackage{lineno,hyperref}
\modulolinenumbers[5]
\journal{Nonlinear Dynamics}

\usepackage{amsmath}
\usepackage{algorithm}
\usepackage{algpseudocode}
\usepackage{dsfont}
\usepackage{color}
\usepackage{amssymb}
\usepackage[english]{babel}
\usepackage{graphicx,color}
\usepackage{extarrows}
\usepackage{mathrsfs}
\usepackage{booktabs}
\usepackage{multirow}
\usepackage{upgreek}
\usepackage{threeparttable}
\usepackage{textcomp}
\usepackage[justification=raggedright]{caption}
\usepackage[misc]{ifsym}
\usepackage{array}
\usepackage{graphicx}
\usepackage{subfloat}
\usepackage{float}
\usepackage{subfigure}

\makeatletter
\@addtoreset{equation}{section}
\makeatother

\graphicspath{ {./figures/} }

\begin{document}
	\begin{frontmatter}
		
		\title{Machine Learning for Complex Systems with Abnormal Pattern by Exception Maximization Outlier Detection Method}

		\author[inst1]{Zhikun Zhang}
		\affiliation[inst1]{organization={School of Mathematics and Statistics},
			addressline={Huazhong University of Science and Technology}, 
			city={Wuhan},
			postcode={430074},
			country={China}}
		
		\author[inst2]{Yiting Duan}
		\affiliation[inst2]{organization={School of Reliability and Systems Engineering},
			addressline={Beihang University}, 
			city={Beijing},
			postcode={100091},
			country={China}}
		
		\author[inst1]{Xiangjun Wang}
		\ead{xjwang@hust.edu.cn}
		
		\author[inst3]{Mingyuan Zhang\corref{mycorrespondingauthor}}
		\cortext[mycorrespondingauthor]{Corresponding author}
		\ead{zmyxinyang@126.com}
		\affiliation[inst3]{
			addressline={National Key Laboratory of Science and Technology on Vessel Integrated Power System}, 
			city={Wuhan},
			postcode={430033},
			country={China}}

		\begin{abstract}
			This paper proposes a novel fast online methodology for outlier detection called the exception maximization outlier detection method(EMODM), which employs probabilistic models and statistical algorithms to detect abnormal patterns from the outputs of complex systems. The EMODM is based on a two-state Gaussian mixture model and demonstrates strong performance in probability anomaly detection working on real-time raw data rather than using special prior distribution information. We confirm this using the synthetic data from two numerical cases. For the real-world data, we have detected the short circuit pattern of the circuit system using EMODM by the current and voltage output of a three-phase inverter. The EMODM also found an abnormal period due to COVID-19 in the insured unemployment data of 53 regions in the United States from 2000 to 2024. The application of EMODM to these two real-life datasets demonstrated the effectiveness and accuracy of our algorithm.
		\end{abstract}
		
		
	\end{frontmatter}
	
	
	\section{Introduction}\label{First_sec}
	In recent years, the complexity of nonlinear systems has been increasing dramatically, and complex systems often switch back and forth between normal and abnormal states. To maintain stability and improve the system's reliability, the key to improving system reliability is to find an efficient abnormal state detection method that yields reliable failure probability results. Abnormal pattern switching of complex systems can be caused by various factors such as parameter drift, which typically occurs during system operation\cite{xu2019sensor,don2019dynamic}, and unforeseen external forces involved such as the short condition in circuit systems and the iterative rise and fall of economic markets. Nowadays, machine learning for abnormal patterns of nonlinear complex systems has emerged as an engaging and challenging field in related fields.
	
	Traditional online abnormal detection methods are inherently restricted by mathematical models and observation distributions\cite{ammiche2018combined}. Filter-based models such as the Kalman filter method, based on dynamic variance minimization, are designed for fault diagnosis and exhibit strong robustness\cite{brumback1987chi,xiong2018design}. However, these methods require the model to be linear and the availability of prior statistical information on noise. The improved extended Kalman filter method can linearize nonlinear models by Taylor expansion but requires that the data noise be Gaussian distributed\cite{an2005hydraulic}. The particle filter algorithm provides finer state estimation results for nonlinear systems with non-Gaussian noise\cite{yin2015intelligent} but may require the user to have some prior knowledge about the overall system state which may not be able to achieve this in practice. In most scenarios, researchers' experience alone may not provide insight into faults from a global perspective, and accurate results are often not obtained for a model output\cite{stuart2015data}.
	On the other hand, classical machine learning methods have also been successfully used in abnormal detection and diagnosis\cite{kanovic2011generalized}, and data-driven models have made significant progress in the field of abnormal detection, as seen in the example of aircraft detection\cite{cartocci2022aircraft}. Moreover, numerous control methods, such as $H_{\infty}$ control\cite{fan2021h}, sliding mode control\cite{li2019sensor}, and back-stepping control\cite{zhao2019finite}, have been widely employed in abnormal detection and diagnosis for nonlinear systems. It is important to note that an effective abnormal detection algorithm should be highly accurate and robust, given that observation data in the real world are frequently affected by Gaussian observation noise, which inevitably impacts the outcomes of circuit fault diagnosis\cite{wang2006robust}.
	
	The abnormal detection and diagnosis problems related to time-dependent differential equations can be addressed using modern statistical methods, such as change-point analysis or outlier detection, rooted in industrial quality monitoring\cite{kano2008data,su2013line}. The development and refinement of statistical analysis methods have led to significant progress in both theoretical and practical research applications, finding use in various fields including biomedicine, pattern recognition, and econometrics\cite{fruhwirth2006finite}. The introduction of statistical analysis algorithms is expected to bring new vitality to the traditional field of pattern recognition and abnormal detection for complex nonlinear systems. After detecting and diagnosing outliers in system output, evaluating complex systems for small abnormal probability failures is an important topic in modern control theories, as the reliability of the system could be affected even by minor failures. Various statistical sampling methods for the circuit system output have been proposed\cite{khalid2014novel} but obtaining actual probability values requires performing an extremely large number of simulation iterations which leads to an infeasible huge computational burden\cite{belloni2009computational}. Therefore, a more accurate and stable sampling method is needed to estimate the failure probability of complex systems.
	
	In this study, a statistical two-classification problem for complex system outputs was proposed to transform abnormal detection and diagnosis into a more efficient and robust statistical algorithm. This was achieved by introducing the Gaussian mixture model(GMM) and then developing an exception maximization outlier detection method(EMODM) based on a two-state mixture model for time-series data of the complex system output. Accurate abnormal detection was achieved for complex systems with observation noise using a finite mixture model. This method enabled effective abnormal detection through probability estimation in a short observation time segment with automatic online abnormal detection.
	
	The paper is organized as follows. In section \ref{Second_sec}, the EMODM used for abnormal detection and diagnosis by the complex system output is presented. In section \ref{Third_sec}, we present the numerical simulation of a Sallen-key low-pass filter case and an information processing and storing for the spin-magnet circuits case. In section \ref{forth_sec}, we show empirical verification results with data on real circuit short-circuit failures and data on the United States(U.S.) insured unemployment including the COVID-19 epidemic time. Section \ref{Comparison} discusses the comparison of EMODM with other classical methods. Section \ref{fifth_sec} is the conclusion of the paper.

	\section{Exception Maximization Outlier Detection Method}\label{Second_sec}
	In this section, we will introduce how the GMM in statistics and the associated Exception Maximization(EM) algorithm can be extended into the EMODM which will be used in statistical inference and abnormal pattern detection for the output of complex systems.
	
	\subsection{Complex System with Abnormal Pattern}
	Firstly, we consider a $n$-dimensional ordinary differential equations on $t \in [t_0, T]$ as
	\begin{equation}\label{eeq:2.1}
		\begin{cases}
			\mathrm{d}x_{1}(t)=f_{1}(t,x_{1}(t),x_{2}(t),\cdots,x_{n}(t))\mathrm{d}t, \\
			\mathrm{d}x_{2}(t)=f_{2}(t,x_{1}(t),x_{2}(t),\cdots,x_{n}(t))\mathrm{d}t, \\
			\cdots \cdots  \\
			\mathrm{d}x_{n}(t)=f_{n}(t,x_{1}(t),x_{2}(t),\cdots,x_{n}(t))\mathrm{d}t,
		\end{cases}
	\end{equation}
	where the functions 
	\begin{equation}\label{eeq:2.2}
		f_{i}(t,x_{1}(t),x_{2}(t),\cdots,x_{n}(t)),\:i=1,2,\cdots,n
	\end{equation}
	are defined in an open region $G$ in $\mathbb{R}^{n+1}$ space. By setting vector function $\mathbf{x}(t)=(x_{1}(t),x_{2}(t),\cdots,x_{n}(t))$ and
	\begin{equation}
		\mathbf{f}(t,\mathbf{x}(t))=(f_{1}(t,\mathbf{x}(t)),f_{2}(t,\mathbf{x}(t)),\cdots,f_{n}(t,\mathbf{x}(t))),
	\end{equation}
	we could rewrite \eqref{eeq:2.1} as a vector form 
	\begin{equation}\label{eeq:2.5}
		\mathrm{d}\mathbf{x}(t)=\mathbf{f}(t,\mathbf{x}(t))\mathrm{d}t,
	\end{equation}
	where $\mathbf{f}:\mathbb{R}^{n+1} \rightarrow \mathbb{R}^n$ with initial condition $(t_0, \mathbf{x}_0) \in G \subset \mathbb{R}^{n+1}$.
	
	Now assume that \eqref{eeq:2.5} contains a finite number of unknown parameters, and rewrite it as the following parametric form
	\begin{equation}\label{eeq:2.9}
		\mathrm{d}\mathbf{x}(t)=\mathbf{f}(t,\mathbf{x}(t);\boldsymbol{\Lambda})\mathrm{d}t,
	\end{equation}
	where $\boldsymbol{\lambda}=\{\lambda_1,\lambda_2,\cdots,\lambda_d\}$ is a $d$-dimensional vector of system parameters, in other words, a set of the equation coefficients.
	Let stochastic process $S(t,\omega),\: t\in [t_0, T], \: \omega \in \Omega$ be an independent and identically distributed process, where $\Omega$ is sample space in probability theory and could simply be written as $S_t$. This process is taking value in a discrete finite index set $\mathbf{I}=\{1, 2\}$, where $S_t=1$ means there is no anomaly in the system and $S_t=2$ means that this system is in an abnormal pattern. It corresponds to the equation coefficients changes in normal and abnormal patterns of the complex system in parameter switching space $\boldsymbol{\Lambda}=\{\boldsymbol{\lambda}_1, \boldsymbol{\lambda}_2\}$ including two entirely different sets of equation coefficients.
	
	Moreover, we claim that the continuous functions
	\begin{equation}\label{eeq:2.4}
		x_{i}(t)=\varphi_i(t),\quad  t \in [t_0, T],\quad i=1,2,\cdots,n
	\end{equation}
	defined in $G$ and satisfying \eqref{eeq:2.1} is a solution of this system. Then in this study, we will talk about how to finish the abnormal detection and diagnosis from a solution $\mathbf{x}(t)$ of the complex system with an abnormal pattern as the following form
	\begin{equation}\label{eeq:2.11}
		\mathrm{d}\mathbf{x}(t)=\mathbf{f}^{\ast}(t,\mathbf{x}(t),S_t;\boldsymbol{\Lambda})\mathrm{d}t,
	\end{equation}
	where $\mathbf{f}^{\ast}:\mathbb{R}^{n+1} \times \mathbf{I} \rightarrow \mathbb{R}^n$. The initial condition satisfies $(t_0, \mathbf{x}_0) \in G \subset \mathbb{R}^{n+1}$ and the initial value of the system pattern is set to $S_0=1$ so that the system is normal at the beginning of observation. In the next statistical inference work, by formula
	\begin{equation}
		y_i=\frac{x_{i}-x_{i-1}}{x_{i-1}}, \quad 1 \leq i \leq N,
	\end{equation}
	we compute the relative change rate $\mathbf{y}=\{y_1,y_2,\cdots,y_N\}$ for the discrete solution $\mathbf{x}=\{x_0,x_1,x_2,\cdots,x_N\}$ of the equation \ref{eeq:2.11} and treat it as a sample set for the EMODM.
	
	\subsection{Classification for Two-state Gaussian Mixture Model}
	To begin with, a two-component GMM is obtained by the random variable $Y$
	\begin{equation}\label{eq:2.1}
		Y \sim (1-\eta)\mathcal{N}(\mu_1, \sigma_1^2)+\eta\mathcal{N}(\mu_2, \sigma_2^2),
	\end{equation}
	where $\eta_1=1-\eta$ is the proportion of time for correct working and $\eta_2=\eta$ is the proportion of time for working in an abnormal pattern. The density function of the Gaussian distribution is
	\begin{equation}\label{2.2}
		f_{\mathcal{N}}(x;\, \mu, \sigma^2)=\frac{1}{\sigma \sqrt{2 \pi}}\mathrm{exp}{\left\{-\frac{(x-\mu)^2}{2 \sigma^2}\right\}}.
	\end{equation}
	Considering the existence of the random error in observation, the full-time observations of the whole system model which has the correct and abnormal patterns are equivalent to a mixture of two Gaussian submodels.
	
	In our model, we assumed that $N$ sample $\mathbf{y}=\{y_1,y_2,\cdots,y_N\}$ in finite time $[0, T]$ comes from a finite GMM with two component. For each observation $y_i$ can be attributed a latent label variable $\mathbf{S}=\{S_1,S_2,\cdots,S_N\}$ and it takes value in finite state space $I=\{1,2\}$. So that $S_t$ can only hold two values, specifying that $k=1$ means the system is correctly working and the opposite $k=2$ indicates that the system is abnormal. All observations $y_i$ with label variable $S_i= k$ come from the same Gaussian distribution $\mathcal{N}(\mu_k, \sigma_k^2)$. The complete-data likelihood function defined earlier in \eqref{2.2} for GMM is
	\begin{equation}\label{eq:2.5}
		p(\mathbf{y},\mathbf{S}|\vartheta)=\prod_{k=1}^K\left(\prod_{i:S_{i}=k} f_{\mathcal{N}}(y_i;\, \mu_k, \sigma_k^2) \right)\left(\prod_{k=1}^K \eta_k^{N_k(\mathbf{S})} \right),
	\end{equation}
	where $N_k(\mathbf{S})=\#\{S_i=k,\;i=1,2,\cdots, N\}$ denotes the count assigned to class $k$ of data $\mathbf{y}$. Therefore $\vartheta=\{\mu_1,\mu_2,\sigma_1,\sigma_2,\eta\}$ is the two-state GMM parameters for which statistical inference is required. When the complex system is in an abnormal pattern, it can be considered that $K = 2$ in \eqref{eq:2.5}.
	
	Both the observed variable $\mathbf{y}$ and the hidden variable $\mathbf{S}$ exist within the GMM. It is more difficult to estimate model parameters directly using the maximum likelihood method or Bayesian when there exists a hidden variable. The EM algorithm can then be used to efficiently find the maximum likelihood solution of the parameters of a finite mixture model with latent variables. To implement the EM algorithm of GMM, the complete data log-likelihood function $p(\mathbf{y},\mathbf{S}|\vartheta)$ defined in \eqref{eq:2.5} is rewritten as
	\begin{equation}\label{eq:2.6}
		\mathrm{log}\, p(\mathbf{y},\mathbf{S}|\vartheta)=\sum_{i=1}^{N}\sum_{k=1}^{K}D_{ik}\mathrm{log}\Big(\eta_k f_{\mathcal{N}}(y_i;\,\mu_k,\sigma_k)\Big),
	\end{equation}
	where the latent variable $D_{ik}$ is a $0/1$ encoding of the assignment $D_{ik}=1$ if and only if $S_i=k$.
	
	The EM algorithm needs to start by determining the initial values of the model parameters $\vartheta^{\{0\}}=\{\mu_k^{\{0\}},\sigma_k^{\{0\}},\eta^{\{0\}},\,k=1,2\}$. This algorithm for the two-component GMM model is iterated between the expectation step and the maximization step. The expectation step uses the parameter estimations of the model from the previous step to calculate the conditional expectation of the log-likelihood function for the complete data, and the estimate of the latent variable $D_{ik}$ in $m$th iteration is
	\begin{equation}\label{eq:2.7}
		D^{\{m\}}_{ik}=\frac{\hat{\eta}_{k}^{\{m-1\}}f_{\mathcal{N}}(y_i;\,\hat{\mu}_k^{\{m-1\}},\hat{\sigma}_k^{\{m-1\}})}{\sum_{j=1}^{K} \hat{\eta}_{j}^{\{m-1\}}f_{\mathcal{N}}(y_i;\,\hat{\mu}_j^{\{m-1\}},\hat{\sigma}_j^{\{m-1\}})}.
	\end{equation}        
	The maximization step determines the parameters $\hat{\vartheta}_j^{\{m-1\}}$ for maximizing the log-likelihood function of the complete data obtained in the expectation step
	\begin{equation}\label{eq:2.8}
		\arg\mathop{\max}\limits_{\hat{\vartheta}^{\{m\}}}\sum_{i=1}^{N}\sum_{k=1}^{K} D^{\{m\}}_{ik}\mathrm{log}\Big(\hat{\eta}^{\{m\}}_k f_{\mathcal{N}}(y_i;\,\hat{\mu}^{\{m\}}_k,\hat{\sigma}^{\{m\}}_k)\Big).
	\end{equation}
	The updates of GMM parameters in $m$th iteration are
	\begin{equation}\label{eq:2.9}
		\begin{split}
			& \hat{\mu}^{\{m\}}_k=\frac{\sum_{j=1}^N D^{\{m\}}_{jk} y_j} {\sum_{j=1}^N D^{\{m\}}_{jk}}, \\
			& \hat{\sigma}^{\{m\}}_k=\frac {\sum_{j=1}^N D^{\{m\}}_{jk} (y_j-\hat{\mu}^{\{m\}}_k)^2} {\sum_{j=1}^N D^{\{m\}}_{jk}}, \\
			& \hat{\eta}^{\{m\}}=\frac{\sum_{j=1}^{N} D^{\{m\}}_{js}} {N},\;s=2.  
		\end{split}
	\end{equation}
	Then in continuous iterations, until the algorithm converges, the final two-state GMM parameter estimation is generated as
	\begin{equation}\label{2.15}
		\hat{\vartheta}=\{\hat{\mu}_1,\hat{\mu}_2,\hat{\sigma}_1,\hat{\sigma}_2,\hat{\eta}\}.
	\end{equation}
	
	\subsection{Online Outlier Detection and Abnormal Probability Estimation}
	By the assumption that outliers always have a small proportion, we regard the group with a small proportion $\eta$ of the two-state GMM as the output from a complex system in an abnormal pattern. By Bayes theorem
	\begin{equation}\label{eq:3.11}
		p(S_i=2|y_i,\vartheta)=\frac{\eta p(y_i|\mu_2,\sigma_2)}{(1-\eta) p(y_i|\mu_1,\sigma_1)+\eta p(y_i|\mu_2,\sigma_2)}.
	\end{equation}
	Then we can calculate a corresponding sequence of abnormal probabilities by \eqref{2.2}. By setting certain thresholds $\alpha_f$, generally as a constant close to 1.
	
	Then in this study, we define the abnormal pattern set as
	\begin{equation}
		\mathbf{F}^{\mathrm{error}}=\{e_i;\,f(e_i)\geq \alpha_f,\,1\leq i \leq N \}.
	\end{equation}
	Its elements are time points of abnormal patterns in the complex system which are captured by EMODM. It can be updated at any time as the time series data continue to increase without a significant computational burden. Thereby realizing our online detection of abnormal patterns in complex systems.
	
	At last, by statistical results due to our EMODM, the global abnormal probability of the circuit system can be easily given as
	\begin{equation}\label{eq:2.10}
		P_{f}=\hat{\eta}, 
	\end{equation}
	which is the incidental result of GMM classification in \eqref{2.15}. In this way, we can easily obtain the evaluation failure probability without using any sampling methods thus achieving real-time early warning and abnormal probability estimation at high speed.
	
	\subsection{Convergence of Algorithm}
	Let $P(\mathbf{Y}|\vartheta^{(i)})$ be the likelihood function of observed data $\mathbf{Y}$, $\vartheta$ be a variable and $\vartheta^{(i)}$ be the estimated parameters of GMM as \eqref{2.15}. Since
	\begin{equation}
		P(\mathbf{Y}|\vartheta)=\frac{P(\mathbf{Y},\mathbf{S}|\vartheta)}{P(\mathbf{S}|\mathbf{Y},\vartheta)}.
	\end{equation}
	By logarithm derivation, we have
	\begin{equation}
		\log P(\mathbf{Y}|\vartheta)=\log{P(\mathbf{Y},\mathbf{S}|\vartheta)}-\log{P(\mathbf{S}|\mathbf{Y},\vartheta)}.
	\end{equation}
	Also, we have
	\begin{equation}
		Q(\vartheta,\vartheta^{i}) =\sum_{i:S_i=k}\log{P(\mathbf{S}|\mathbf{Y},\vartheta^{(i)})}\log{P(\mathbf{Y},\mathbf{S}|\vartheta)}.
	\end{equation}
	Set
	\begin{equation}\label{2}
		H(\vartheta,\vartheta^{i}) =\sum_{i:S_i=k}\log{P(\mathbf{S}|\mathbf{Y},\vartheta^{(i)})}\log{P(\mathbf{S}|\mathbf{Y},\vartheta)}.
	\end{equation}
	Then we can rewrite the log-likelihood formula as 
	\begin{equation}
		\log{P(\mathbf{Y}|\vartheta)}=Q(\vartheta,\vartheta^{(i)})-H(\vartheta,\vartheta^{(i)})\label{siran}.
	\end{equation}
	On substituting $\vartheta$ in \eqref{siran} by $\vartheta^{(i)}$ and $\vartheta^{(i+1)}$, we have
	\begin{equation}\label{4}
		\begin{aligned}
			&\log P(\mathbf{Y}|\vartheta^{(i+1)})-\log P(\mathbf{Y}|\vartheta^{(i)})\\=&[Q(\vartheta^{(i+1)},\vartheta^{(i)})-Q(\vartheta^{(i)},\vartheta^{(i)})]-[H(\vartheta^{(i+1)},\vartheta^{(i)})-H(\vartheta^{(i)},\vartheta^{(i)})].
		\end{aligned}
	\end{equation}
	We can get the conclusion if (\ref{4}) is non-negative on its right. Consequently, our task is to prove non-negativity on its right. From the expectation step \eqref{eq:2.8} of
	\begin{equation}
		\vartheta^{(i+1)}=\arg \max _{\vartheta} \sum_{i:S_i=k}\log{P(\mathbf{S}|\mathbf{Y},\vartheta^{(i)})}\log{P(\mathbf{Y},\mathbf{S}|\vartheta)},
	\end{equation}
	we could deduce that
	\begin{equation}\label{5}
		Q(\vartheta^{(i+1)},\vartheta^{(i)})-Q(\vartheta^{(i)},\vartheta^{(i)}) \geq 0.
	\end{equation}
	As for the second, by (\ref{2}) and Jensen inequality, it follows
	\begin{equation}\label{6}
		\begin{aligned}
			&H(\vartheta^{(i+1)},\vartheta^{(i)})-H(\vartheta^{(i)},\vartheta^{(i)})\\=&
			\sum_{i:S_i=k}\Bigg(\log \frac{P(\mathbf{S}|\mathbf{Y},\vartheta^{(i+1)})}{P(\mathbf{S}|\mathbf{Y},\vartheta^{(i)})}\Bigg)P(\mathbf{S}|\mathbf{Y},\vartheta^{(i)}) \Bigg)\\
			\leq& \log\Bigg(\sum_{i:S_i=k}\Bigg(\frac{P(\mathbf{S}|\mathbf{Y},\vartheta^{(i+1)})}{P(\mathbf{S}|\mathbf{Y},\vartheta^{(i)})}\Bigg)P(\mathbf{S}|\mathbf{Y},\vartheta^{(i)})\Bigg)\\
			=&\log \Bigg( \sum_{i:S_i=k}P(\mathbf{S}|\mathbf{Y},\vartheta^{(i)})\Bigg)=0.
		\end{aligned}
	\end{equation}
	
	At last, we can verify by (\ref{5}) and (\ref{6}) that $P(\mathbf{Y}|\vartheta^{(i)})$ is monotonic increasing
	\begin{equation}
		P(\mathbf{Y}|\vartheta^{(i+1)})\geq P(\mathbf{Y}|\vartheta^{(i)}).
	\end{equation}  
	This illustrates the convergence of EMODM with $P(\mathbf{Y}|\vartheta) \leq 1$. Finally, the EMODM algorithm is summarized in algorithm \ref{algo}.

	\begin{algorithm}
		\caption{\label{algo}Exception Maximization Outlier Detection Method(EMODM)}
		\begin{algorithmic}[1]
			\State \textbf{Input:} Observations of system output $\mathbf{x} = \{x_0, x_1, x_2, \cdots, x_N\}$, Threshold $\alpha_f$
			\State \textbf{Compute:} Relative Change Rate
			\For{$i = 1$ to $N$}
			\State $y_i=(x_{i}-x_{i-1})/(x_{i-1})$
			\EndFor
			\State \textbf{Initialize:} Parameters $\vartheta^{(0)} = \{\mu_k^{(0)}, \sigma_k^{(0)}, \eta^{(0)}\}$ for $k = 1, 2$
			\Repeat
			\State \textbf{E-step:} Compute the responsibilities
			\For{$i = 1$ to $N$}
			\For{$k = 1$ to $2$}
			\State $D_{ik}^{(m)} = \Big(\eta_k^{(m-1)} f_{\mathcal{N}}(y_i; \mu_k^{(m-1)}, \sigma_k^{(m-1)})\Big)/\Big(\sum_{j=1}^{2} \eta_j^{(m-1)} f_{\mathcal{N}}(y_i; \mu_j^{(m-1)}, \sigma_j^{(m-1)})\Big)$
			\EndFor
			\EndFor
			
			\State \textbf{M-step:} Update parameters
			\For{$k = 1$ to $2$}
			\State $\mu_k^{(m)} = \Big(\sum_{i=1}^{N} D_{ik}^{(m)} y_i\Big)/\Big(\sum_{i=1}^{N} D_{ik}^{(m)}\Big)$
			\State $\sigma_k^{(m)} = \sqrt{\Big(\sum_{i=1}^{N} D_{ik}^{(m)} (y_i - \mu_k^{(m)})^2\Big)/\Big(\sum_{i=1}^{N} D_{ik}^{(m)}}\Big)$
			\State $\eta_k^{(m)} = \Big(\sum_{i=1}^{N} D_{ik}^{(m)}\Big)/N$
			\EndFor
			
			\Until{convergence of $\vartheta^{(m)}$}
			
			\State \textbf{Output:} Parameters $\hat{\vartheta} = \{\hat{\mu}_1, \hat{\mu}_2, \hat{\sigma}_1, \hat{\sigma}_2, \hat{\eta}\}$
			
			\State \textbf{Online Outlier Detection:}
			\For{$i = 1$ to $N$}
			\State \textbf{Compute:} Abnormal probability
			\State $p(S_i = 2 | y_i, \vartheta) = \Big(\hat{\eta} f_{\mathcal{N}}(y_i; \hat{\mu}_2, \hat{\sigma}_2)\Big)/\Big((1 - \hat{\eta}) f_{\mathcal{N}}(y_i; \hat{\mu}_1, \hat{\sigma}_1) + \hat{\eta} f_{\mathcal{N}}(y_i; \hat{\mu}_2, \hat{\sigma}_2)\Big)$
			\If{$p(S_i = 2 | y_i, \vartheta) \geq \alpha_f$}
			\State Mark $y_i$ as an outlier $e_i$
			\EndIf
			\EndFor
			
			\State \textbf{Output:} Sequence of identified outliers $\mathbf{F}^{\mathrm{error}}=\{e_i;\,f(e_i)\geq \alpha_f,\,1\leq i \leq N \}$
			
		\end{algorithmic}
	\end{algorithm}

	\section{Numerical Simulation}\label{Third_sec}
	In this section, we evaluate the performance of our EMODM through two numerical simulations. Two numerical cases are taken from two electronic components in microelectronics. The first case is the Sallen-Key low-pass filter system which is a high-order linear system. The second case is the information processing and storing for spin-magnet circuits, which is described by the classical Landau-Liftshitz-Gilbert equation(LLG) equation. Unlike the previous example, the LLG equation is a nonlinear case and we shall see that our novel method works well in both two cases and produces high accuracy in abnormal detection and diagnosis. Additionally, we implement the EMODM to realize online abnormal detection for the above circuit systems.  
	
	\subsection{Sallen-Key Low-pass Filter}\label{3.1}
\begin{figure}[ht!]
		\centering
		\includegraphics[width=0.8\textwidth]{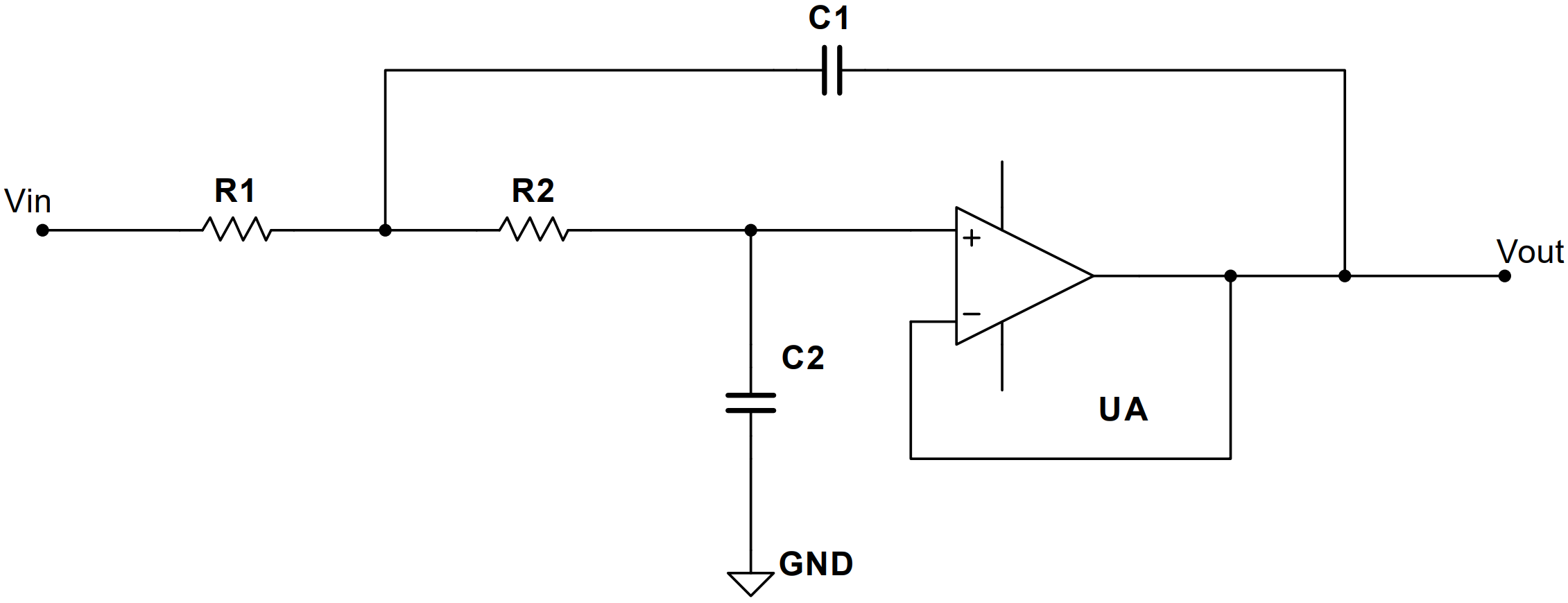}
		\caption{\label{Sallen-Key} Topology of Sallen-Key Low-pass Filter.}
	\end{figure}

	The Sallen-Key low-pass filter is a common electronic device in industrial circuit systems and its structure is shown in figure \ref{Sallen-Key}. This filter was proposed in \cite{sallen1955practical} and consists of a single operational amplifier and a low-pass filter consisting of a resistance and a capacitance. The fault of the Sallen-Key low-pass filter always causes the deterioration of the circuit and even leads to the collapse of the whole system. Currently, a wide range of abnormal detection and diagnosis methods for the Sallen-Key low-pass filter has been proposed\cite{aminian2002analog,xu2017research}. Here, we use this example to demonstrate the accuracy of our EMODM.  
	
	The following second-order linear ordinary differential equation represents the Sallen-Key low-pass filter as follows:
	\begin{equation}\label{eq:3.1}
		R_{1}R_{2}C_{1}C_{2}\ddot{V}_{\text{out}}+( R_{1}+R_{2})C_{2}\dot{V}_{\text{out}}+V_{\text{out}}=V_{\text{in}},
	\end{equation}
	where $V_\text{{in}}$ denotes the excitation voltage which is the sine excitation voltage here, $R_1$, $R_2$ represents the resistance R1, R2, and $C_1$, $C_2$ represents the capacitance C1, C2, respectively. Here, the Radau method is a fully implicit method that is used to solve the differential equation, thus obtaining the simulation results of the Sallen-Key low-pass filter circuit.

	Based on the online abnormal detection EMODM proposed in section \ref{Second_sec}, we explore the ability of our EMODM to implement fault detection. Through the Radau ordinary differential equation solver, we begin our simulation experiment by getting the corresponding outputs according to the voltage of the correct pattern and the voltage of the abnormal pattern separately. We generate 1000 pairs of results of the fault circuit and then obtain the mean value as the output result of the fault circuit at that time period. The total time length of the simulation is 0.02\,$s$, and we divided it into 630 discrete time periods including 7 unequal-length time segments. In the global observation time, 4 segments in a correct pattern containing 600 time periods and 3 segments in the abnormal pattern containing 30 time periods. The voltage value of the system is calculated at each discrete time period and then taken into the numerical solver corresponding to the differential equation that obtains the response to the result of the circuit system at the corresponding time. 
	
\begin{figure}[p]
		\centering
		\subfigure{
			\begin{minipage}{0.48\textwidth}
				\centering
				\includegraphics[scale=0.45]{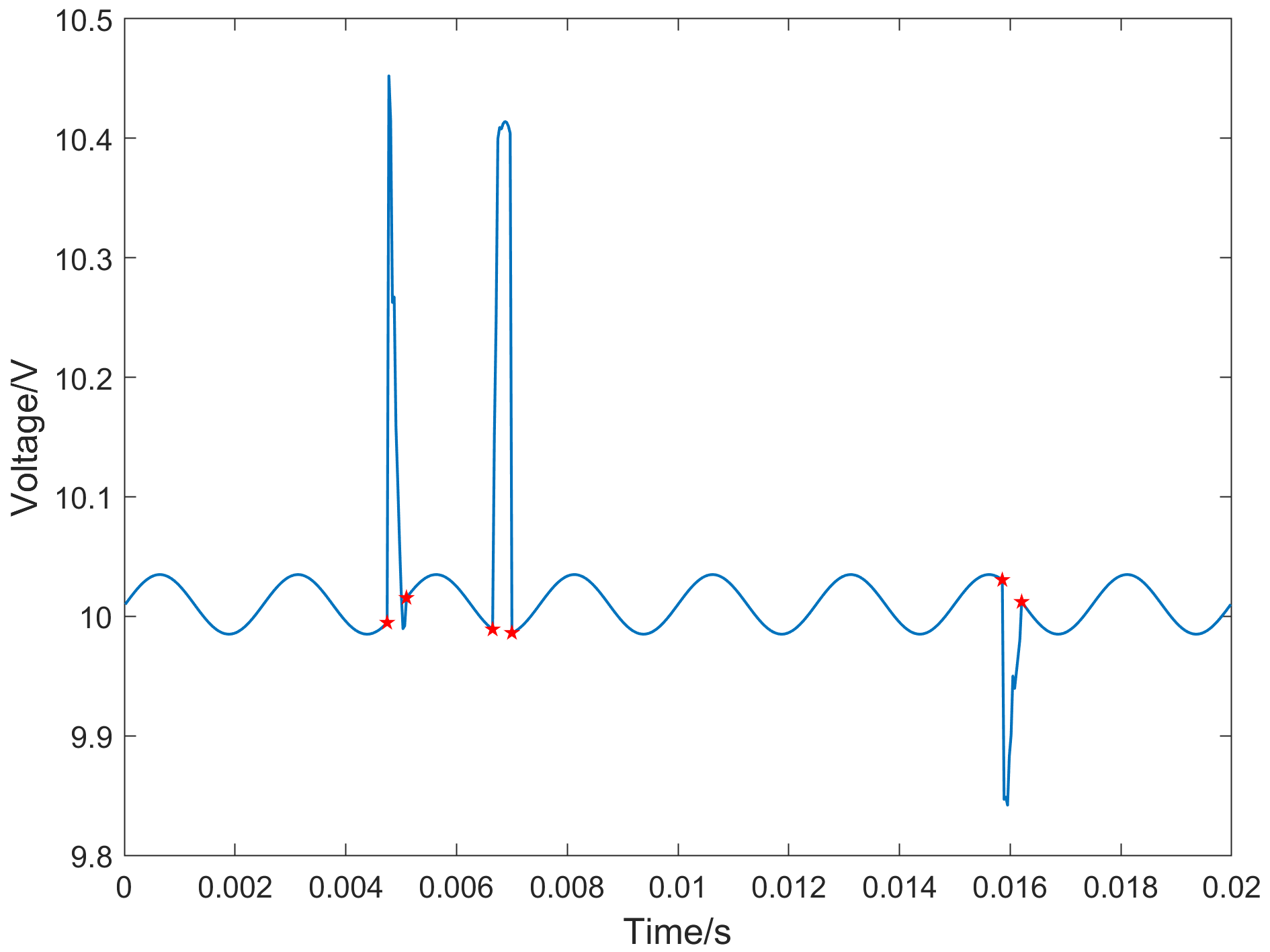}
			\end{minipage}
			\begin{minipage}{0.48\textwidth}
				\centering
				\includegraphics[scale=0.45]{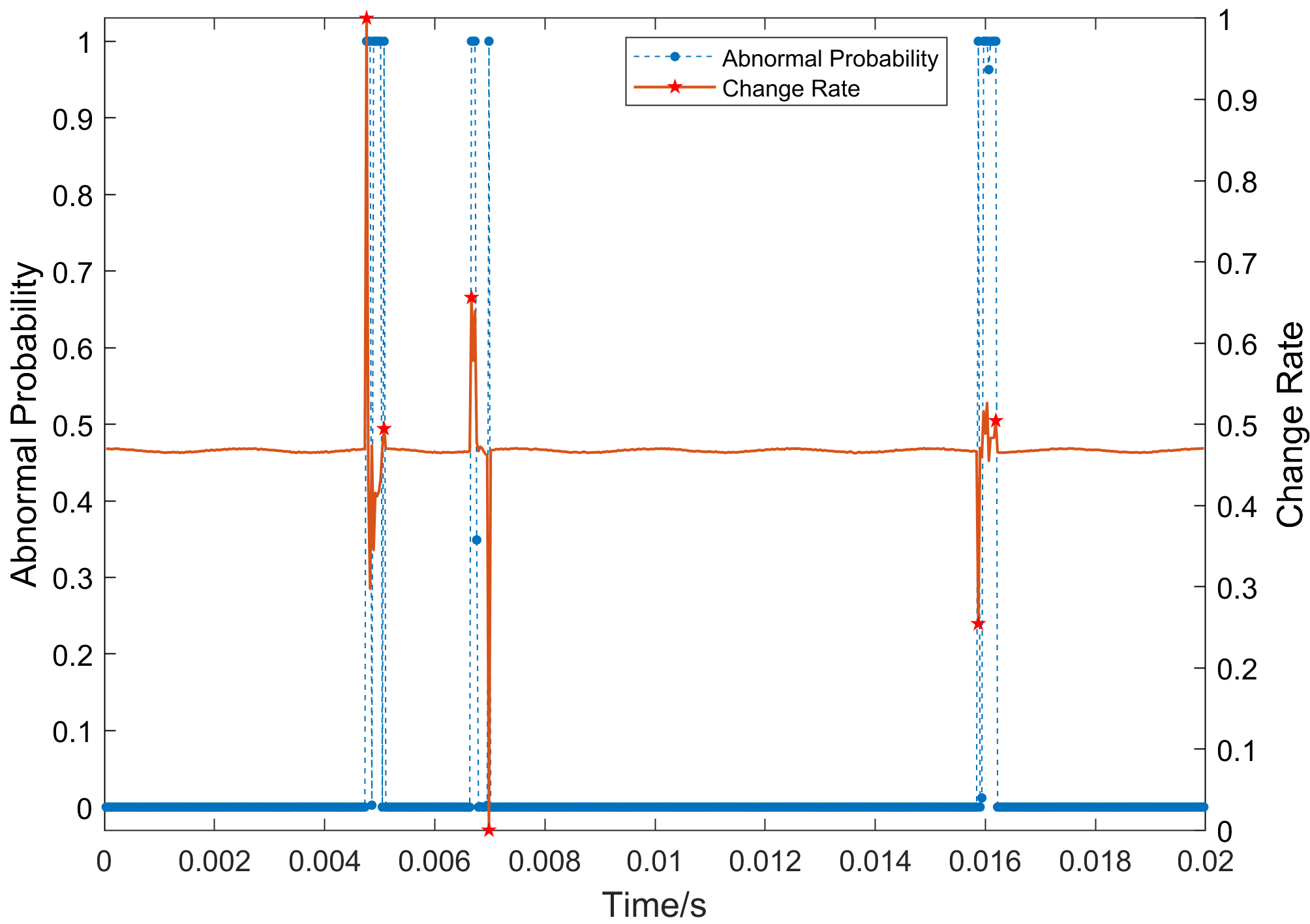}
			\end{minipage}
		}
		\caption{\label{result1}Left: Sallen-Key Low-pass Filter Output with Single-component Voltage Input; Right: EMODM Results by Change Rate of System Output.}
		
		\subfigure{
			\begin{minipage}{0.48\textwidth}
				\centering
				\includegraphics[scale=0.45]{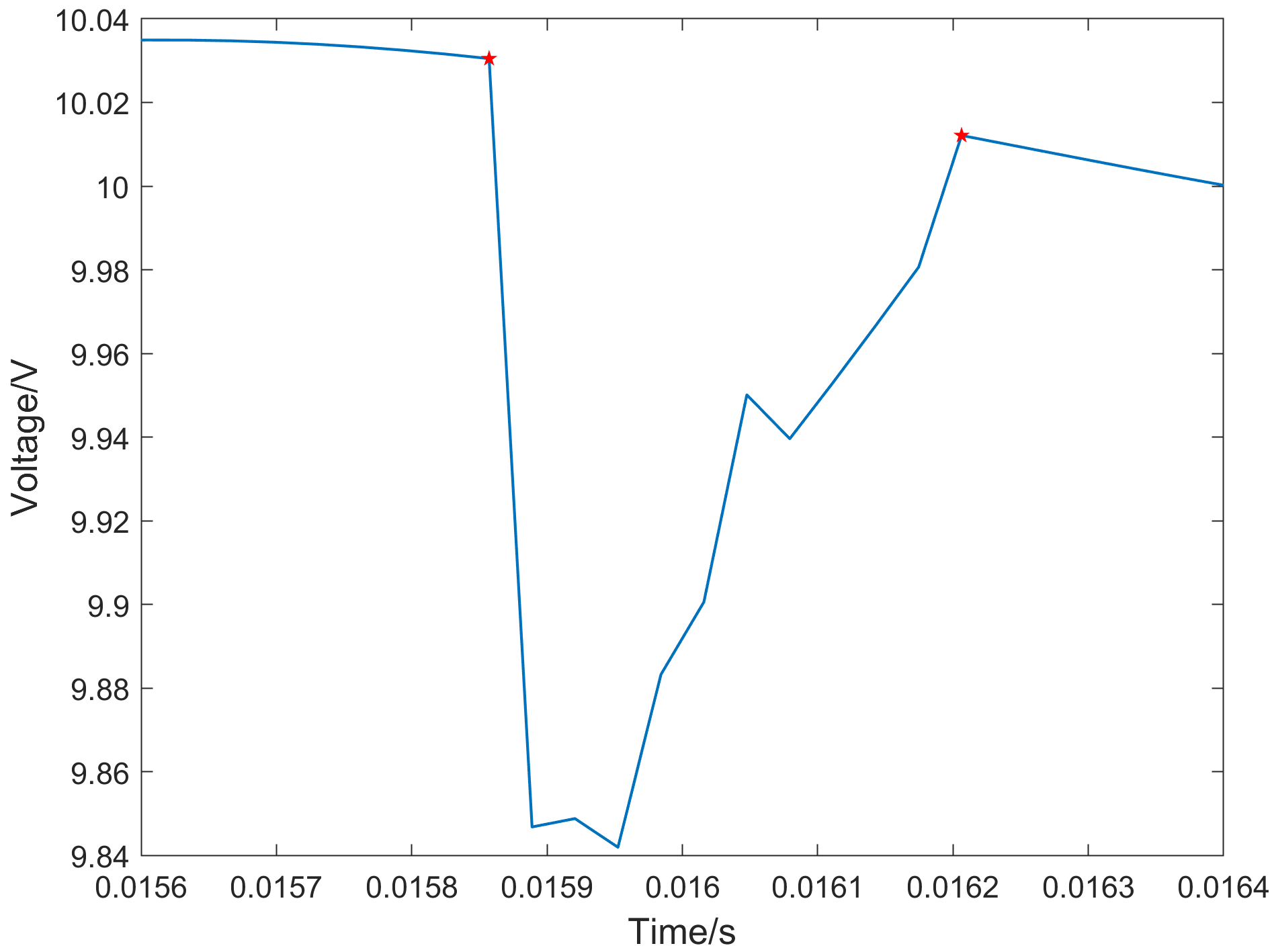}
			\end{minipage}
			\begin{minipage}{0.48\textwidth}
				\centering
				\includegraphics[scale=0.45]{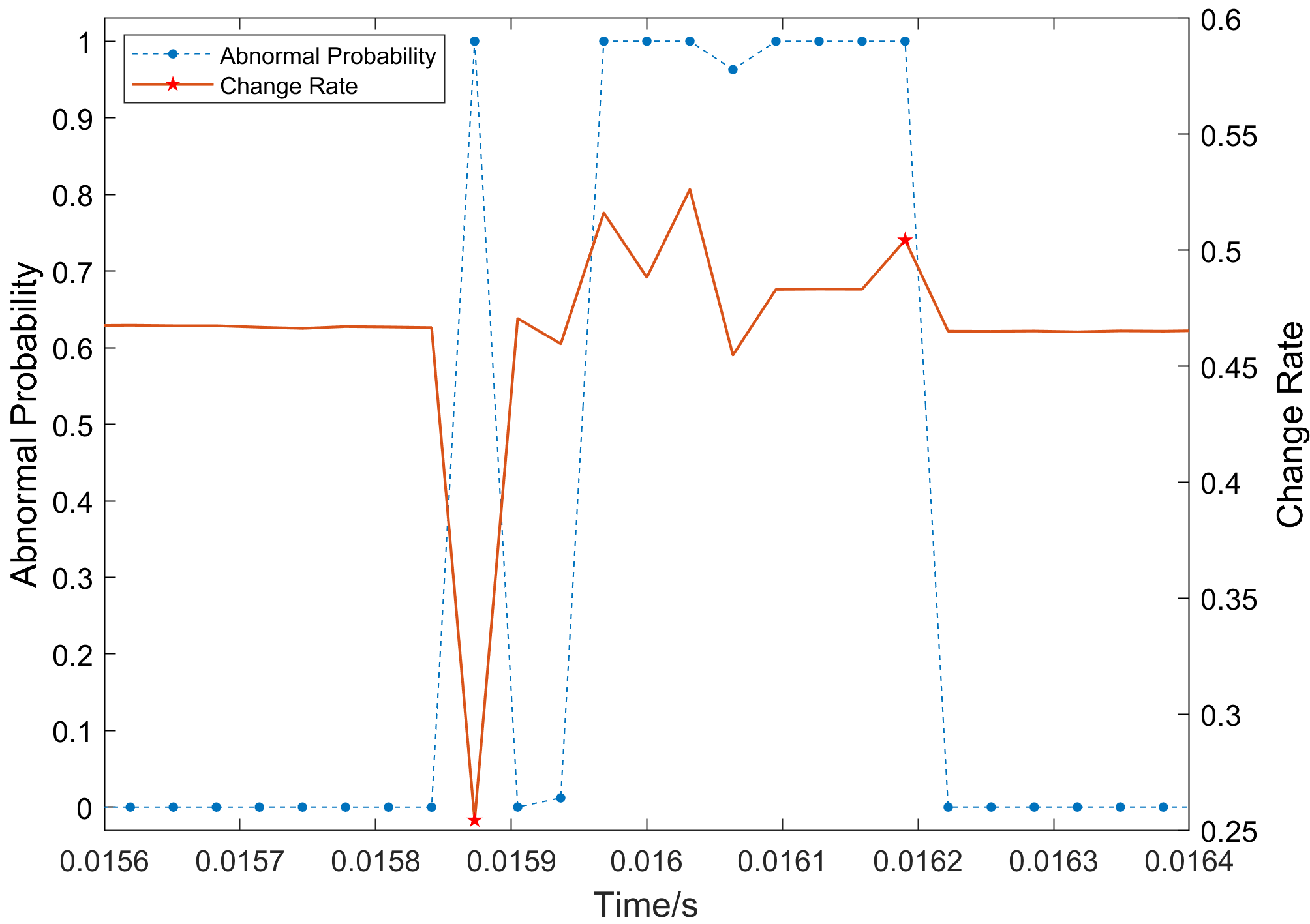}
			\end{minipage}
		}
		\caption{\label{result3}Left: Circuit System Output with Single-component Voltage Input In Time Segment 495-515; Right: EMODM Results Applied in Local Time with Abnormal Time Segment 501-510.}
		\subfigure{
			\includegraphics[width=0.95\textwidth]{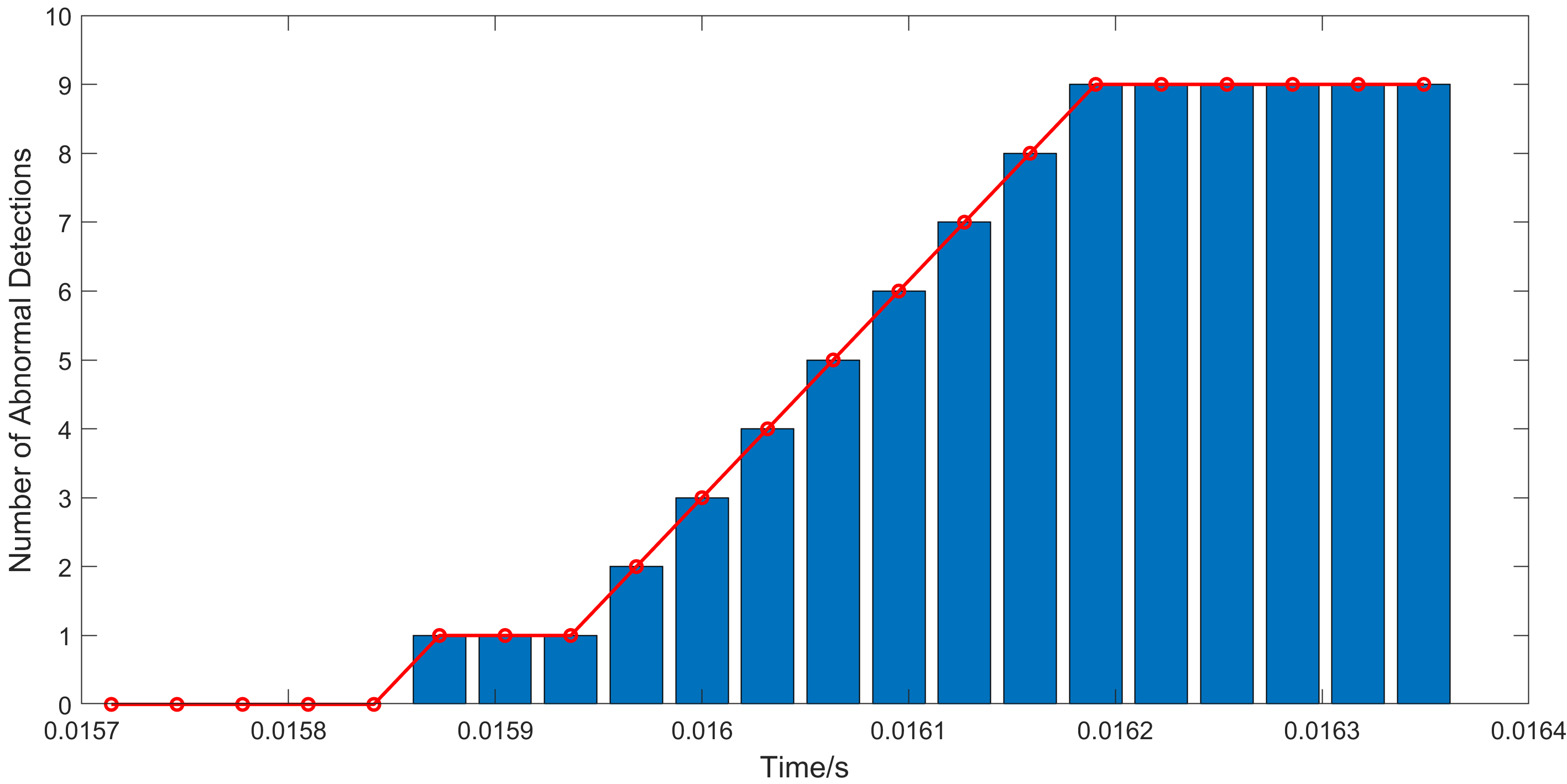}
		}
		\caption{\label{result4} Online Abnormal Detection Result: Abnormal Time Periods Capture Number of Circuit System with Single-component Voltage Input In Time Segment 495-515 with Abnormal Time Segment 501-510.}
	\end{figure}

	Since the fault the large parameter drift of the circuit will cause the system to enter an abnormal pattern, we assume the resistance R1 conforms to Gaussian distribution as $\mathcal{N}(1000,1000)$, capacitance C1 and C2 conform to Gaussian distribution as $\mathcal{N}(2,1)$ and $\mathcal{N}(2,1)$, where the unit of resistance is Omega and the unit of capacitance is Farad. We set the rejection threshold as the 4\,\% two-tail probability of the result distribution to get the outliers. By sampling from this domain and averaging the sample results, we obtain the estimated value at the corresponding time period. After repeating the above process and combining the corresponding results of the correct working and fault time segments, the whole simulation results of the circuit can be obtained. We set the initial phase of the single-component voltage as zero, the expressing formula is as follows
	\begin{equation}\label{eq:3.6}
		U_{\text{in}}=100\mathrm{sin}(800\pi t).
	\end{equation}
	The outputs of the Sallen-Key low-pass filter circuit system with single-component voltage input are shown on the left-hand side of figure \ref{result1}. The red pentagons represent the probability of fault occurrence and termination time points in the Sallen-Key low-pass filter circuit system.  
	
	First, the EMODM is used to detect faults in the outputs of circuits with single-component voltage input. The mixture probability model of the two-state Gaussian distribution could be obtained by utilizing the double-component EMODM in section \ref{Second_sec} and employed for later work on circuit system fault diagnosis in this study. The statistical inference results derived from the EMODM are shown in table \ref{tab.1}. It can be noticed that the abnormal pattern in two-state GMM always has a larger variance, so the probability of generating outliers is greater. Meanwhile, the percentage of the abnormal time segment in the single-component voltage and double-component voltage input experiment in the numerical simulation data is set to 30/630=4.76\,\%, which is less than 5\,\%, which corresponds to a small probability abnormal scenario for practical applications in statistical significance. So the circuit system is more likely to output outliers when it enters a small probability of abnormal pattern. Thus we can use the mixture model obtained to detect the system's abnormal pattern caused by circuit faults. Here, we use the relative change rate of the time series data from the circuit output and the corresponding system abnormal probabilities to obtain the fault diagnosis visualization results based on the EMODM for the Sallen-Key low-pass filter circuit system with single-component voltage input shown on the right-hand side of figure \ref{result1}. Note that the red orbit in this figure represents the change rate of the circuit output and the blue orbit represents the probabilistic output results based on the EMODM.
	
	\begin{table}[H]
		\centering
		\scalebox{0.9}{
		\begin{tabular}{c|cccc}
			\hline
			\hline
			Voltage Input&Pattern&Proportion&Mean&Variance\\
			\hline
			Single-& 1:\,Correct&95.34\,\%&-0.0017&2.1996\\
			component& 2:\,Abnormal& 4.66\,\%&0.7997&2.0397e$^4$ \\
			\hline
			Double-& 1:\,Correct & 95.72\,\% & 0.0471 & 13.2613\\
			component& 2:\,Abnormal& 4.28\,\% & 0.6727 & 2.9164e$^4$ \\
			\hline
			\hline
			Voltage Input &Abnormal Detection &Setting Abnormal& Abnormal Detection & Abnormal Detection  \\
			&Time Periods &Time Segment&Time Segment &Segment Ratio \\
			\hline
			Single- & 9&151-160&151-160 &100\,\%\\
			component & 4 &211-220&211-220 &100\,\%\\
			& 9 &501-510&501-510&100\,\%\\
			\hline
			Double-& 4 &151-160 &151-158 & 80\,\%  \\
			component& 6 &211-220& 211-220 &100\,\% \\
			&6 &501-510&501-510&100\,\%\\
			\hline
			\hline
		\end{tabular}}
		\caption{\label{tab.1}EMODM Results for Sallen-Key Low-pass Filter Circuit System with Single-component and Double-component Voltage Input.}
	\end{table}
	
	According to the results shown in table \ref{tab.1}, all three abnormal time segments and the most of abnormal time periods that existed in the circuit system were detected by our EMODM. In the single-component voltage input experiment, the abnormal time periods detected in the circuit system global time outputs accounted for 22/30=73.33\,\% of the total abnormal time periods. All three abnormal time segments of circuit systems set up were successfully diagnosed and the global fault diagnosis rate of the EMODM in detecting the occurrence of three faults is 100\,\%. Compared to the artificially fixed anomaly ratio of 4.76\,\%, the percentage of the abnormal time segment in the single-component voltage input experiment in the numerical simulation data is 4.66\,\%. By analyzing the above results, it can be proved that the EMODM proposed in this paper has a very impressive diagnostic capacity on the circuit time series output with fault. Moreover, figure \ref{result3} shows the output of the circuit system with single-component voltage input in a selected time segment containing an abnormal segment and demonstrates the magnified details of the EMODM at a short local time segment in time periods 495-515. Based on the numerical modeling of single-component voltage input circuit systems, we utilize the EMODM repeatedly to perform online abnormal detection experiments by using the growing time series data in a selected local time segment shown in figure \ref{result3} with the third abnormal time segment of time periods 501-510. In figure \ref{result4}, we find that the EMODM can already achieve real-time capture of faults within a delay of one observation time of the fault occurrence. During the abnormal time segment, a new abnormal time segment is constantly detected and an alarm signal could be immediately issued. Furthermore, the method would not continue to incorrectly detect new abnormal time points after the circuit system reenters the correct pattern.
	
\begin{figure}[H]
		\centering
		\begin{minipage}{0.48\textwidth}
			\centering
			\includegraphics[scale=0.45]{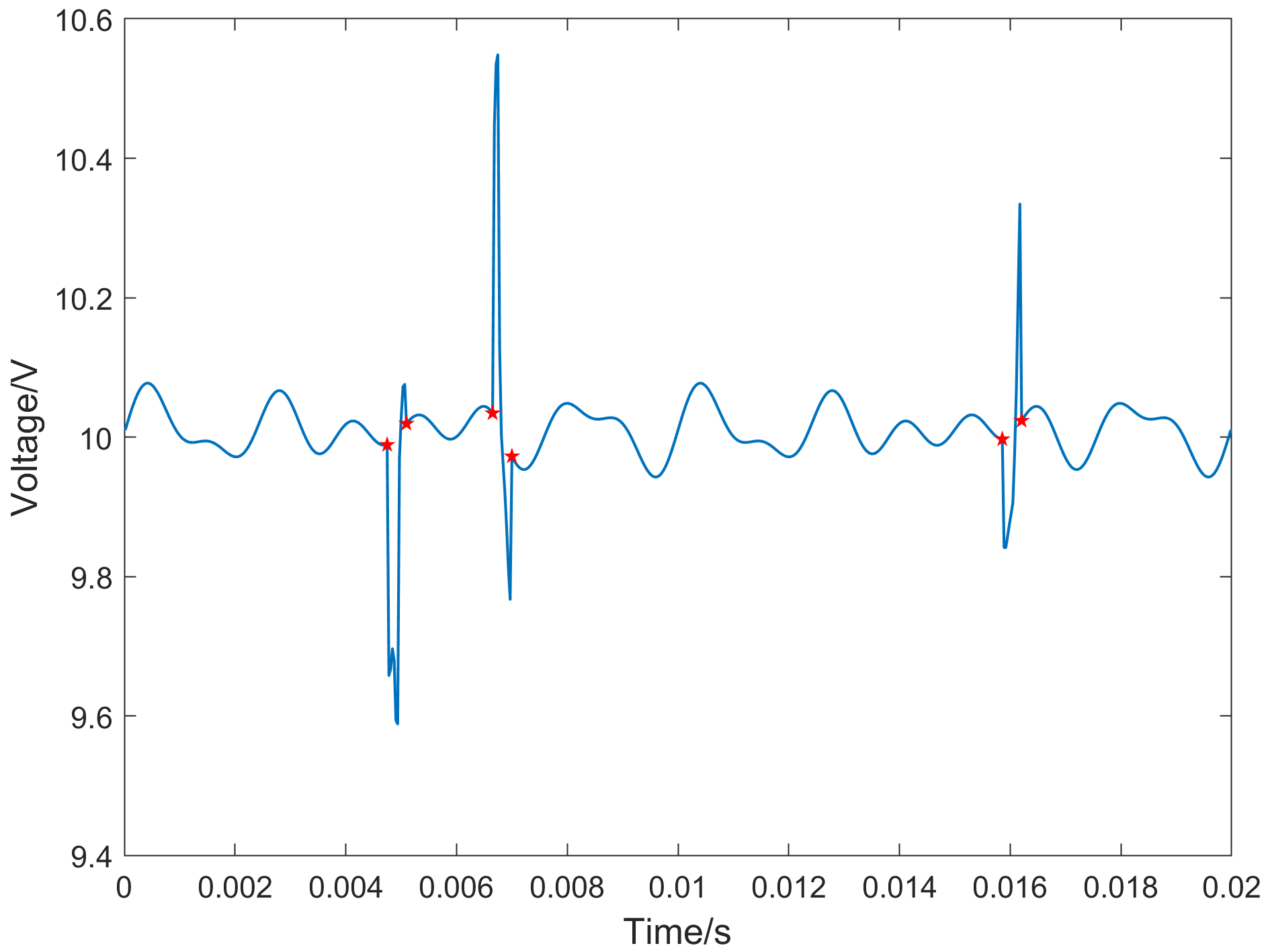}
		\end{minipage}
		\begin{minipage}{0.48\textwidth}
			\centering
			\includegraphics[scale=0.45]{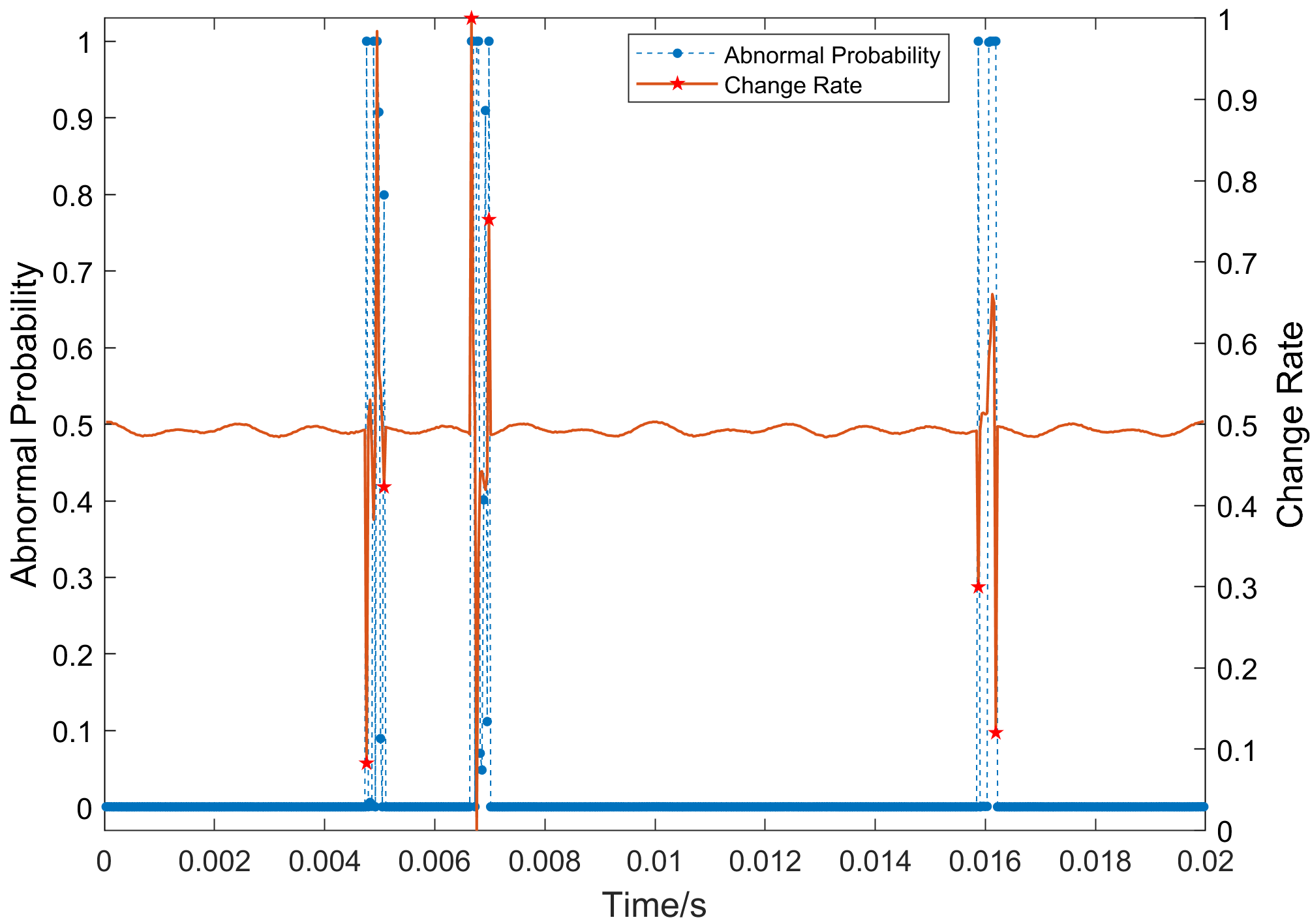}
		\end{minipage}
		\caption{Left: Sallen-Key Low-pass Filter Output with Double-component Voltage Input; Right: EMODM Results by Change Rate of System Output.}
		\label{result5}
	\end{figure}

	Furthermore, we consider a bit more complicated complex double-component voltage input. We set the initial phase of the double-component voltage as zero, with the unchanged initial value, the expressing formula is as follows
	\begin{equation}\label{eq:3.7}
		U_{\text{in}}=100\mathrm{sin}(800\pi t)+100\mathrm{sin}(1600\pi t).
	\end{equation}
	Its output is shown on the left-hand side of figure \ref{result5}. The EMODM is used to detect potential faults in the outputs of circuits with double-component voltage input as same as the previous single-component voltage input experiment. The EMODM results for the Sallen-Key low-pass filter circuit system with double-component voltage input are also shown in table \ref{tab.1}. The results of the probabilistic visualization based on the change rate of the circuit output are presented on the right-hand side of figure \ref{result5}. The percentage of the abnormal component mixture in the double-component voltage input experiment is 4.28\,\% which is close to the real setting abnormal time percentage of 4.76\,\%. It can be seen that the performance of EMODM is degraded in complex situations but is completely acceptable. Since the same three abnormal segments were correctly detected, it can be considered that the effectiveness of our method in the complicated case was further verified.
	
	\subsection{Information Processing and Storing for Spin-magnet Circuits}\label{3.2} 
\begin{figure}[H]
		\centering
		\includegraphics[width=0.9\textwidth]{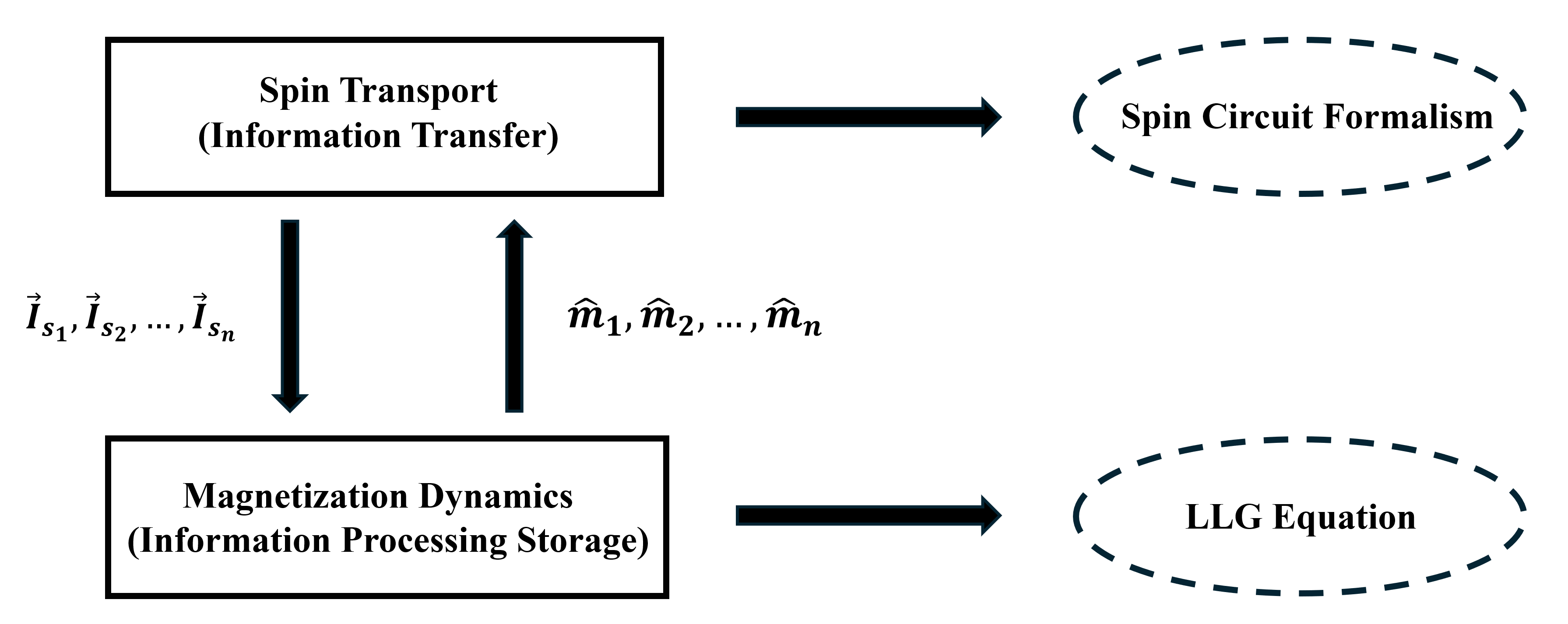}
		\caption{\label{Information Passing Process} A Spin-transport/magnetization-dynamics Model for Spin-magnet Circuits.}
	\end{figure}

	In this case, we take the classical Landau-Liftshitz-Gilbert(LLG) equation which is closely related to the information processing and storing in the device that uses spin currents and nanomagnets \cite{srinivasan2013modeling}. A description of magnetization dynamics containing the information processing and storing within the magnet is shown in figure \ref{Information Passing Process}. This figure shows how information is processed and stored in the nanomagnets which are described by the standard Landau-Lifshitz-Gilbert equation.

	\begin{table}[H]
		\centering
		\begin{tabular}{c|cc}
			\hline
			\hline
			Variable & Symbol & Value  \\
			\hline
			Unidirectional anisotropy constant& $k_{u}$& 3.14e$^{4}$\,(erg)/cm$^{3}$ \\
			Gyromagnetic ratio& $\Gamma$ & 1.76e$^{7}$\,(rad)/(Oe $\cdot$ s) \\
			Gilbert damping parameter& $\lambda$ & 0.007 \\
			Dimensionless demagnetizing field without heat& $h_{d}$ & 0\,Oe \\
			Coulombs& q& 1.6e$^{-19}$ \\
			Saturation Magnetization& $M_{s}$& 780\,(emu)/cm$^{3}$ \\
			volume& $V$& 2.72e$^{-17}$\,cm$^{3}$\\
			Bohr magneton& $\mu_{B}$& 9.274e$^{-21}$ \\
			\hline
			\hline
		\end{tabular}
		\caption{\label{tab.6} The Parameters of LLG Equation Taken From the Experiment.}
	\end{table}

	A typical example of that spin-magnet system mentioned above is the spin-transfer torque magnetic random access memory(STT-MRAM) with fast write speed and other ideal properties\cite{lu2016high}. Due to its high-frequent spin-transfer for the information passing process, the write error rate of STT-MRAM is a key factor that may significantly improve the reliability of the whole system\cite{nowak2016dependence}. Here, we implement our EMODM on the LLG equation and accurately identify the abnormal pattern in that process. The standard Landau-Lifshitz-Gilbert(LLG) nonlinear equation describes the dynamics of instantaneous magnetization which is denoted as $\Vec{m}$ of a magnet subject to the spin currents and is numerically solved by the varied order Adams-Bashforth-Moulton PECE solver\cite{srinivasan2013modeling}. The standard normalized spherical LLG equation is as follows
	\begin{equation}
		(1+\lambda^2)\frac{d\Vec{m}}{dt}=-|\Gamma|(\Vec{m} \times \Vec{H})-\lambda|\Gamma|(\Vec{m} \times \Vec{m} \times \Vec{H})+\Vec{\tau}+\lambda(\Vec{m} \times \Vec{\tau}),
	\end{equation}
	where spin torque $\Vec{\tau}$ is
	\begin{equation}
		\Vec{\tau}=\frac{\Vec{m}\times\Vec{I_{s}}\times\Vec{m}}{qN_{s}},
	\end{equation}
	with $\Vec{I_{s}}$ as electrical current, q as the charge of an electron, and $N_{s}$ as the total number of spins in the nanomagnet which is defined as:
	\begin{equation}
		N_{s}=M_{s}V/\mu_{B}, 
	\end{equation}
	where $M_{s}$ denotes the saturation magnetization, $V$ represents the volume and $\mu_{B}$ refers to the Bohr magneton. In this experiment, we assume the magnet $\Vec{m}$ is a mono-domain whose orientation is
	\begin{equation}\label{eq:3.8}
		\Vec{m}=[m_{x}\quad m_{y}\quad m_{z}] = [\text{sin}\theta \text{cos}\phi\quad \text{sin}\theta \text{sin}\phi\quad \text{cos}\theta],
	\end{equation}
	and its rotation starts from the x-z plane. All values of parameters, in this case, are listed in the following table \ref{tab.6} which are taken as dimensionless by using the constant $M_{s}/2k_{u}\Gamma$.
	
\begin{figure}[t]
		\centering
		\subfigure{
			\begin{minipage}{0.48\textwidth}
				\centering
				\includegraphics[scale=0.45]{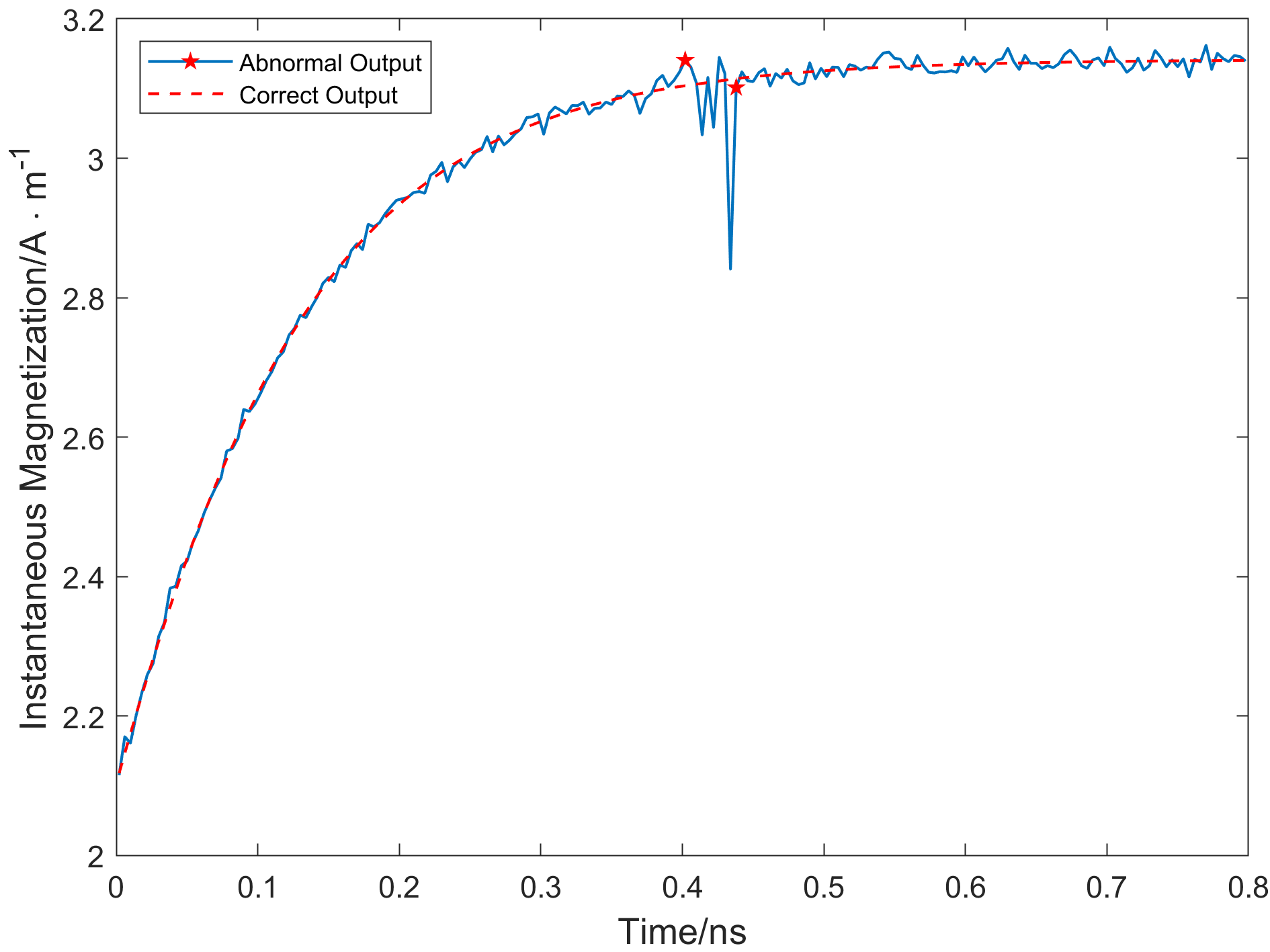}
			\end{minipage}
			\begin{minipage}{0.48\textwidth}
				\centering
				\includegraphics[scale=0.45]{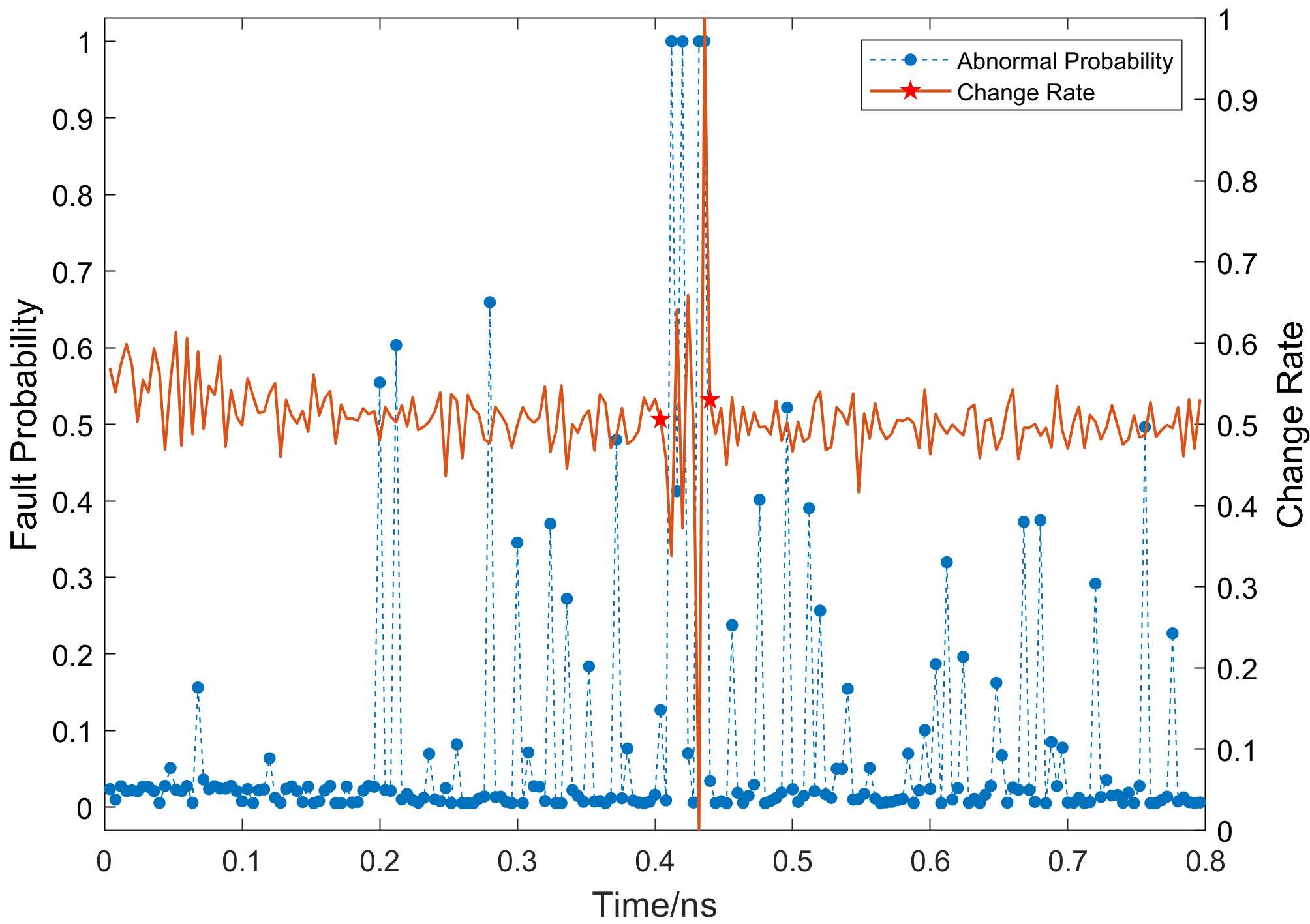}
			\end{minipage}
		}
		\caption{\label{llg2}Left: The Single-fault Output of Instantaneous Magnetization $\Vec{m}$ of the LLG Equation adds 1\,\% Gaussian White Noise; Right: EMODM Results for Instantaneous Magnetization $\Vec{m}$ of the LLG Equation.}
		\subfigure{
			\begin{minipage}{0.48\textwidth}
				\centering
				\includegraphics[scale=0.45]{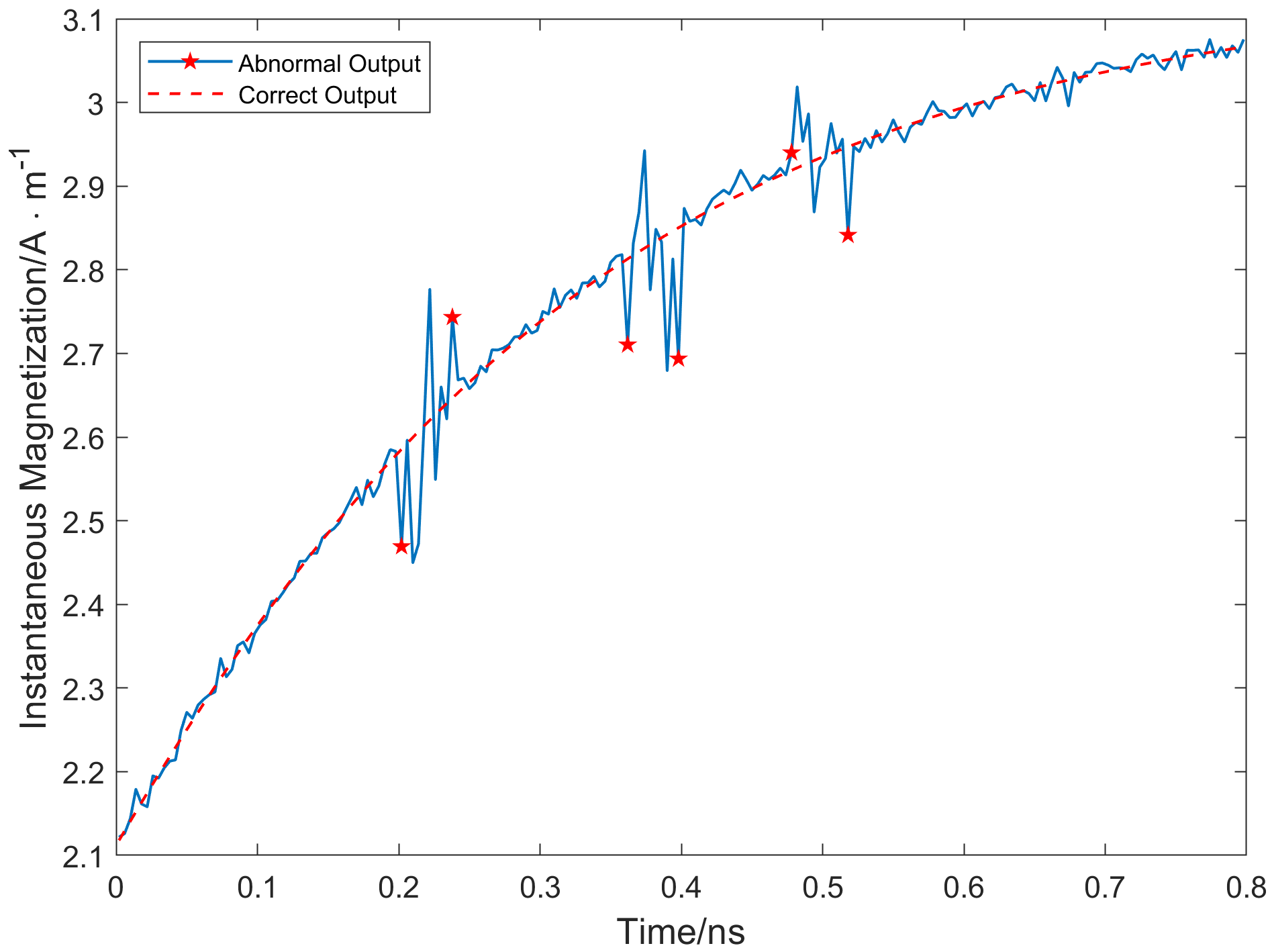}
			\end{minipage}
			\begin{minipage}{0.48\textwidth}
				\centering
				\includegraphics[scale=0.45]{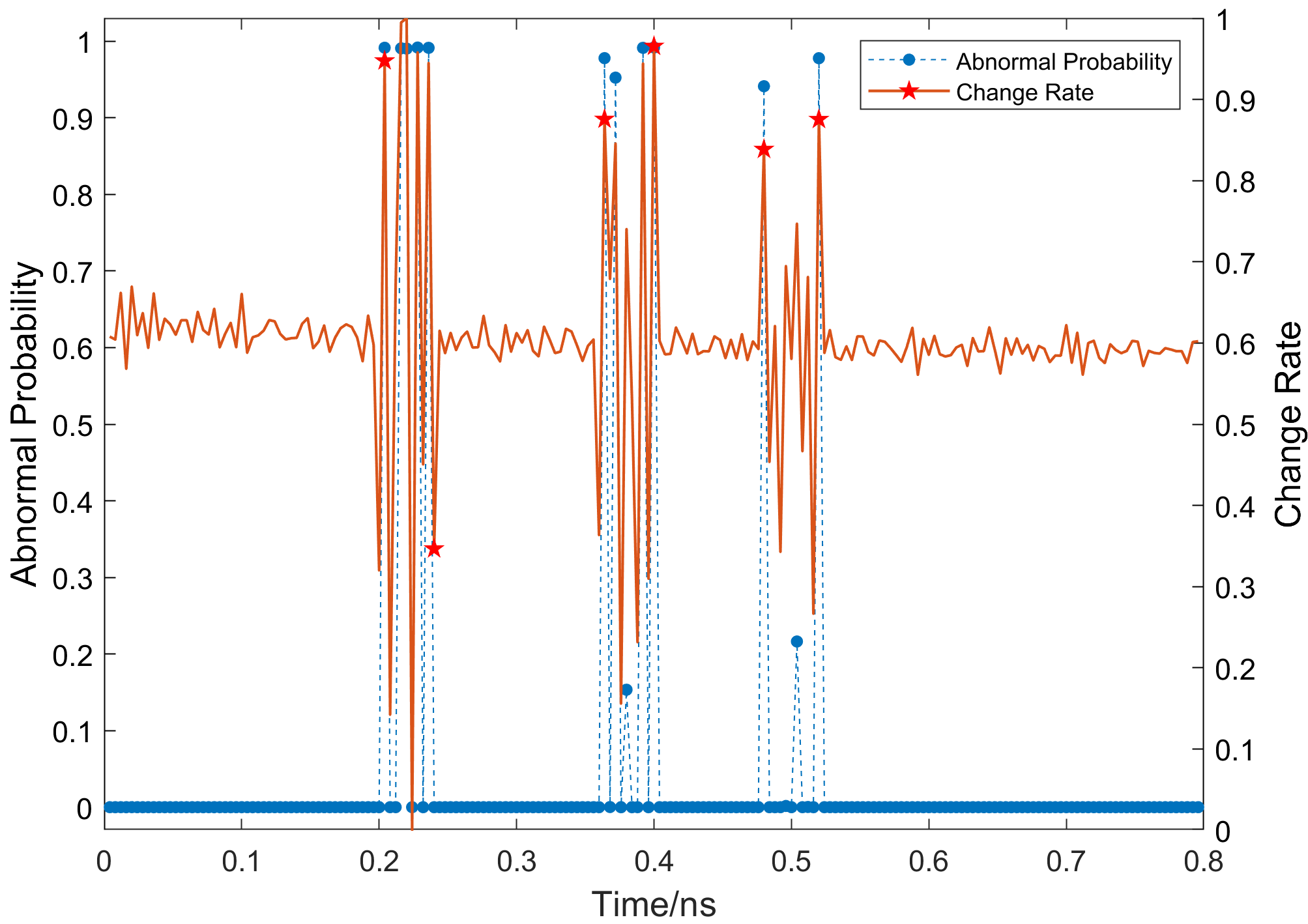}
			\end{minipage}
		}
		\caption{\label{llg3}Left: The Multi-fault Output of Instantaneous Magnetization $\Vec{m}$ of the LLG Equation adds 1\,\% Gaussain White Noise; Right: EMODM Results for Instantaneous Magnetization $\Vec{m}$ of the LLG Equation.}
	\end{figure}

	In this study, we set the sum of internal fields and external fields on the magnet as:
	\begin{equation}\label{eq:3.9}
		\Vec{H}=[0\enspace -h_{d}m_{y}\enspace m_{z}], 
	\end{equation}
	where $h_{d}$ is the dimensionless demagnetizing field without heat. There are three considered parameters, initial azimuth angle $\theta_{0}$, current magnitude $I_{s}$, and the final time, separately. The total time length of the simulation is 0.8\,ns divided into 200 time periods in the following experiments. In the first experiment, we set up only one abnormal time segment which contains 10 time periods and two correctly working time segments contain 190 time periods. In another multi-fault case, we set three abnormal pattern time segments containing 30 time periods and four correctly working time segments containing 170 time periods. In both experiments, the current magnitude is set as 1.814e$^{-4}$\,A and we generated the fault caused by the random initial azimuth angle obeying Gaussian distribution $\mathcal{N}(\pi/4,\pi/12)$. The correct output of instantaneous magnetization $\Vec{m}$ of the LLG equation is shown in figure \ref{llg2}, where the initial azimuth angle is $\pi/4$. The red pentagons represent the fault occurrence and termination time points in the information processing and storing for spin-magnet circuits.  
	
	\begin{table}[H]
		\centering
		\scalebox{0.9}{
		\begin{tabular}{c|cccc}
			\hline
			\hline
			Experiment  & Setting Abnormal & Abnormal Detection & Abnormal Detection & Abnormal Detection \\
			Setup & Time Segment & Time Segment & Segment Ratio & Time Periods \\
			\hline
			Single-fault Case & 101-110 & 103-110 & 80\,\% & 4 \\
			\hline
			& 51-60 & 51-60 & 100\,\% & 5\\
			Multi-fault Case & 91-100 & 91-100 & 100\,\% & 4 \\
			& 121-130 & 121-130 & 100\,\% & 2 \\
			\hline
			\hline
		\end{tabular}}
		\caption{\label{tab.7} EMODM  Results for Instantaneous Magnetization $\Vec{m}$ of the LLG Equation.}
	\end{table}
	
	There is usually corresponding Gaussian white noise when observing the circuit system. The observation noise is described as illustrated in section \ref{Third_sec}. In this part, we verify the fault diagnosis capability of the EMODM under 1\,\% Gaussian white noise disturbance. We followed the same procedure as in the previous experiment. Here, we use the relative change rate of the output data from the system including the abnormal pattern, and obtain the fault diagnosis results based on the EMODM for instantaneous magnetization $\Vec{m}$ of the LLG equation presented in table \ref{tab.7}. According to these results, all abnormal time segments in the circuit system were detected by our method. In single-fault and multi-fault experiments, the global fault diagnosis rate in detecting the occurrence of faults is 100\,\%. Under standard noise disturbance, there is no missed and mistaken detection of faults. The visualization of the two experimental results is shown in figure \ref{llg2} and \ref{llg3}. The red orbit represents the change rate of the instantaneous magnetization $\Vec{m}$ of the LLG equation and the blue orbit represents the abnormal probability given by the EMODM. It can be seen that this method can capture abnormal time points continuously in the abnormal output segment of the circuit system. No faults are falsely reported in the correct working segments. The EMODM is well adapted to circuit outputs with varying features caused by faults. Therefore in this experiment, we verified the robustness and stability of the EMODM based on the noise data of the nonlinear circuit system. Further by its online detection feature, people could receive an alarm signal from the algorithm program as soon as it accurately captures the abnormal time point with its characterization.

	\section{Application in real-world datasets}\label{forth_sec}
	In this section, our EMODM proposed in this paper for machine learning for complex systems with an abnormal pattern has been experimentally verified in two real-world datasets on a 7kVA three-phase three-wire inverter and the U.S. insured unemployment dataset.

	\subsection{7kVA Three-phase Three-wire Inverter}
\begin{figure}[htp]
		\centering
		\includegraphics[width=0.9\textwidth]{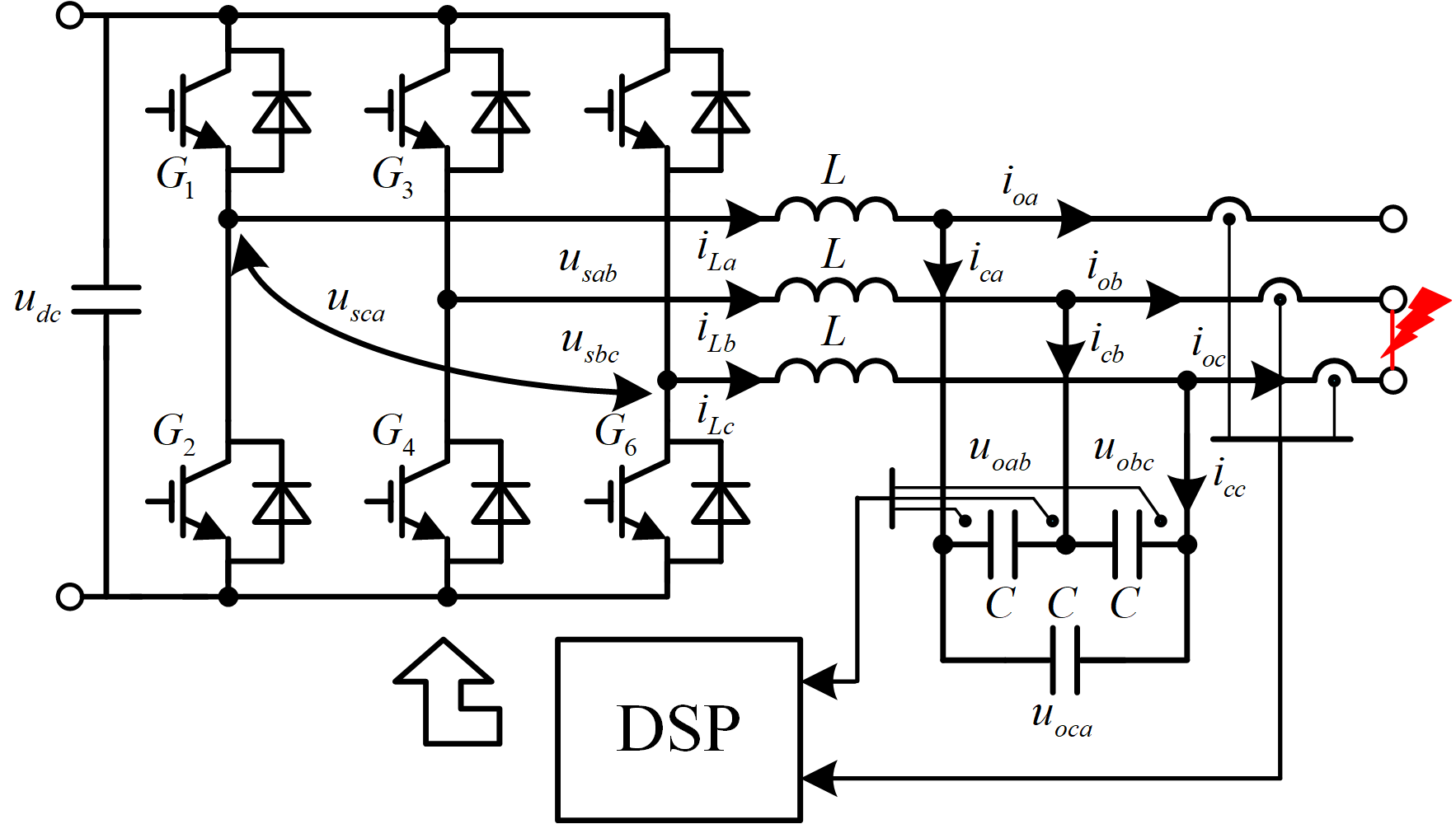}
		\caption{\label{exper1} Topology structure of Three Phase Inverter with A Short Circuit}
	\end{figure}

	The three-phase inverter includes a digital signal processor, a direct current power supply, three insulated gate bipolar transistor half bridges, and a three-phase LC filter\cite{pei2012short}. The abnormal pattern of this complex circuit system is to suddenly experience a short circuit somewhere during normal operation, where the time-series data from the voltmeter and ammeter indications are the outputs of the system. The $i_{La}$, $i_{Lb}$ and $i_{Lc}$ are the inductor currents. The $i_{ca}$, $i_{cb}$ and $i_{cc}$ are the capacitor currents. The $i_{oa}$, $i_{ob}$ and $i_{oc}$ are the output currents. The $u_{oab}$, $u_{obc}$ and $u_{oca}$ are the output line voltages. The experiment simulates the phase-to-phase short circuit condition of the three-phase three-wire inverter by using a circuit breaker on the load side. During a short circuit, the faulted phase inductor output current is limited to 40$A$ and the output current is approximately 42.75$A$. The figure \ref{exper1} and table \ref{tab.8} respectively give the parameters and topology diagram of the three-phase inverter.
	
	\begin{table}[H]
		\centering
		\begin{tabular}{c|cc}
			\hline
			\hline
			Variable & Symbol & Value  \\
			\hline
			Rated capacity& $S$ & 7 kVA \\
			Rated voltage& $V_{\text{RMS}}$ & 209 V \\
			Rated current& $I_{\text{RMS}}$ & 19.33 A\\
			Filter inductor& $L$ & 0.56 mH \\
			Filter capacitor& $C$ & 60 $\upmu$F \\
			Fundamental frequency& $f$ & 50 Hz \\
			DC bus voltage& $V_{\text{dc}}$ & 450 V \\
			\hline
			\hline
		\end{tabular}
		\caption{\label{tab.8} The Parameters Setting of Three-phase Inverter.}
	\end{table}

\begin{figure}[p]
		\centering
		\subfigure{
			\begin{minipage}{0.24\textwidth}
				\centering
				\includegraphics[scale=0.23]{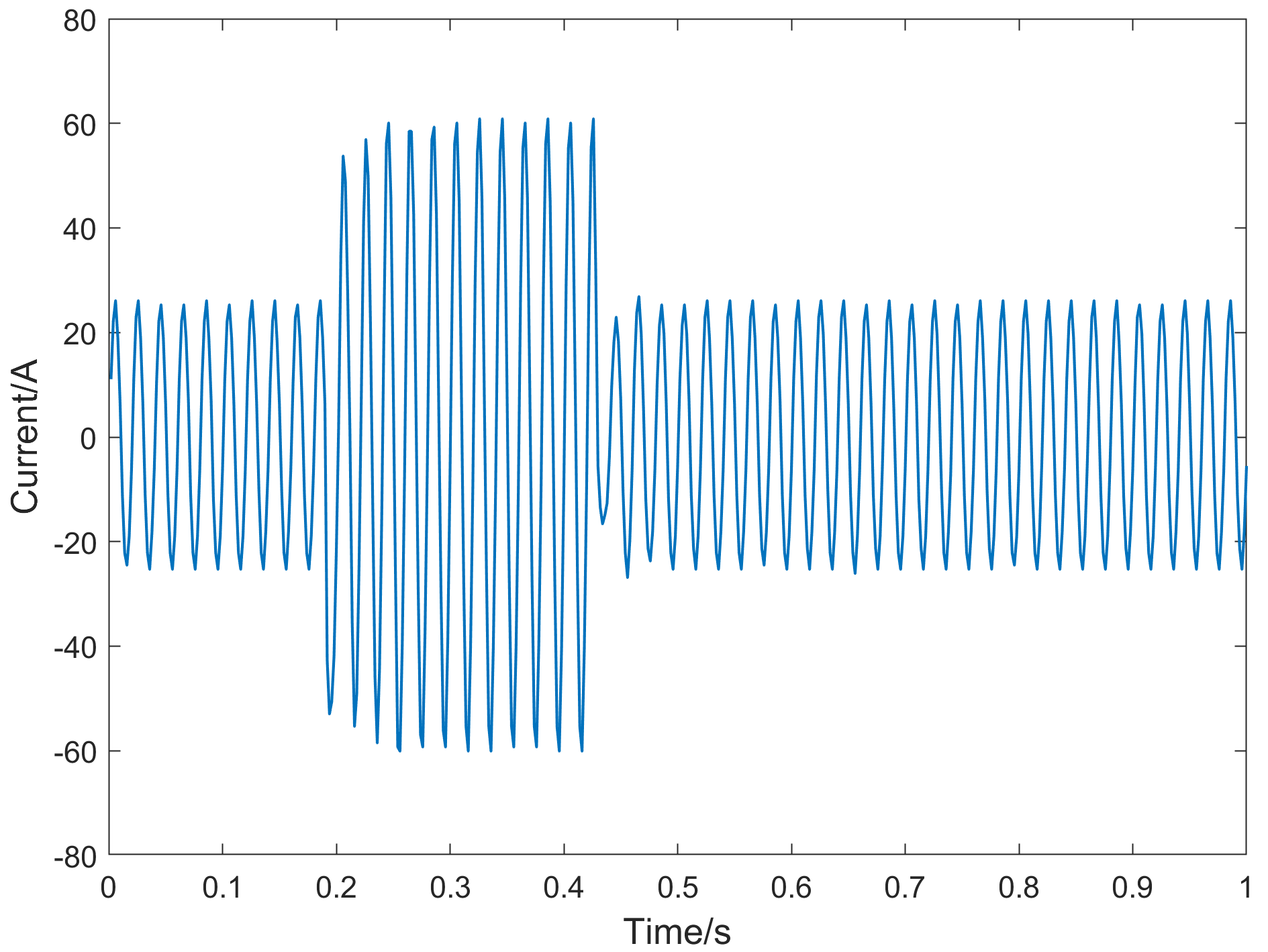}
			\end{minipage}
			\begin{minipage}{0.24\textwidth}
				\centering
				\includegraphics[scale=0.23]{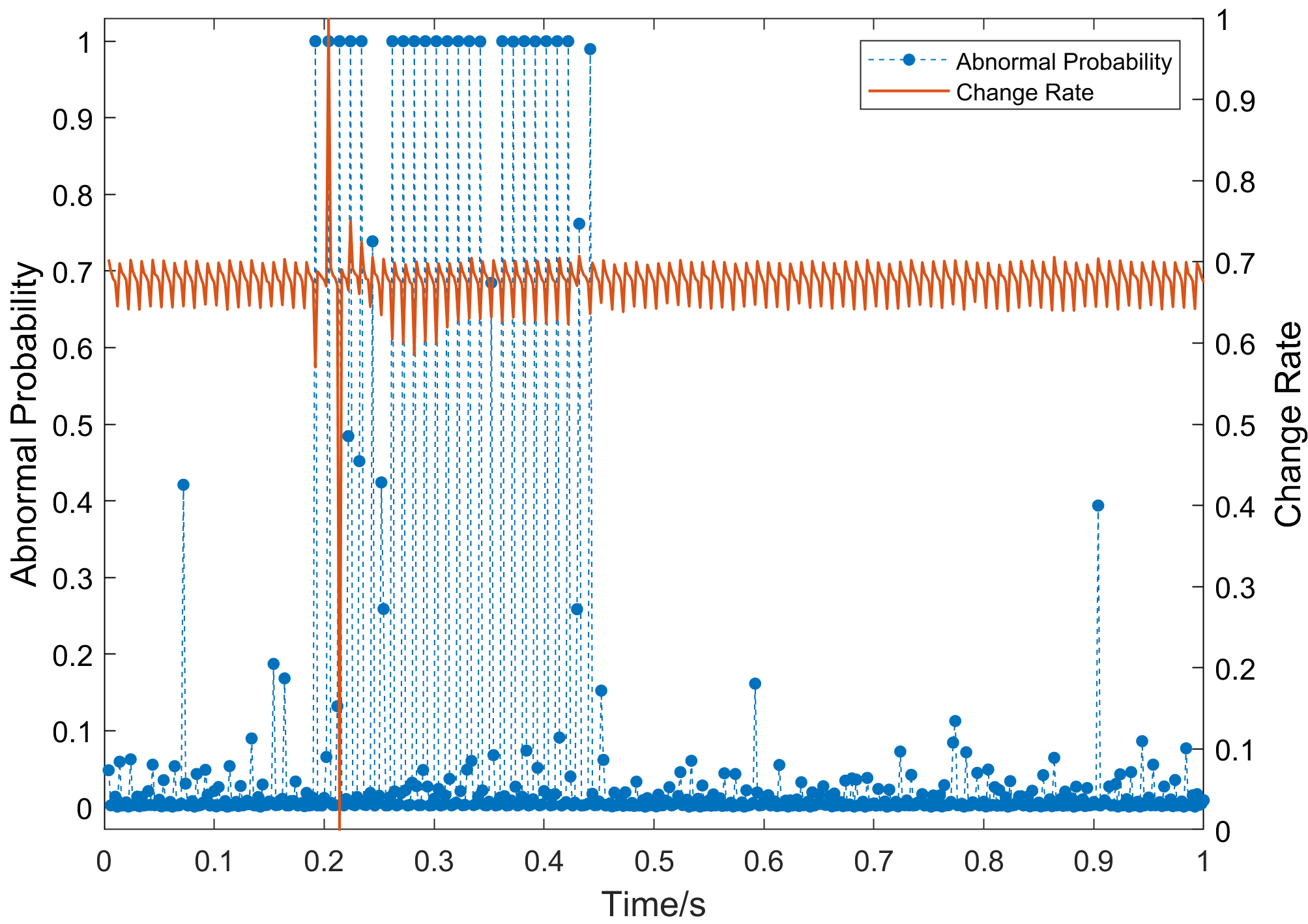}
			\end{minipage}
			\begin{minipage}{0.24\textwidth}
				\centering
				\includegraphics[scale=0.23]{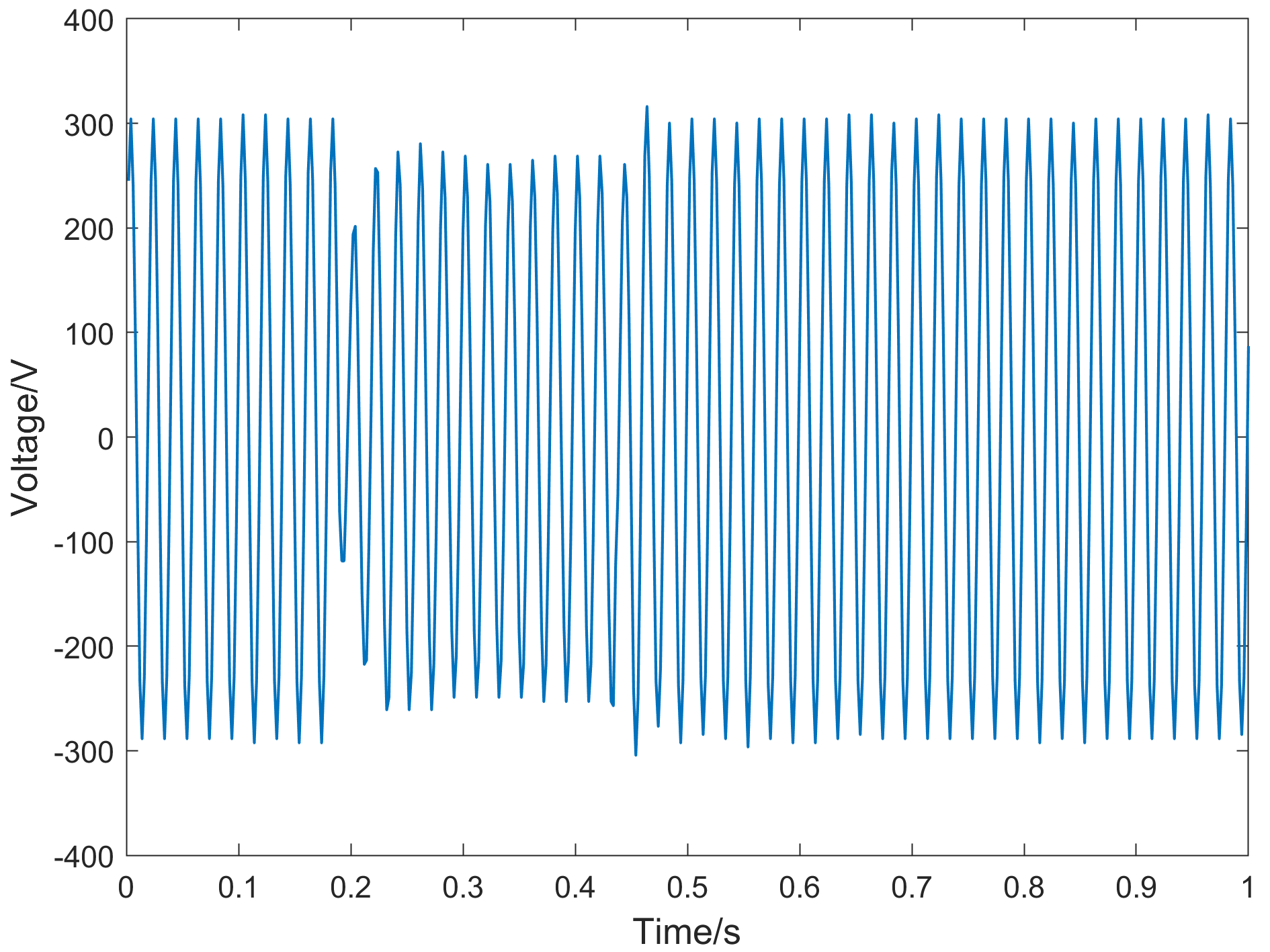}
			\end{minipage}
			\begin{minipage}{0.24\textwidth}
				\centering
				\includegraphics[scale=0.23]{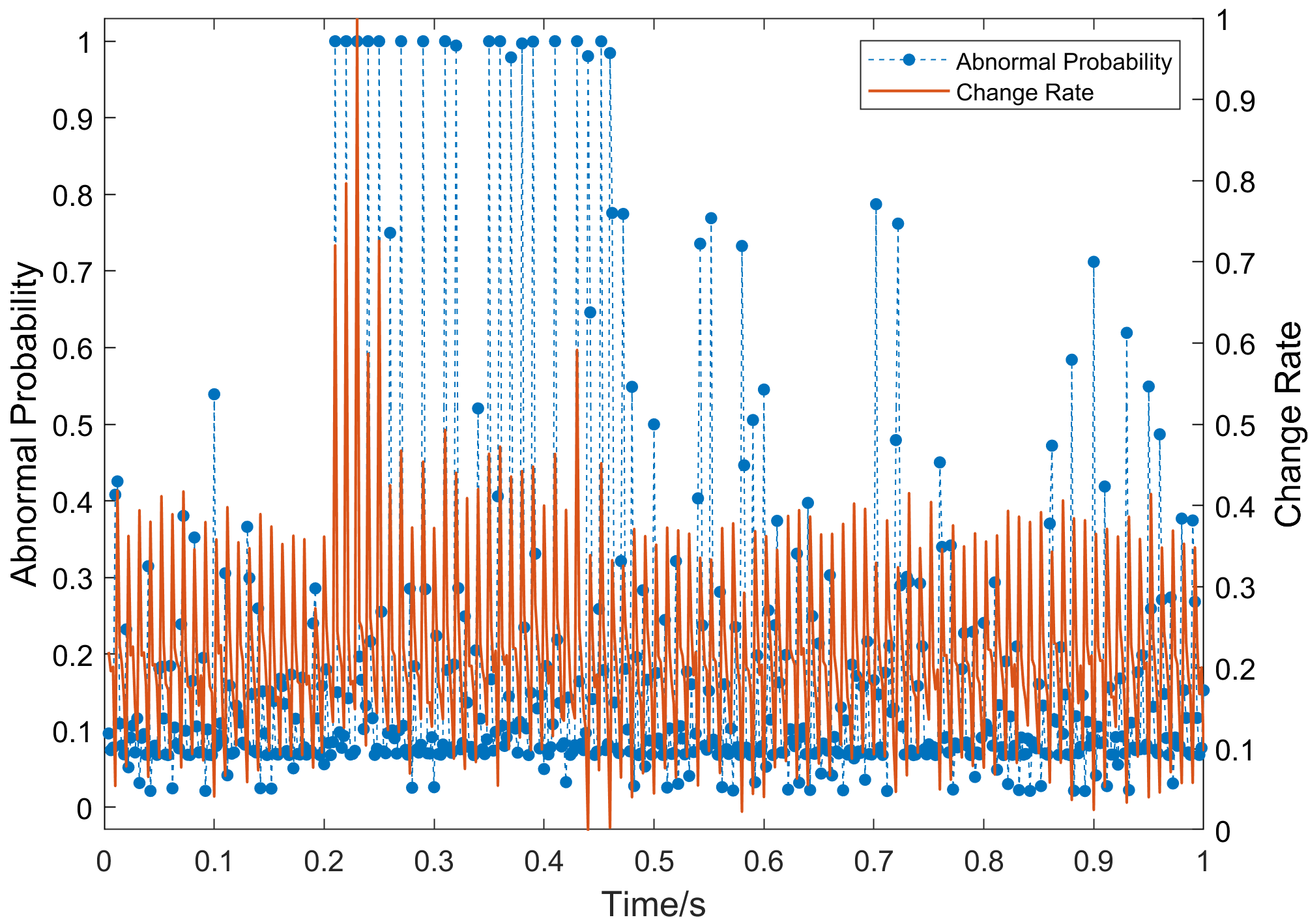}
			\end{minipage}
		}\\
		\caption{Output of Complex Circuit System with A Short Circuit and EMODM Results. Left: Current Signal Output $i_{oa}$; Right: Voltage Signal Output $u_{ab}$.\label{exper2}}
		
		\subfigure{
			\begin{minipage}{0.24\textwidth}
				\centering
				\includegraphics[scale=0.23]{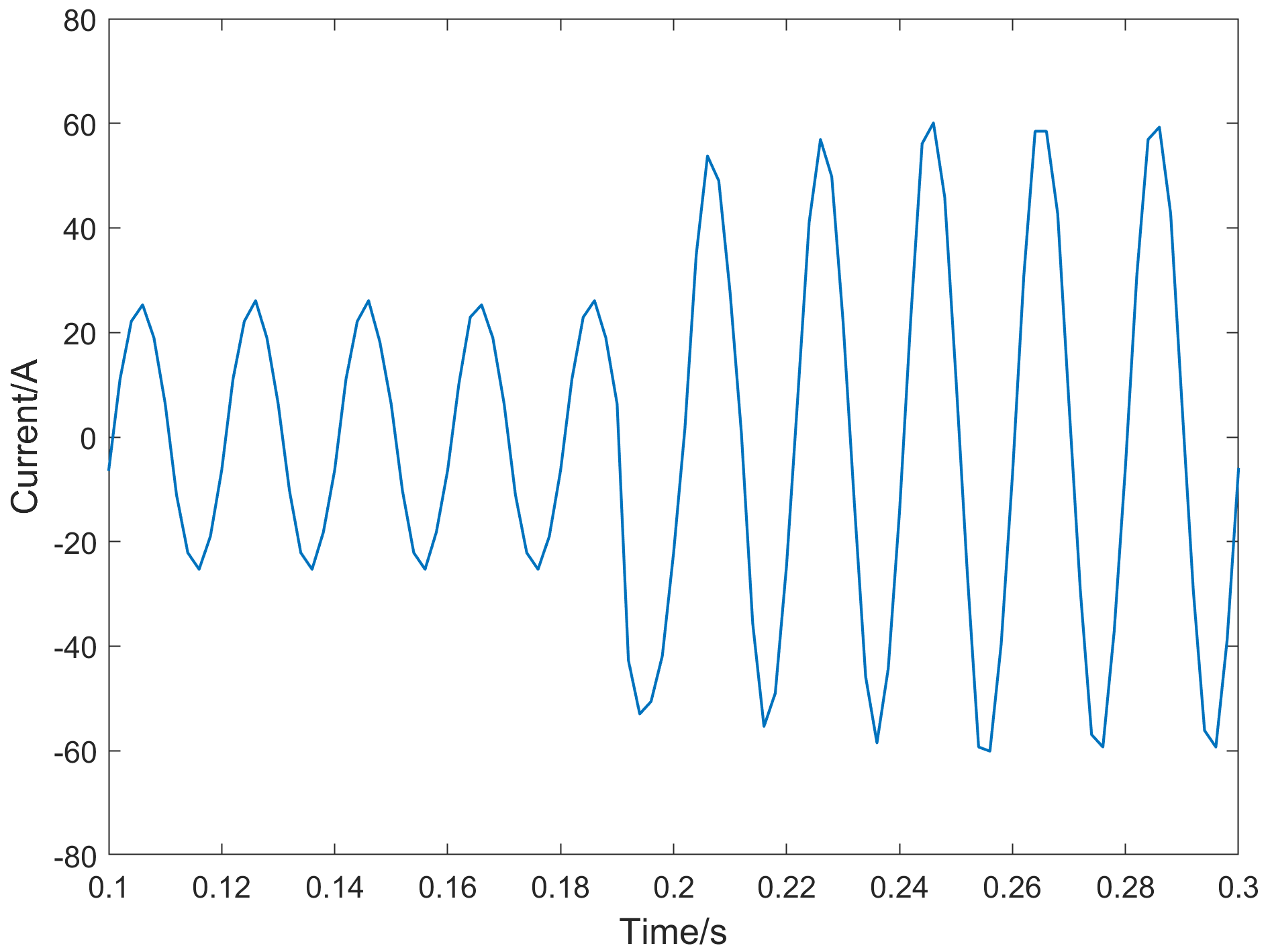}
			\end{minipage}
			\begin{minipage}{0.24\textwidth}
				\centering
				\includegraphics[scale=0.23]{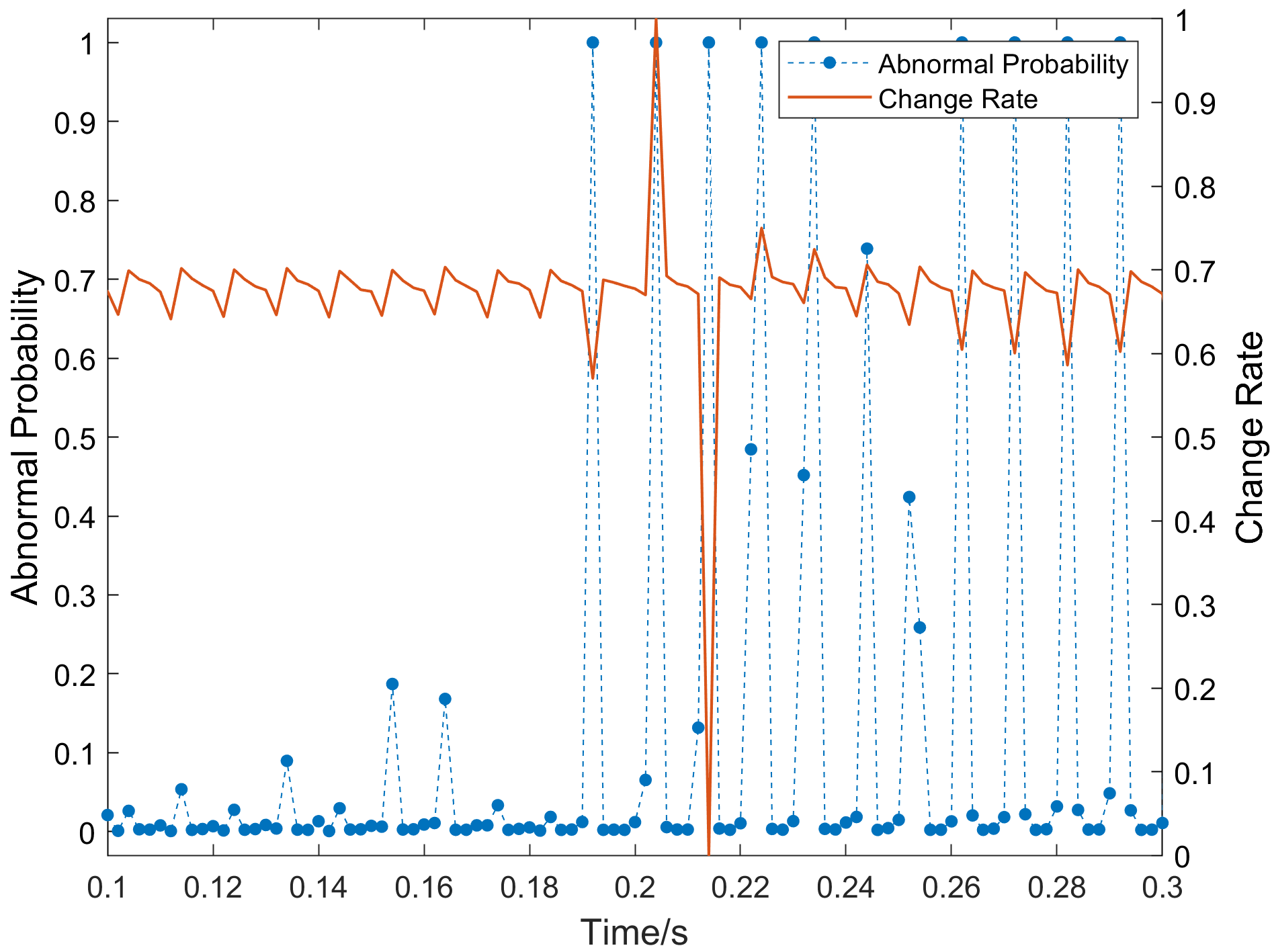}
			\end{minipage}
			\begin{minipage}{0.24\textwidth}
				\centering
				\includegraphics[scale=0.23]{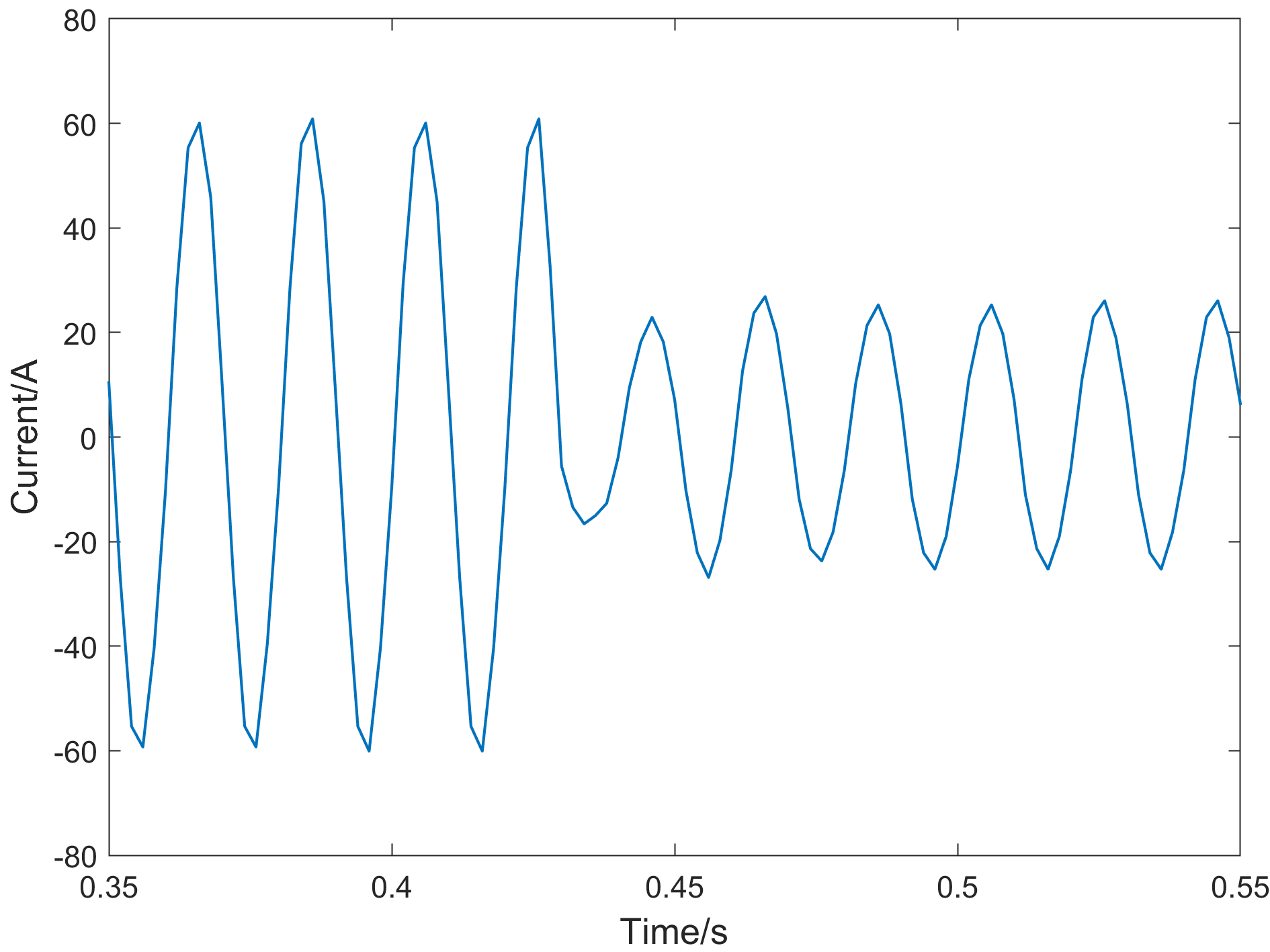}
			\end{minipage}
			\begin{minipage}{0.24\textwidth}
				\centering
				\includegraphics[scale=0.23]{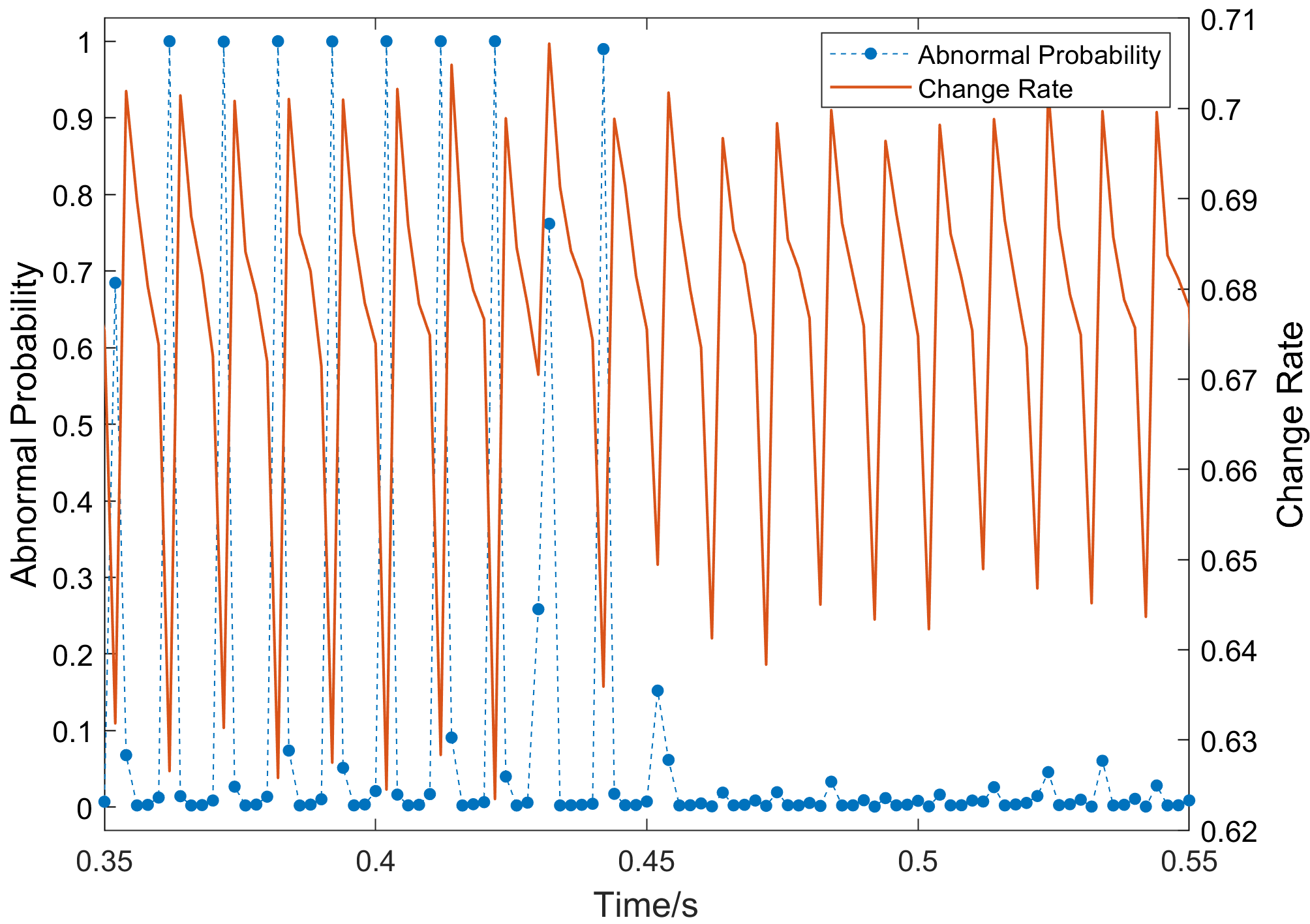}
			\end{minipage}
		}\\
		\subfigure{
			\begin{minipage}{0.24\textwidth}
				\centering
				\includegraphics[scale=0.23]{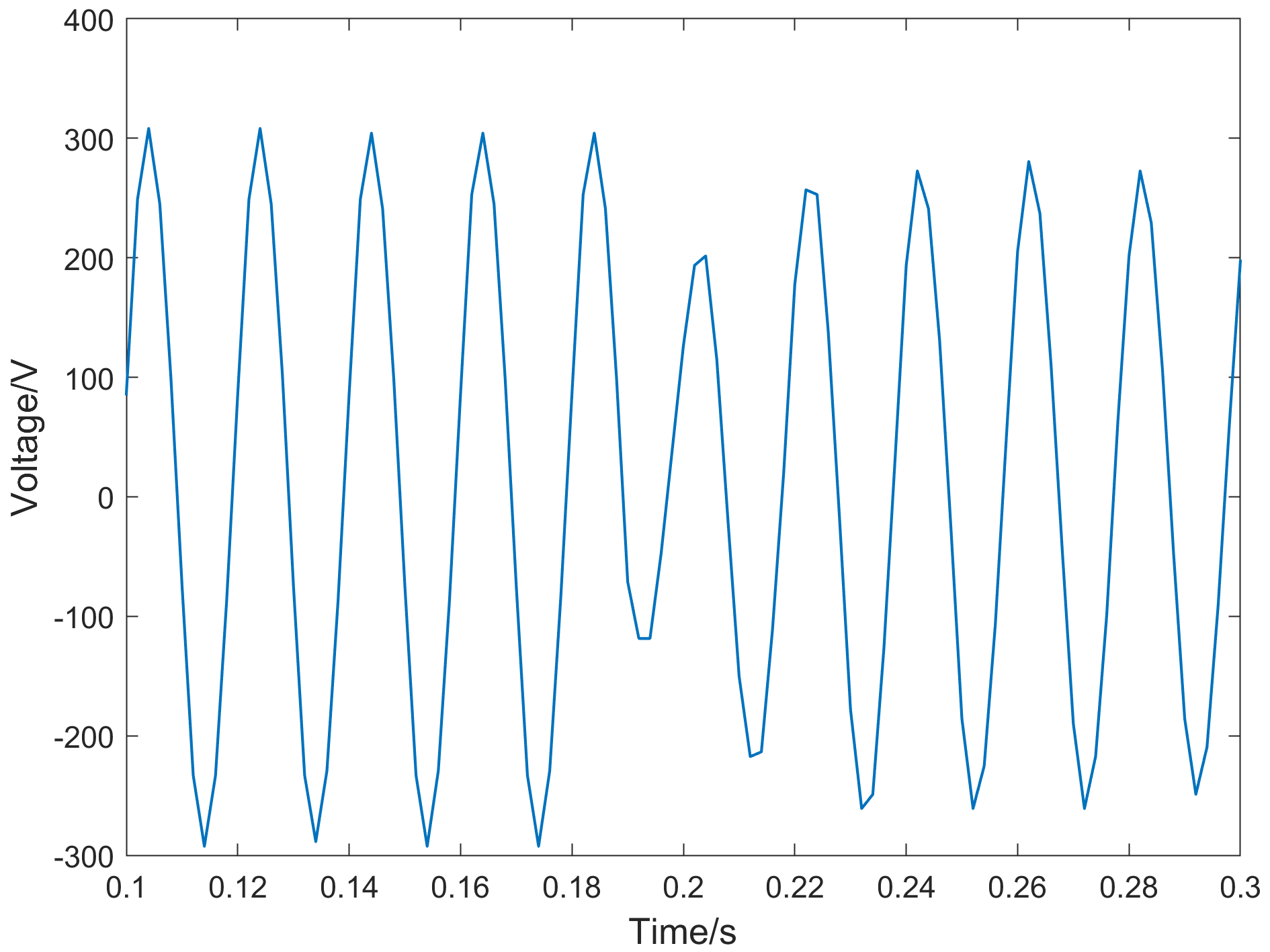}
			\end{minipage}
			\begin{minipage}{0.24\textwidth}
				\centering
				\includegraphics[scale=0.23]{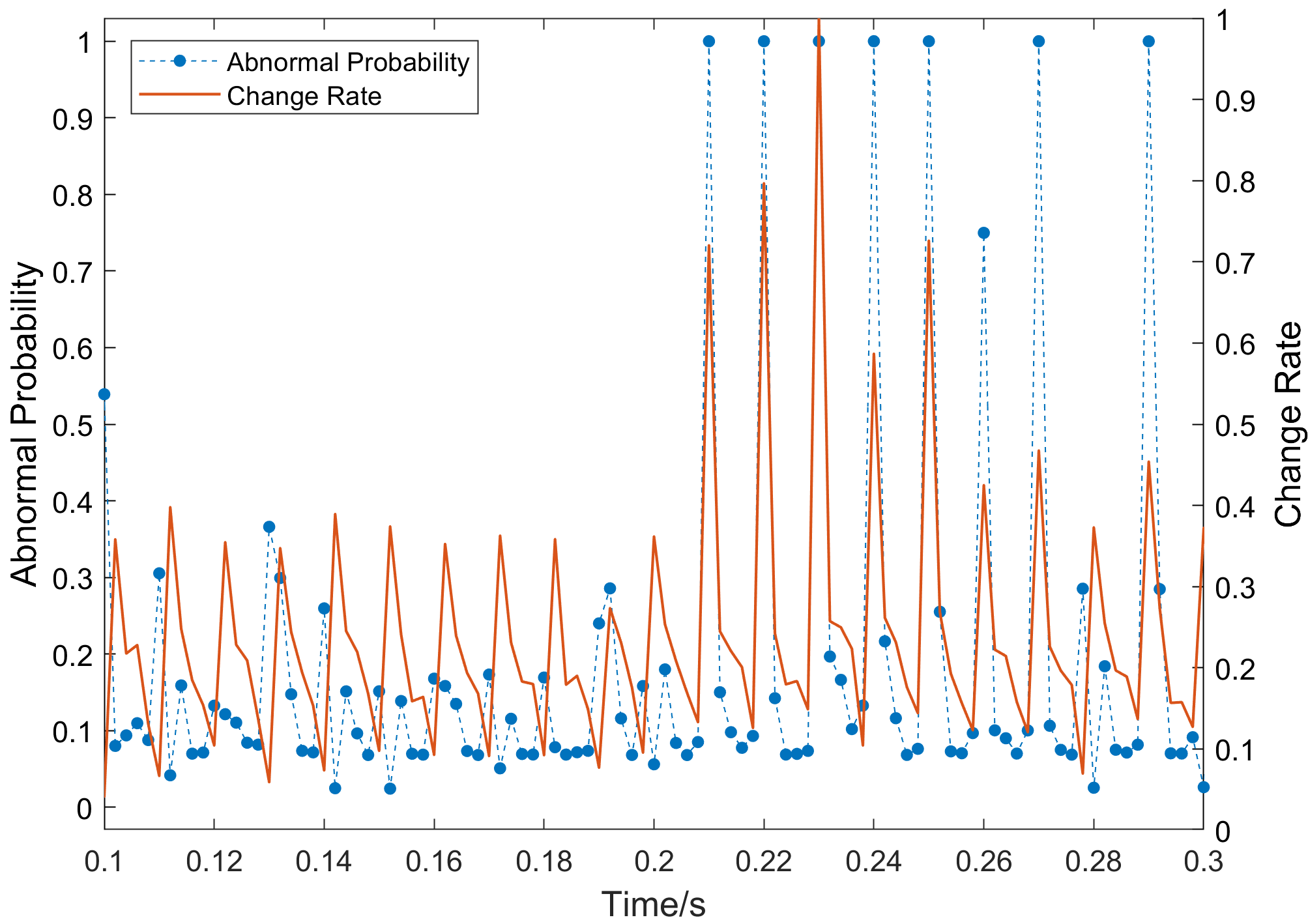}
			\end{minipage}
			\begin{minipage}{0.24\textwidth}
				\centering
				\includegraphics[scale=0.23]{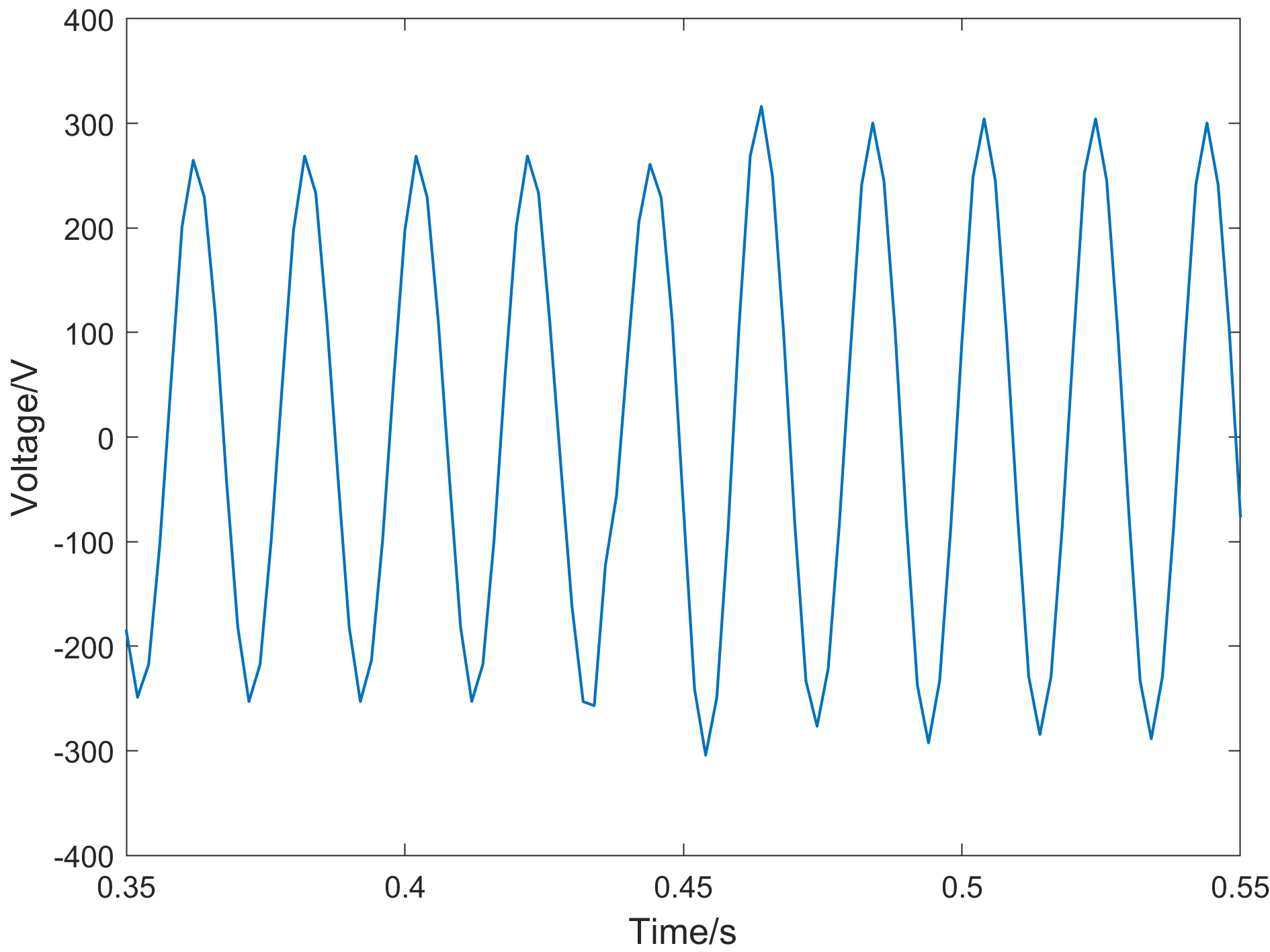}
			\end{minipage}
			\begin{minipage}{0.24\textwidth}
				\centering
				\includegraphics[scale=0.23]{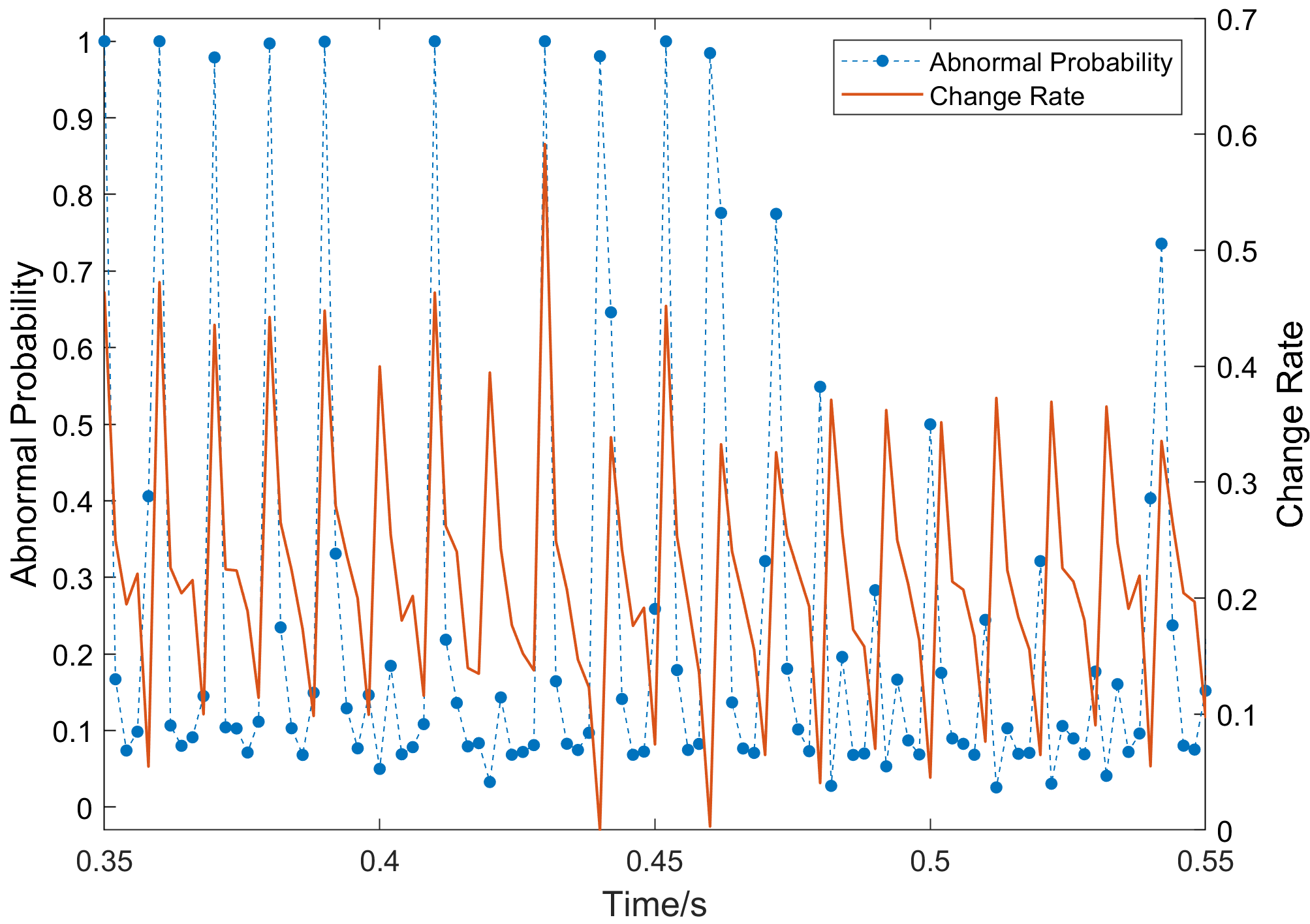}
			\end{minipage}
		}
		\caption{EMODM Results Applied in Local Time 0.1-0.3\,s and 0.35-0.55\,s. Left: short-circuit failure occurrence segment; Right: short-circuit failure end segment; Top: current signal output $i_{oa}$; Bottom: voltage signal output $u_{ab}$.\label{exper3}}
		\subfigure{
			\includegraphics[width=0.95\textwidth]{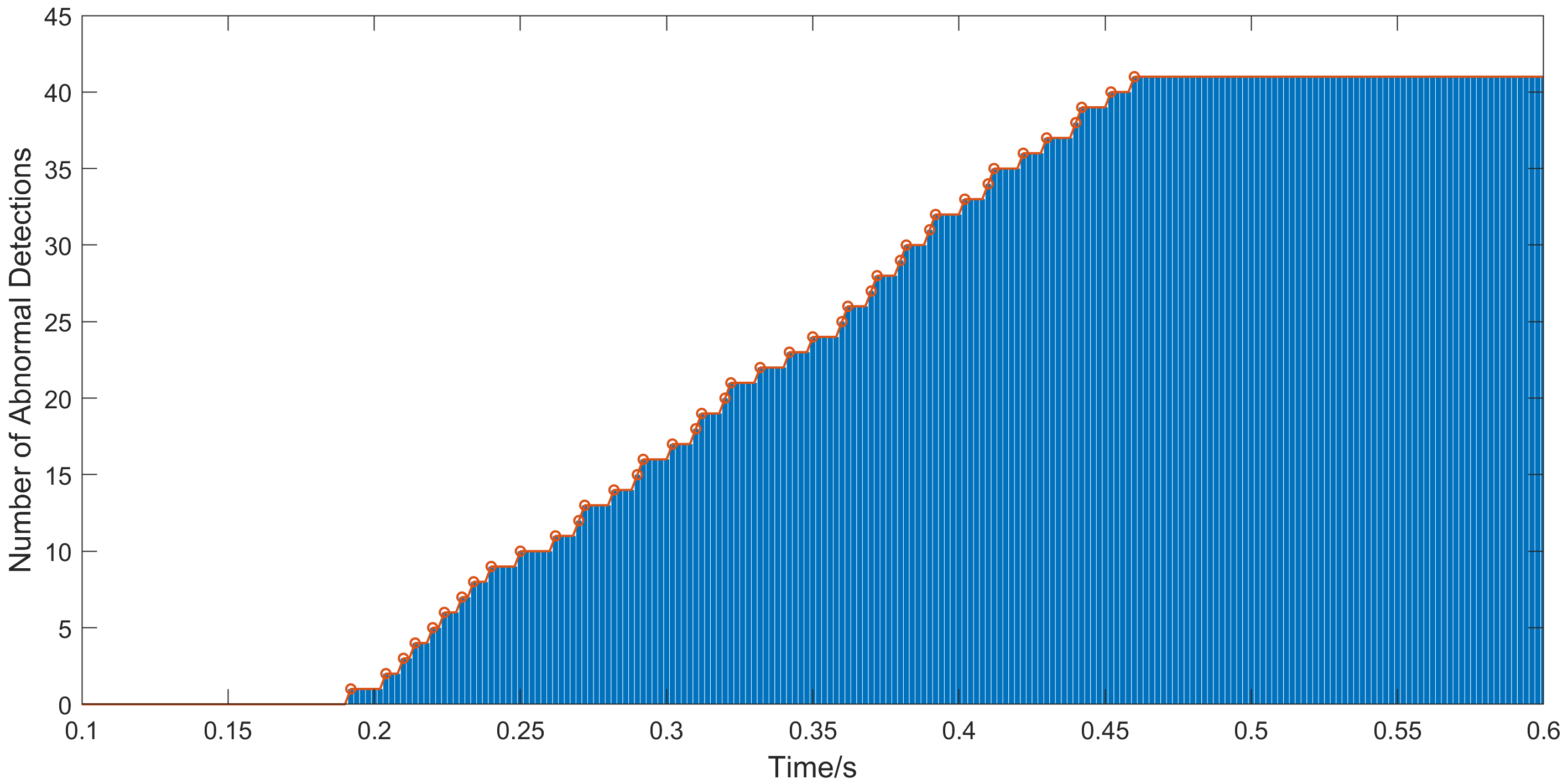}
		}
		\caption{\label{exper4} Online Detection Result: abnormal Time Points Capture Number of Complex Circuit System In Time Segment 0.1-0.6\,s.}
	\end{figure}

	A sinusoidal signal can be used as an output to this complex circuit system. Then we can observe the waveform and amplitude change from front to back or back to front by using an oscilloscope. Here, we use the relative change rate of the time series data from the circuit output and obtain the fault diagnosis results based on the EMODM. Figure \ref{exper2} left-hand side demonstrates the current output from the complex circuit which occurs by a designed short circuit. The global time of the experiment is 1\,s, and the period for abnormal detection is divided into 0.002\,s. The statistical results of the EMODM are shown in table \ref{tab2} and the visualization is shown in figure \ref{exper2} right-hand side. Note that the red orbit represents the change rate of the circuit output with noise and the blue orbit represents the probabilistic output results of the EMODM. In the comparison of the above two figures, it can be found that the EMODM has continuously detected abnormal time points in the current anomaly segment caused by a short circuit. Our method detects an abnormal time segment of 0.190-0.442\,s for current $i_{oa}$ and 0.208-0.458\,s for voltage $u_{ab}$.
	
	To provide the local detection performance of the EMODM, we have made a magnified view of the fault occurrence and termination segment in figure \ref{exper3}. It is necessary to show that our method enables the computer program to constantly detect faults within the time segment of the abnormal circuit system output. The method would not continue to erroneously detect new abnormal time points after the circuit system reenters the correct working state. Furthermore, figure \ref{exper4} presents the online abnormal detection result by the EMODM in a time segment 0.1-0.6\,s. We capture a total of 41 abnormal time points in global time by the current and voltage signal output with a high abnormal detection rate and low false detection rate. Thus it can quickly send an alarm signal after accurately capturing an abnormal time point. Finally, by examining the signal outputs of current $i_{oa}$ and voltage $u_{ab}$, we deduce that the failure probability of this circuit system is 2.09-2.25$\,\%$ due to EMODM results. Overall the feasibility and effectiveness of the EMODM are verified further in the complex circuit system of a real experimental scenario.
	
	\begin{table}[t]
		\centering
		\begin{tabular}{c|cccc}
			\hline
			\hline
			Output & Pattern &Proportion&Mean&Variance\\
			\hline
			Current $i_{oa}$ & 1:\,Correct&97.91\,\%&-0.4822& 1.7299\\
			& 2:\,Abnormal& 2.09\,\%&6.3086&232.0743 \\
			\hline
			Voltage $u_{ab}$ & 1:\,Correct & 97.75\,\% & -0.0552 & 1.7034\\
			& 2:\,Abnormal& 2.25\,\% & 4.6177 & 9.3288 \\
			\hline
			\hline
			Output &Abnormal Detection &  Occurrence & Termination &  Abnormal  \\
			&Time Points &Time & Time & Probability \\
			\hline
			Current $i_{oa}$ & 22 & 0.190s & 0.442s & 2.09\,\% \\
			\hline
			Voltage $u_{ab}$ & 19 & 0.208s & 0.458s & 2.25\,\%  \\
			\hline
			\hline
		\end{tabular}
		\caption{\label{tab2}EMODM Results for Current and Voltage Outputs of Complex Circuit System.}
	\end{table}

	\subsection{U.S. Insured Unemployment Dataset} 
	
	\begin{figure}[t]
		\centering
		\subfigure{
			\begin{minipage}{0.32\textwidth}
				\centering
				\includegraphics[scale=0.3]{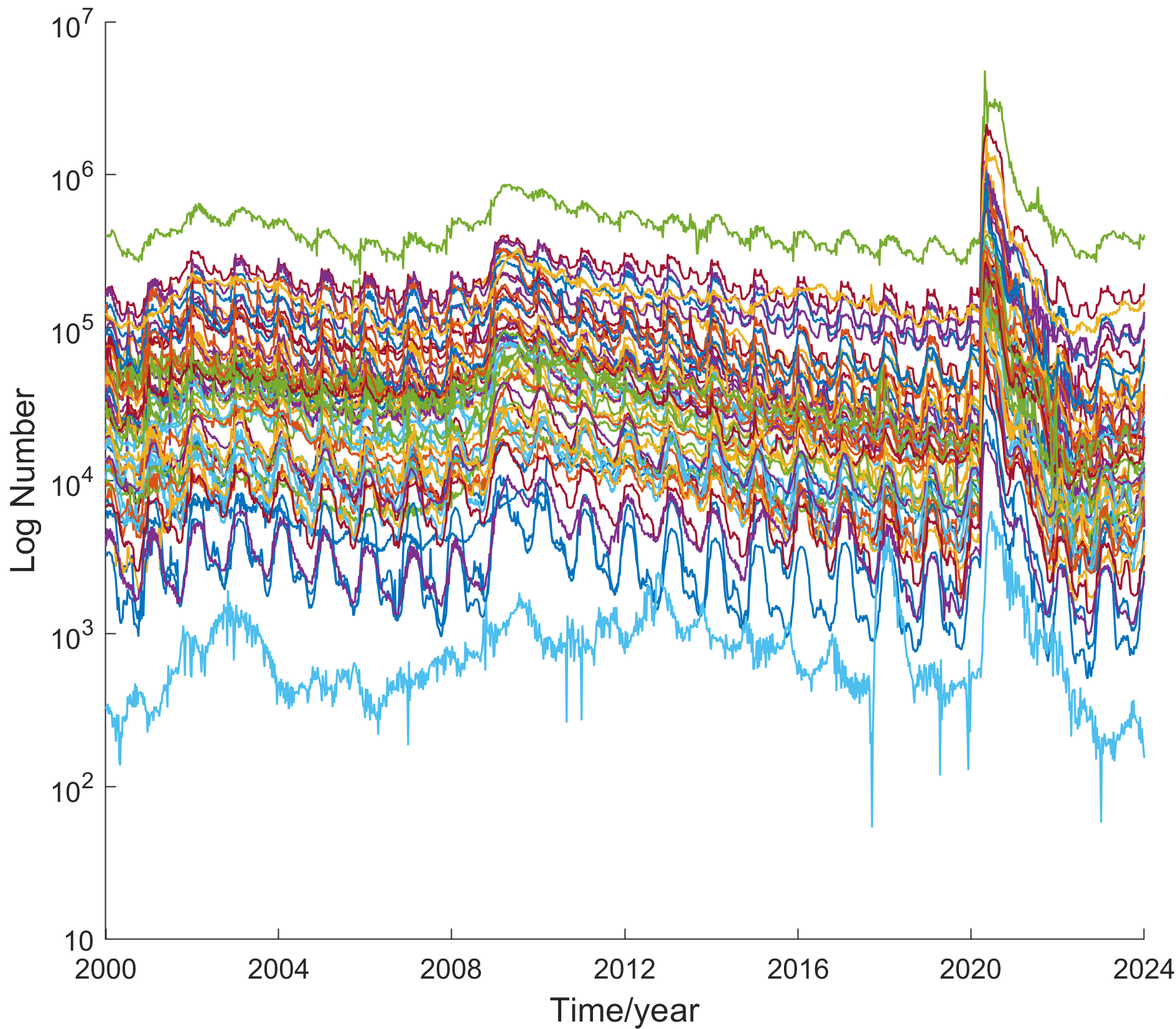}
			\end{minipage}
			\begin{minipage}{0.32\textwidth}
				\centering
				\includegraphics[scale=0.3]{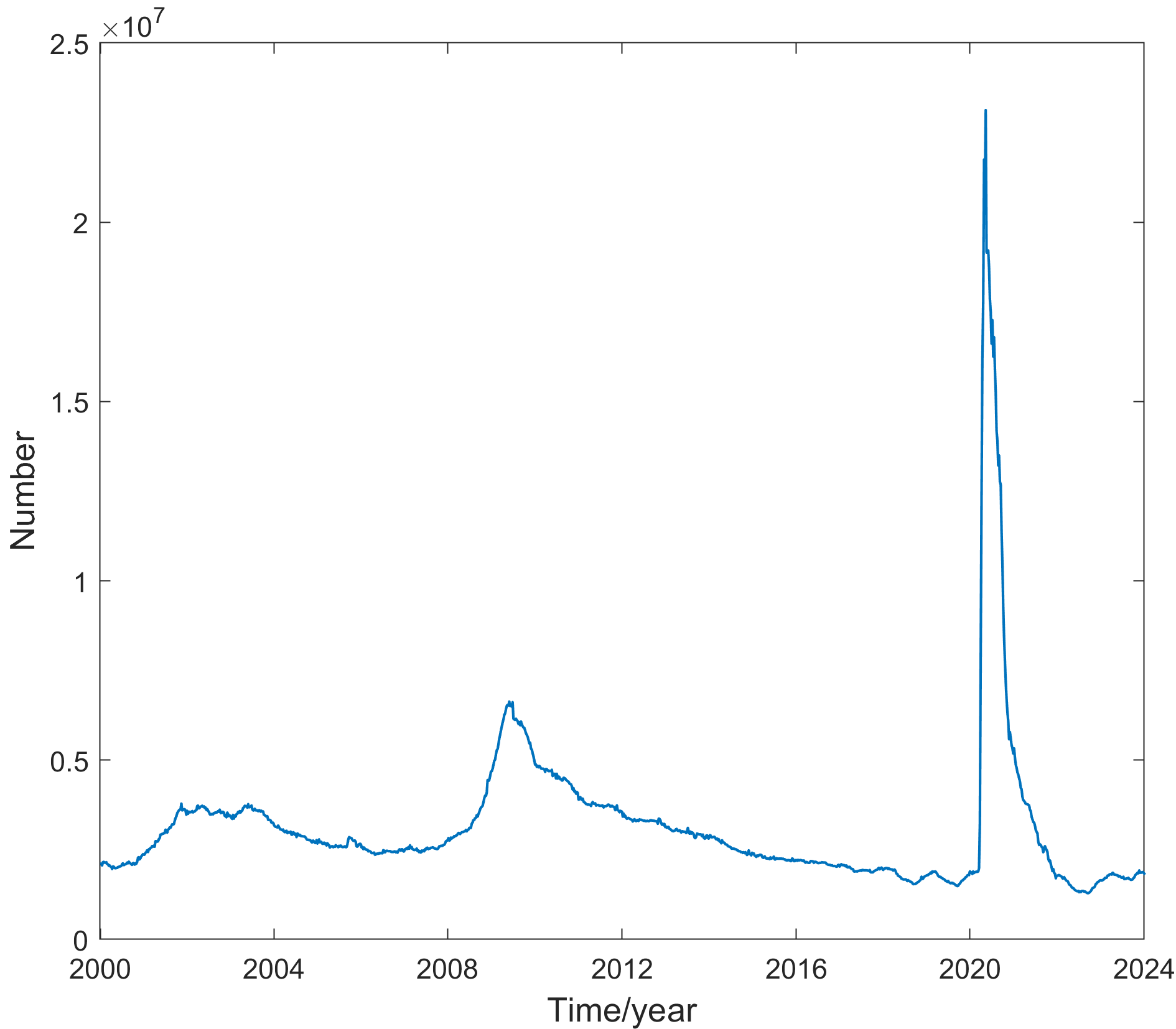}
			\end{minipage}
			\begin{minipage}{0.32\textwidth}
				\centering
				\includegraphics[scale=0.3]{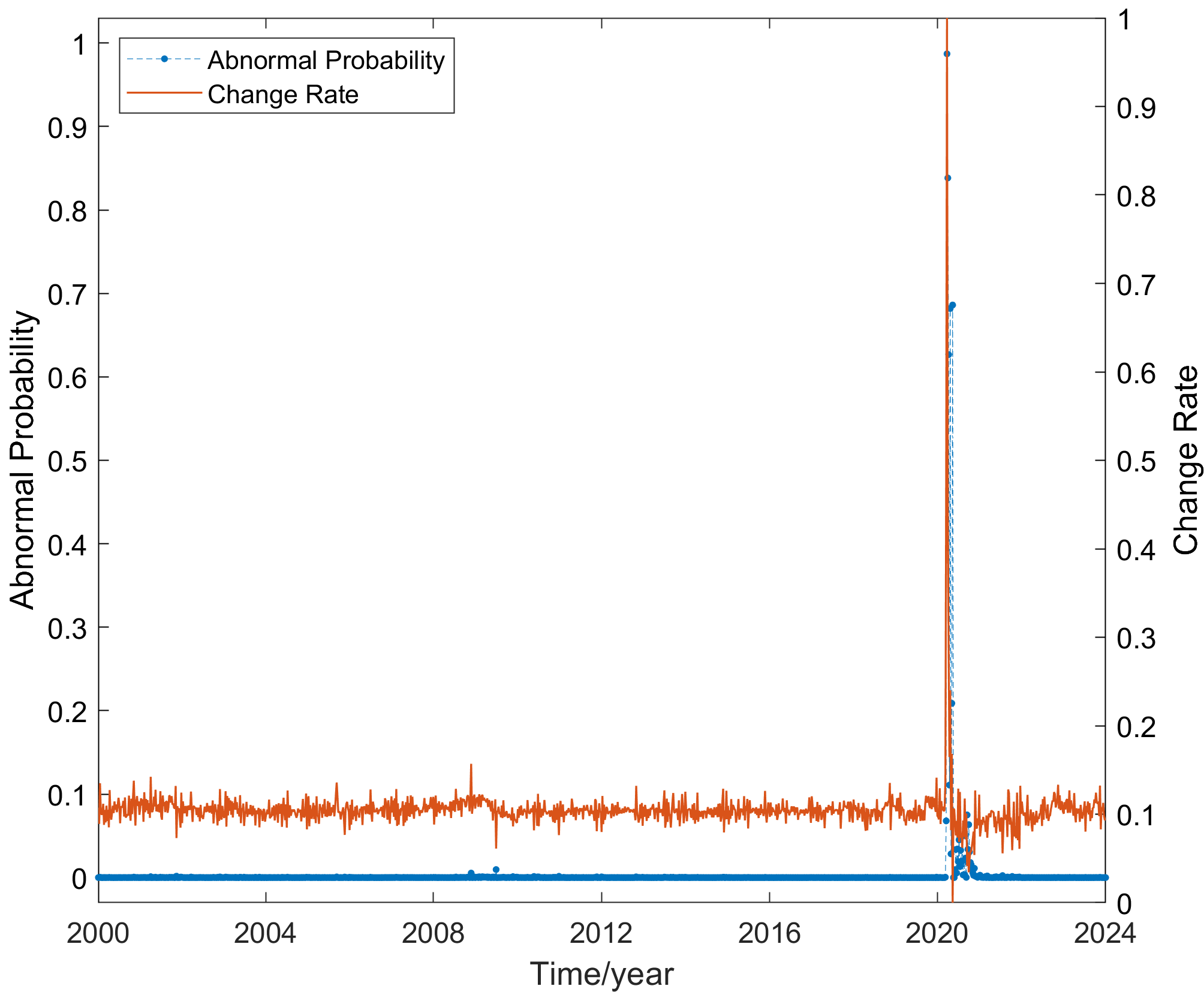}
			\end{minipage}
		}
		\caption{\label{us1} Left: U.S. Insured Unemployment in 53 regions; Middle: Total U.S. Insured Unemployment Data; Right: EMODM Results for Total Data.}
		
		\subfigure{
			\begin{minipage}{0.32\textwidth}
				\centering
				\includegraphics[scale=0.34]{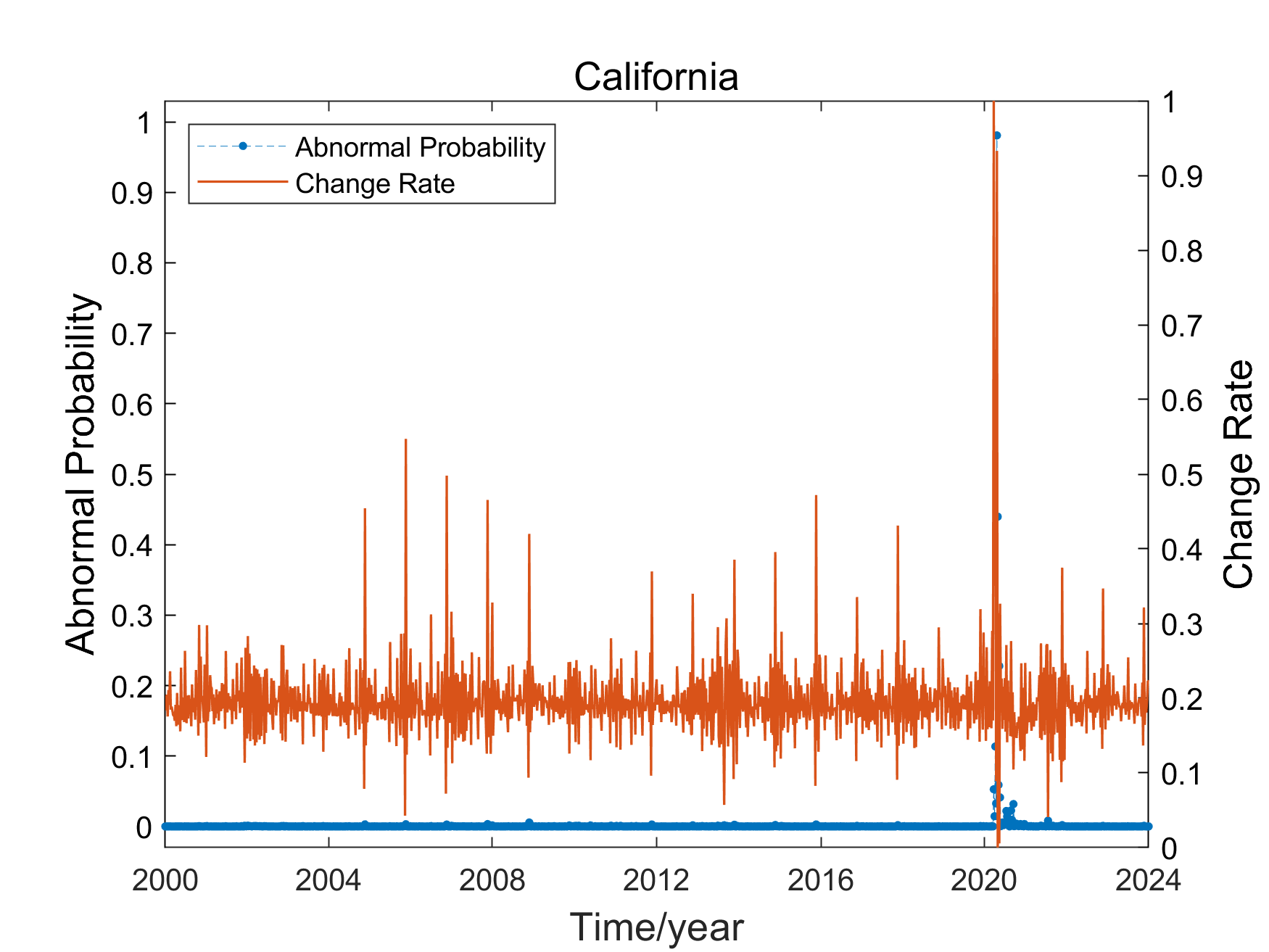}
			\end{minipage}
			\begin{minipage}{0.32\textwidth}
				\centering
				\includegraphics[scale=0.34]{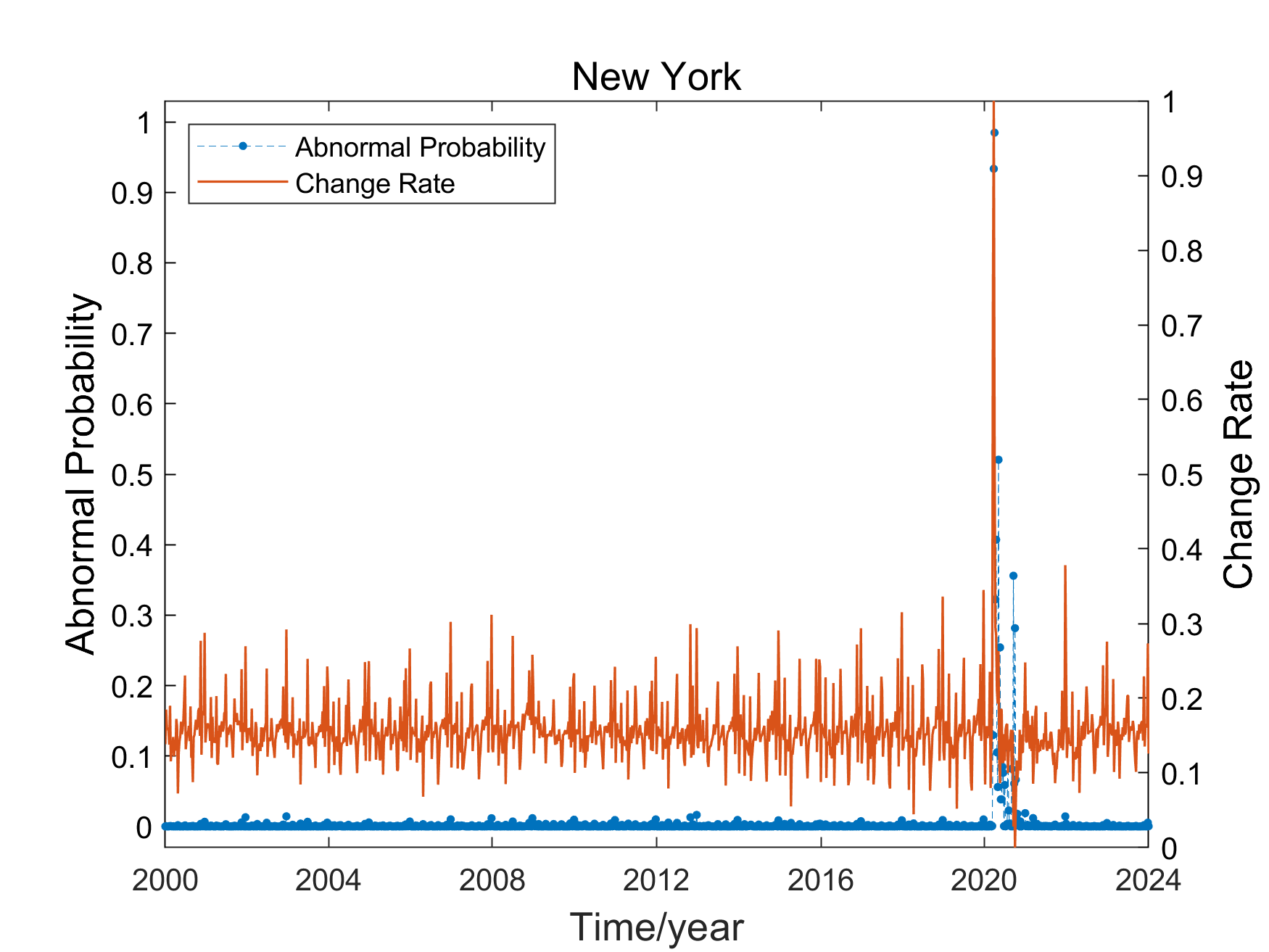}
			\end{minipage}
			\begin{minipage}{0.32\textwidth}
				\centering
				\includegraphics[scale=0.34]{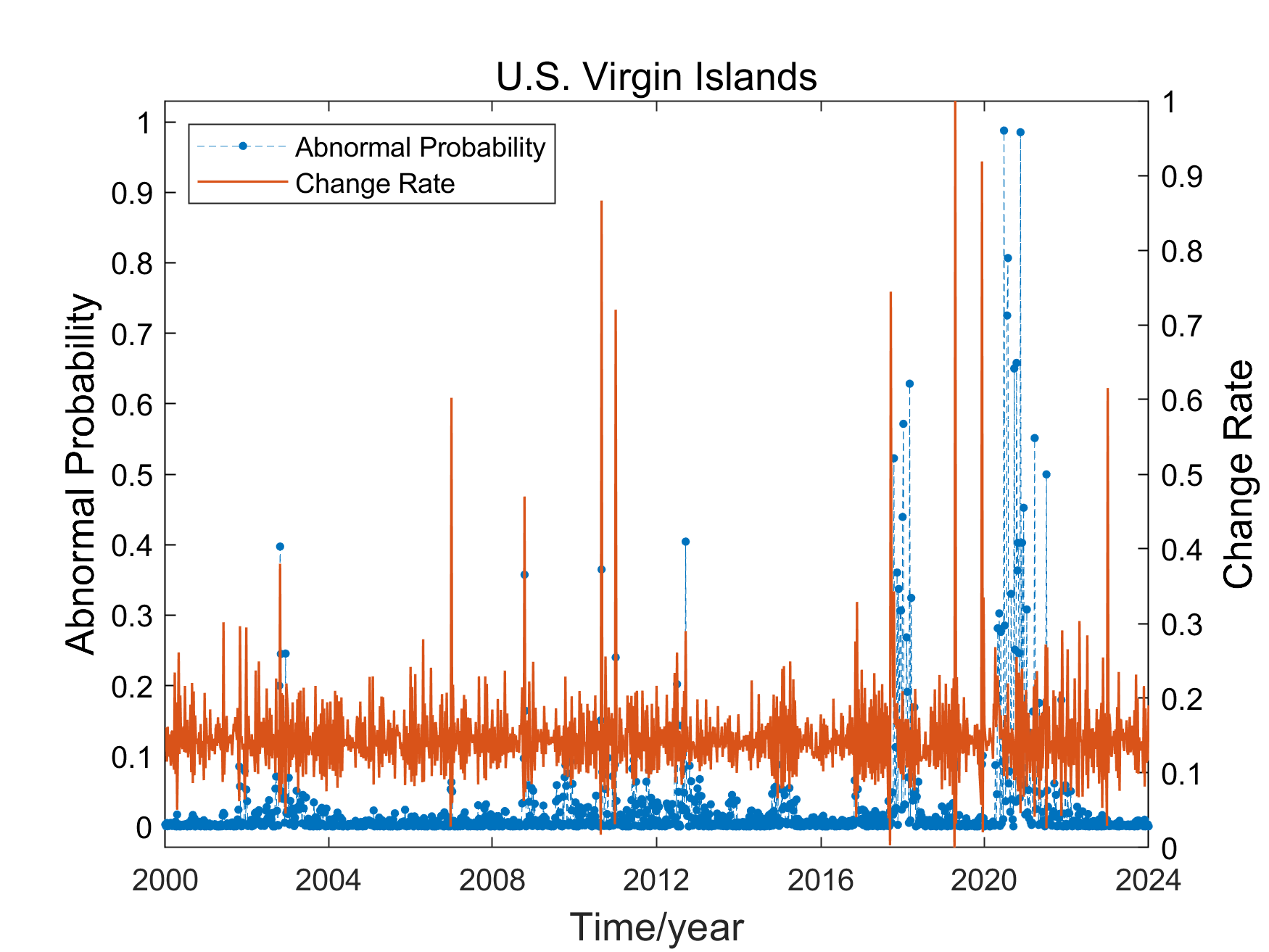}
			\end{minipage}
		}
		\caption{\label{us2}EMODM Results for Insured Unemployment Data in California, New York and U.S. Virgin Islands.}     
	\end{figure}
	
	The economic market, as a complex system, follows certain patterns and undergoes endless cycles of rise and fall over time\cite{niemira1994forecasting}. The insured unemployment data in the U.S. serves as a crucial indicator reflecting the fluctuations for economic markets\cite{fujita2010economic}. During these dynamic processes, abnormal patterns occasionally emerge, as evidenced by the significant anomalies in the data during the COVID-19 pandemic period. The output data of economic markets can usually be modeled by stochastic differential equations
	\begin{equation}
		\begin{cases}
			\mathrm{d}x_{1}(t)=f_{1}(t,x_{1}(t),x_{2}(t),\cdots,x_{n}(t))\mathrm{d}t + \sigma_1 \mathrm{d}B_t,\\
			\mathrm{d}x_{2}(t)=f_{2}(t,x_{1}(t),x_{2}(t),\cdots,x_{n}(t))\mathrm{d}t + \sigma_2 \mathrm{d}B_t,\\
			\cdots \cdots  \\
			\mathrm{d}x_{n}(t)=f_{n}(t,x_{1}(t),x_{2}(t),\cdots,x_{n}(t))\mathrm{d}t + \sigma_n \mathrm{d}B_t,
		\end{cases}
	\end{equation}
	where $\boldsymbol{\sigma} = \{\sigma_1, \sigma_2, \cdots, \sigma_n\}$ is the noise driving this complex system, which is different from the Gaussian observation noise in section \ref{Third_sec}. In this case, the size of $\sigma_i$ is related to the regions in the U.S. The impact of the COVID-19 epidemic will be reflected in a change in functions $f_i$, in other words, this system enters an anomalous pattern. This leads to a significant change in the system output. In this part, we plan to statistically infer the abnormal pattern in the unemployment insurance dataset across the 53 regions in the U.S. from 2000-2024. The dataset is from Federal Reserve Economic Data(FRED) on website \url{https://fred.stlouisfed.org}. We aim to further demonstrate the potential and importance of EMODM in processing real-world data and in abnormal detection in complex systems.
	
	In figure \ref{us1}, the left figure displays the U.S. insured unemployment data across 53 regions from 2000 to 2024. Because the data in different regions varies significantly between regions, and to visualize all data on a single graph, a logarithmic scale with base 10 was used. The figure reveals similar trends across different regions, reflecting the cyclical nature of economic market fluctuations. Notably, there is a discernible anomaly after 2020, corresponding to the widely known black swan event, the COVID-19 pandemic several years ago. This period of anomaly will be a primary focus for further data mining using the EMODM. The middle figure aggregates the data from all regions to represent the nationwide insured unemployment numbers in the U.S. It shows the overall trend of the economic market, which is similar to the characterization of the left figure and highlights a clear anomaly after 2020. This anomaly corresponds to the same period identified in the left plot, indicating a significant impact in the economic market leading to an abnormal pattern due to the COVID-19 pandemic. We will reveal the shock of the pandemic to this complex system in terms of the change rate of system output data with its corresponding abnormal probability. The right figure presents the EMODM results for the U.S. insured unemployment dataset. This figure is a statistical inference for the total data based on the middle figure. It illustrates that around March 21, 2020, the complex system of the U.S. insured unemployment entered an abnormal pattern due to the COVID-19 pandemic.
	
\begin{figure}[p]
		\centering
		\subfigure{
			\begin{minipage}{0.15\textwidth}
				\centering
				\includegraphics[scale=0.16]{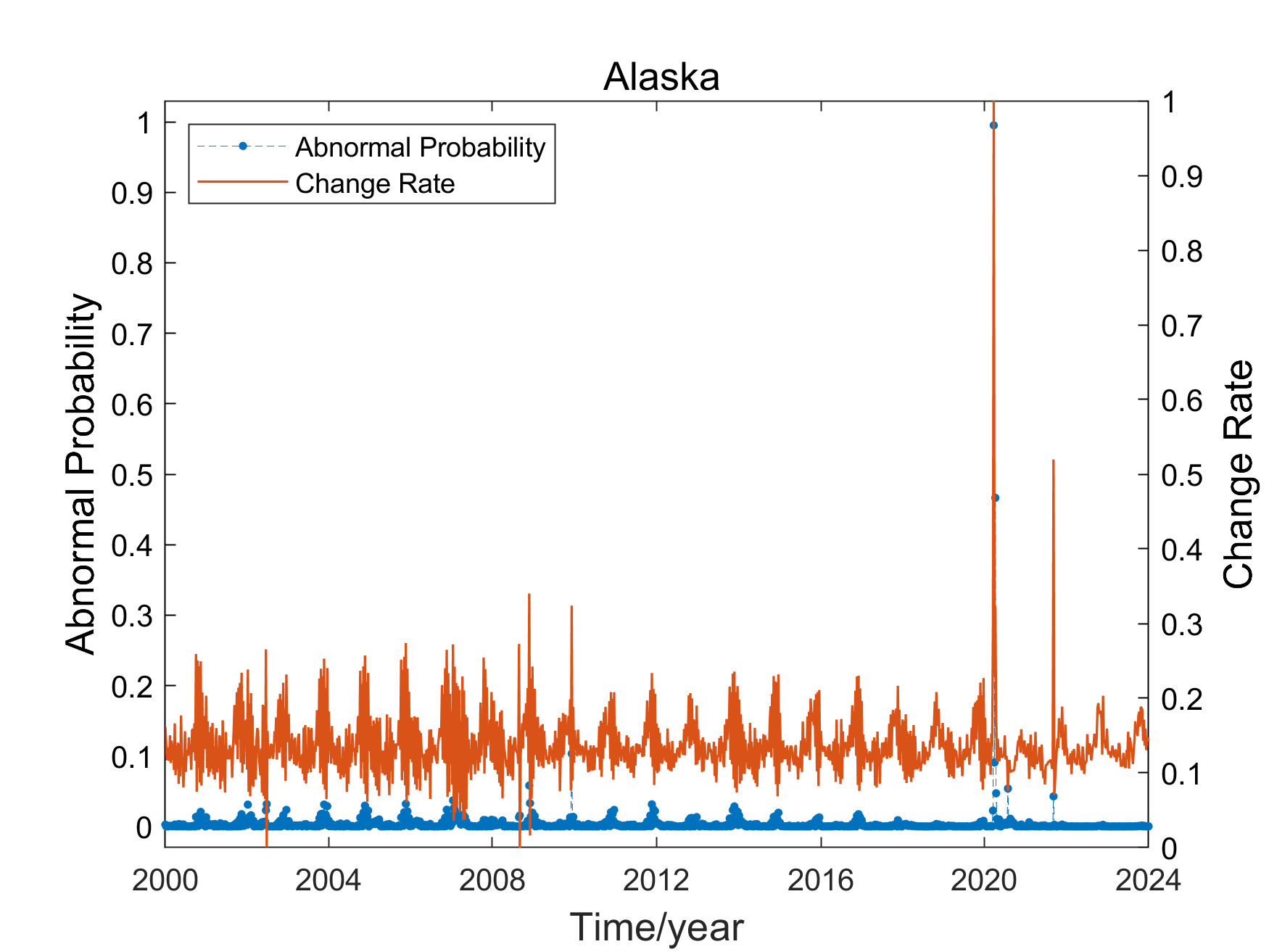}
			\end{minipage}
			\begin{minipage}{0.15\textwidth}
				\centering
				\includegraphics[scale=0.16]{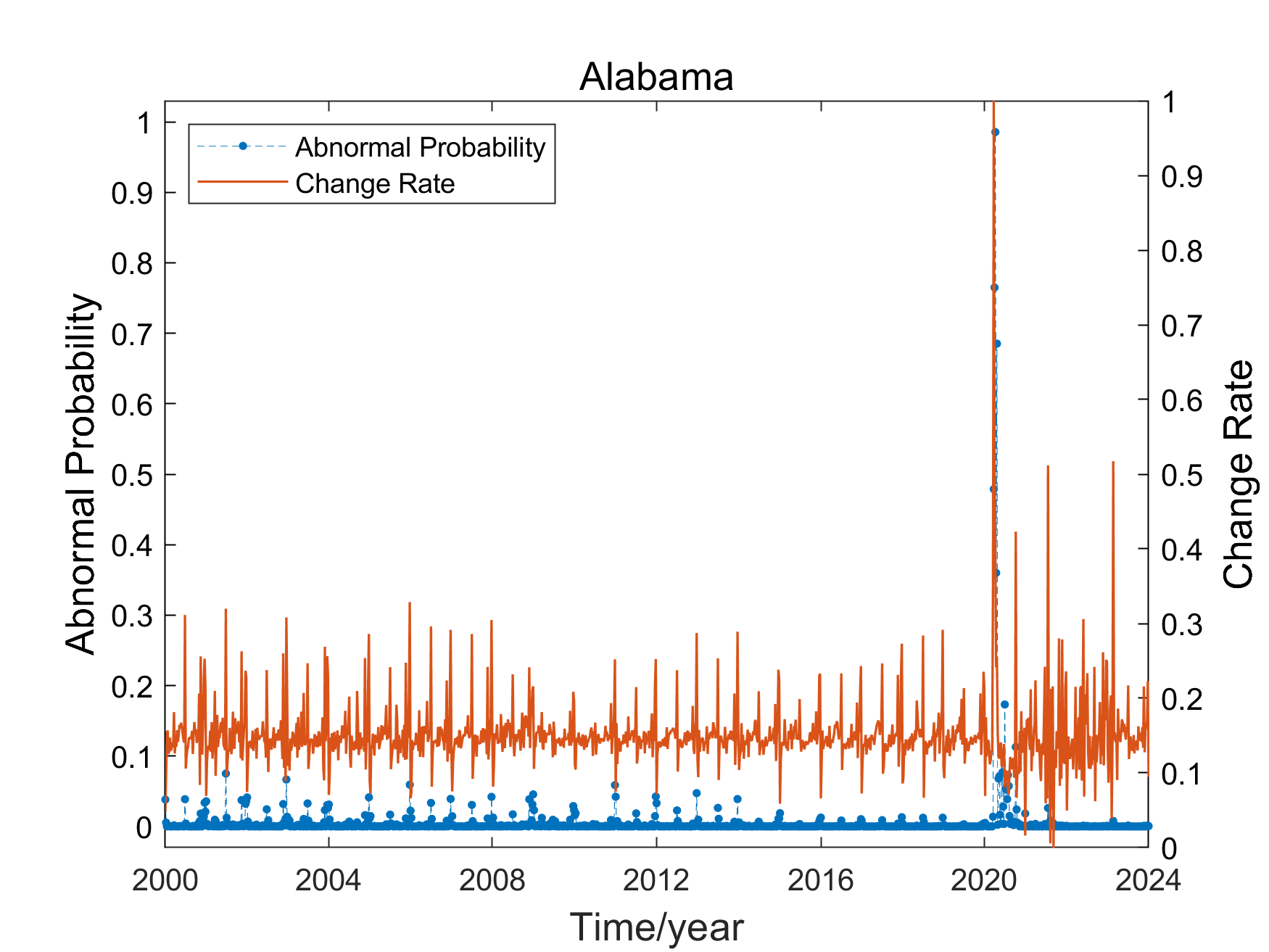}
			\end{minipage}
			\begin{minipage}{0.15\textwidth}
				\centering
				\includegraphics[scale=0.16]{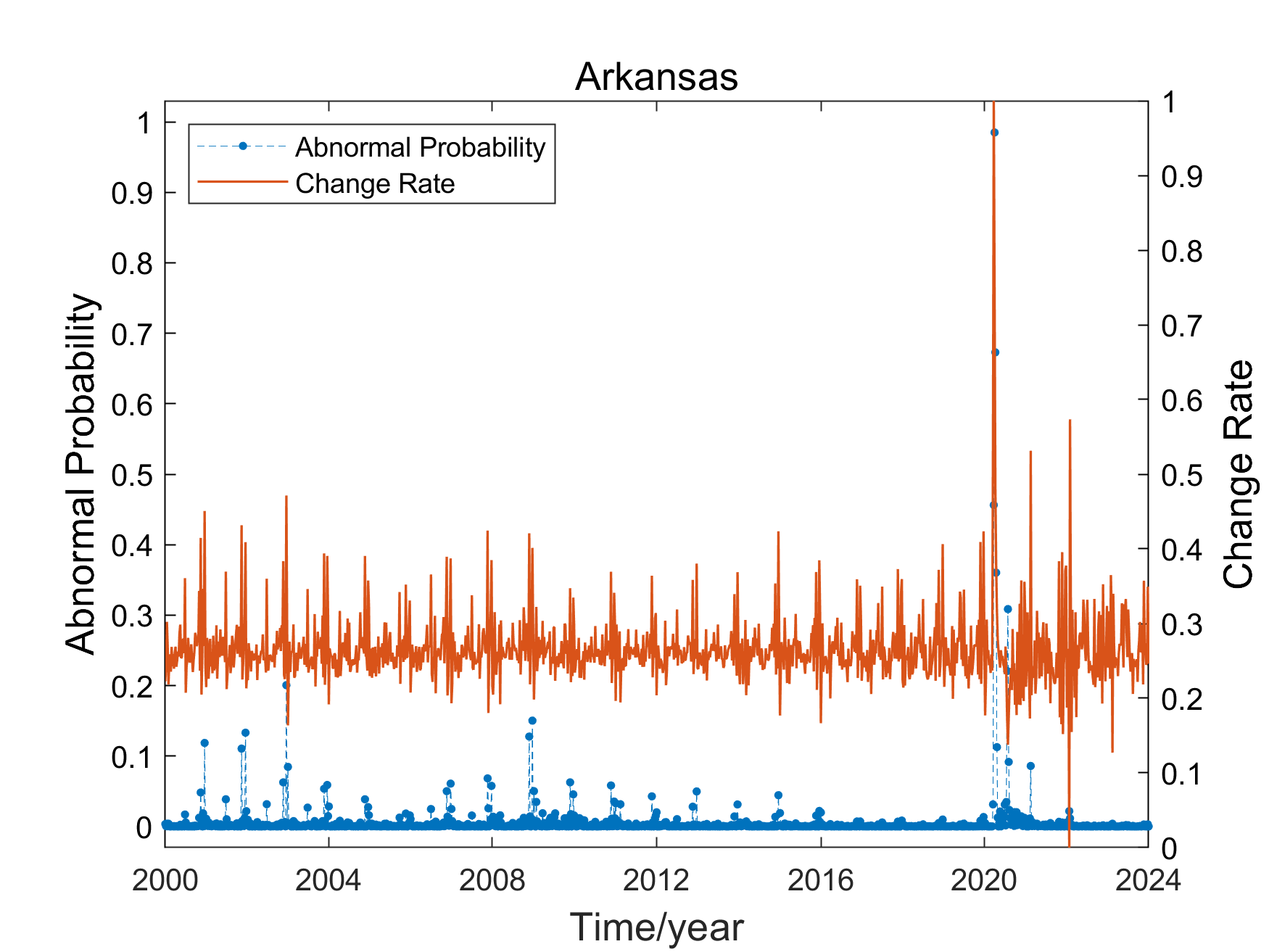}
			\end{minipage}
			\begin{minipage}{0.15\textwidth}
				\centering
				\includegraphics[scale=0.16]{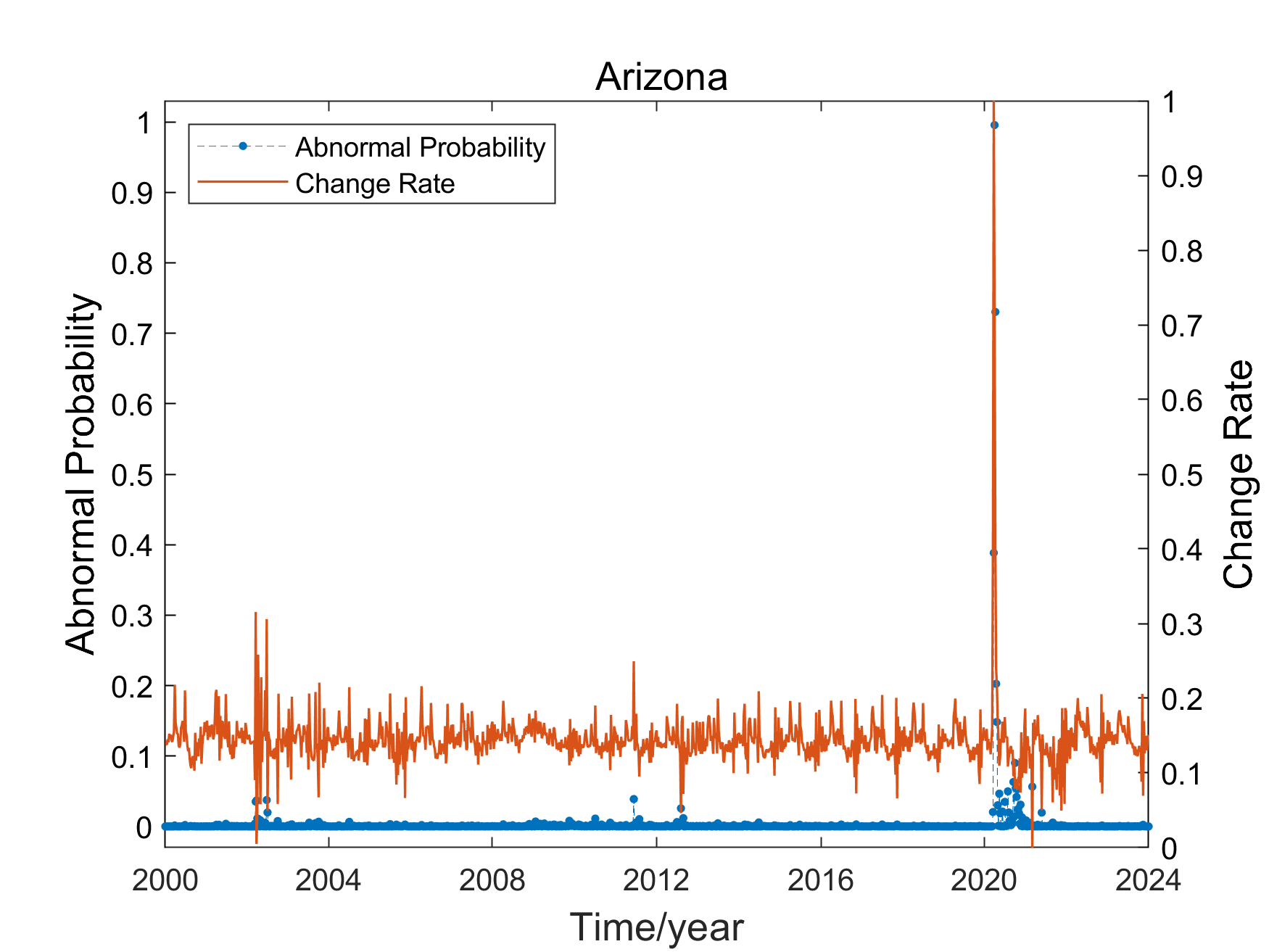}
			\end{minipage}
			\begin{minipage}{0.15\textwidth}
				\centering
				\includegraphics[scale=0.16]{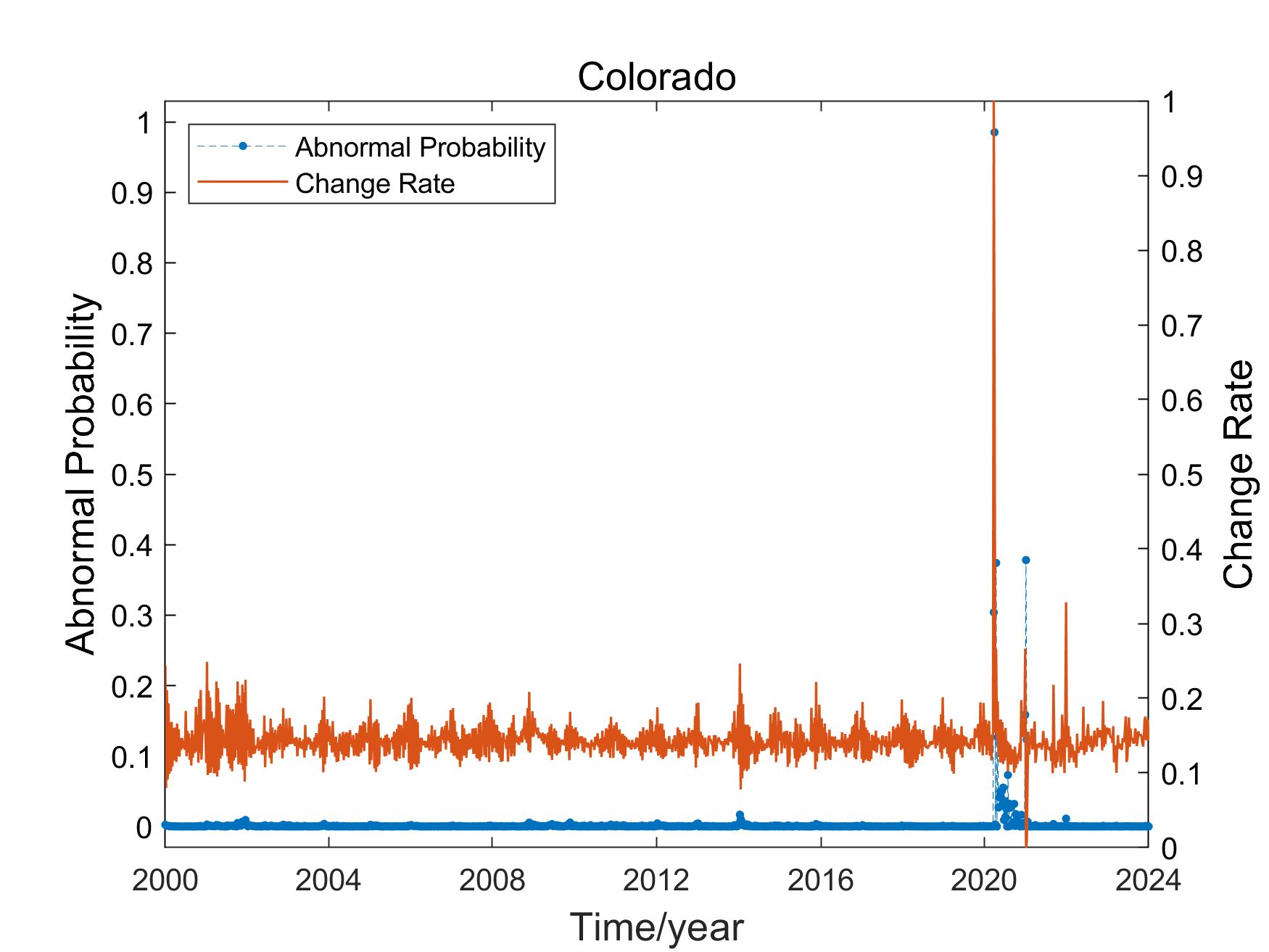}
			\end{minipage}
			
		}\\
		\subfigure{
			\begin{minipage}{0.15\textwidth}
				\centering
				\includegraphics[scale=0.16]{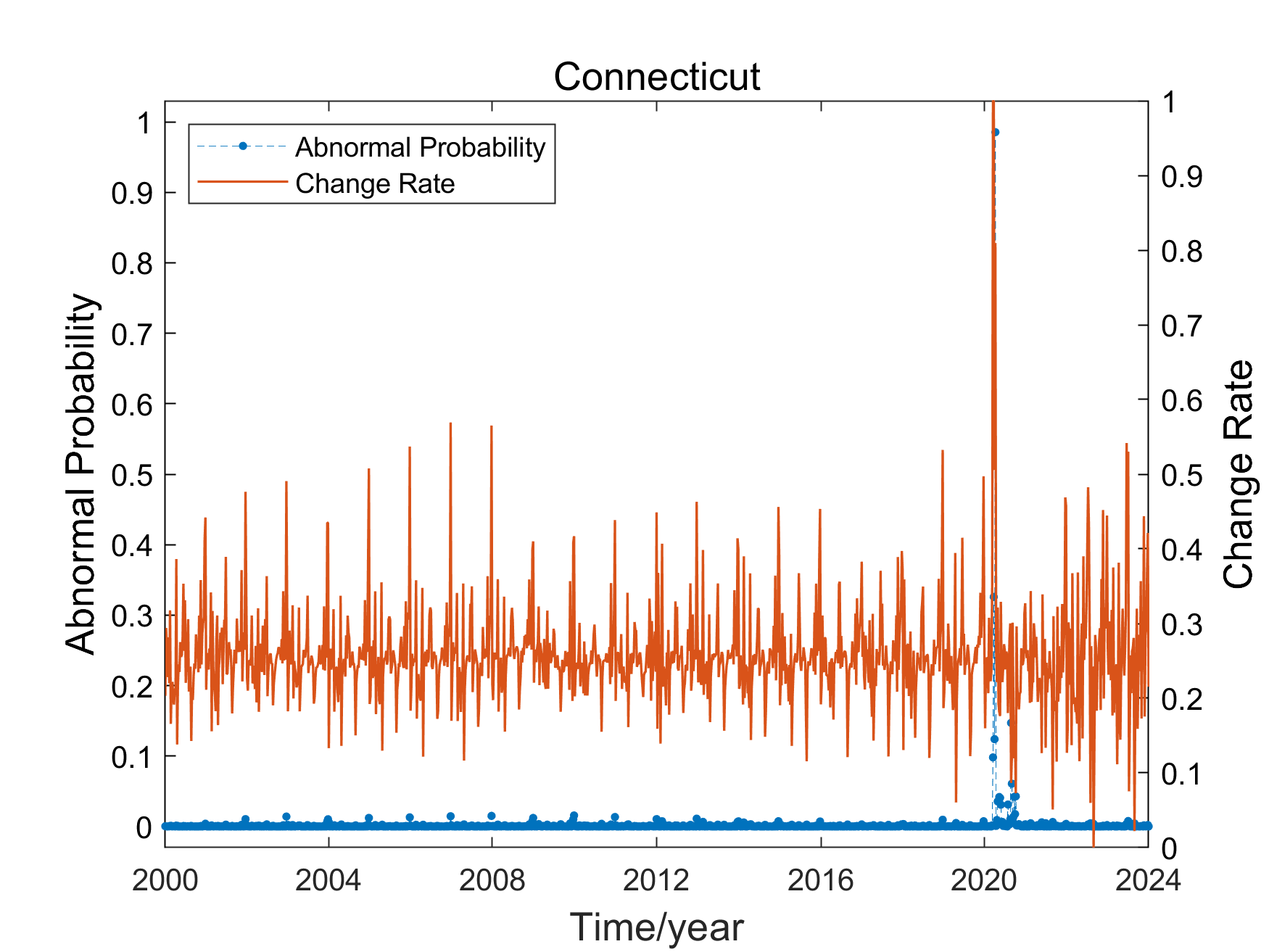}
			\end{minipage}
			\begin{minipage}{0.15\textwidth}
				\centering
				\includegraphics[scale=0.16]{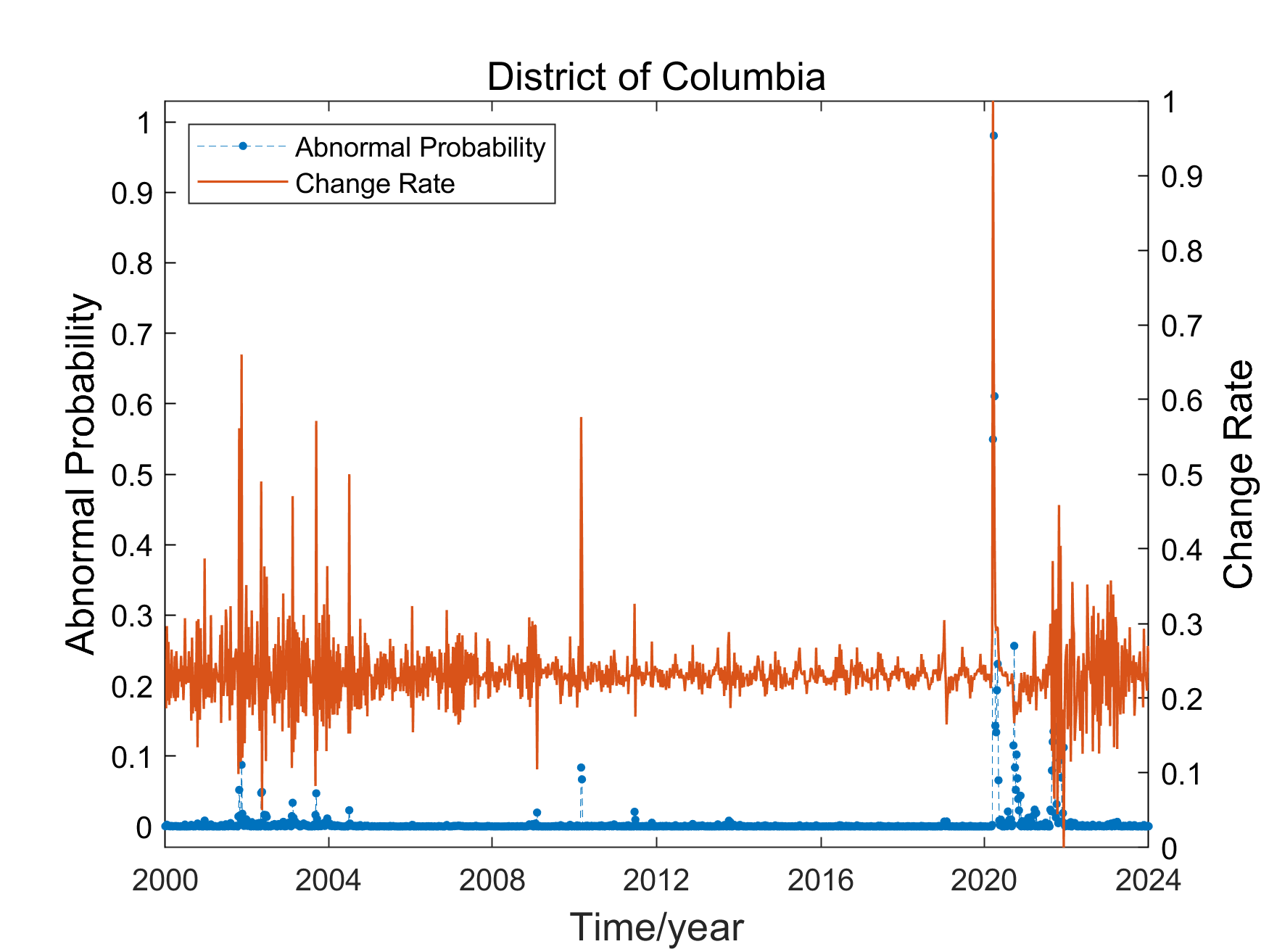}
			\end{minipage}
			\begin{minipage}{0.15\textwidth}
				\centering
				\includegraphics[scale=0.16]{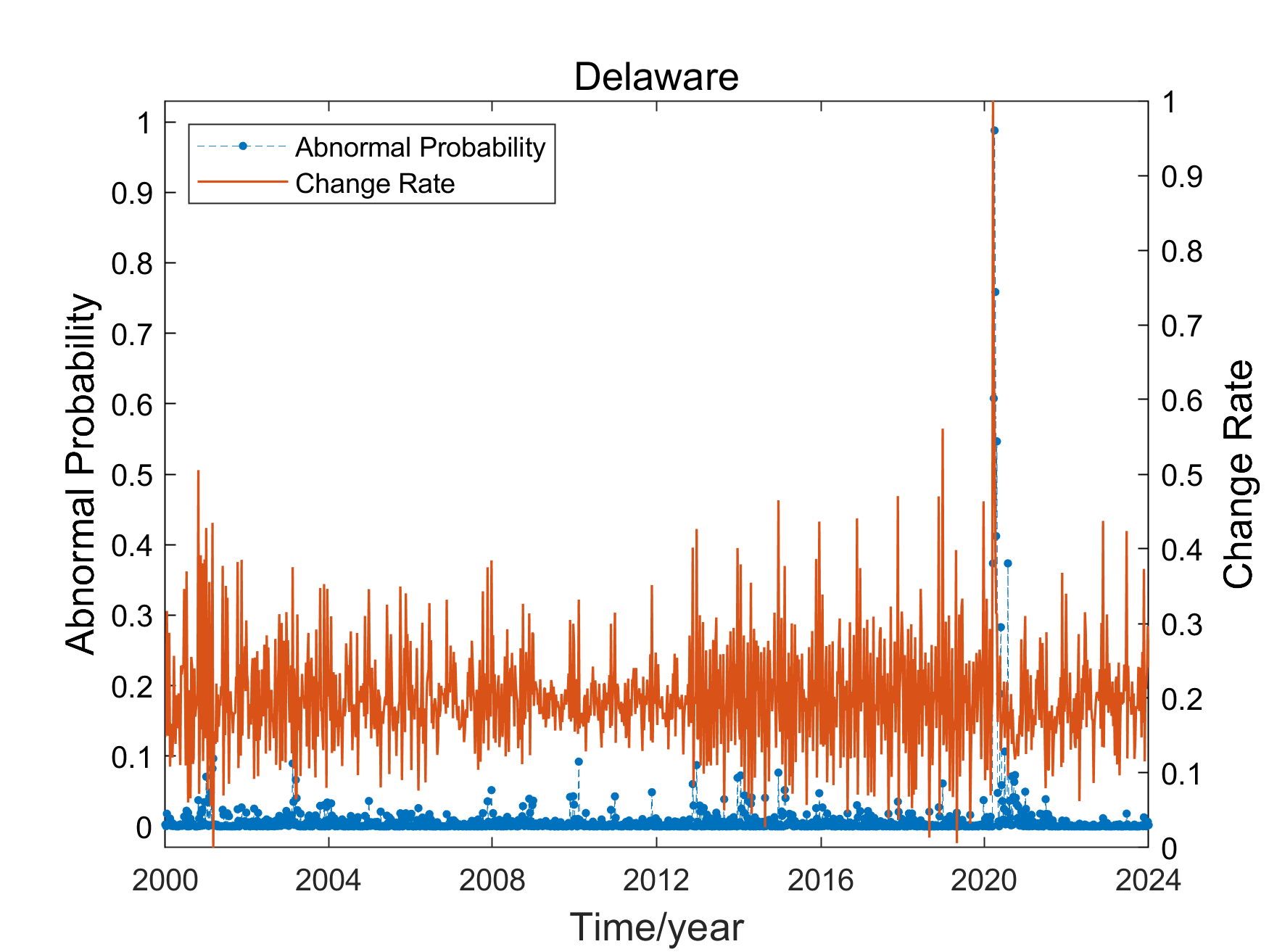}
			\end{minipage}
			\begin{minipage}{0.15\textwidth}
				\centering
				\includegraphics[scale=0.16]{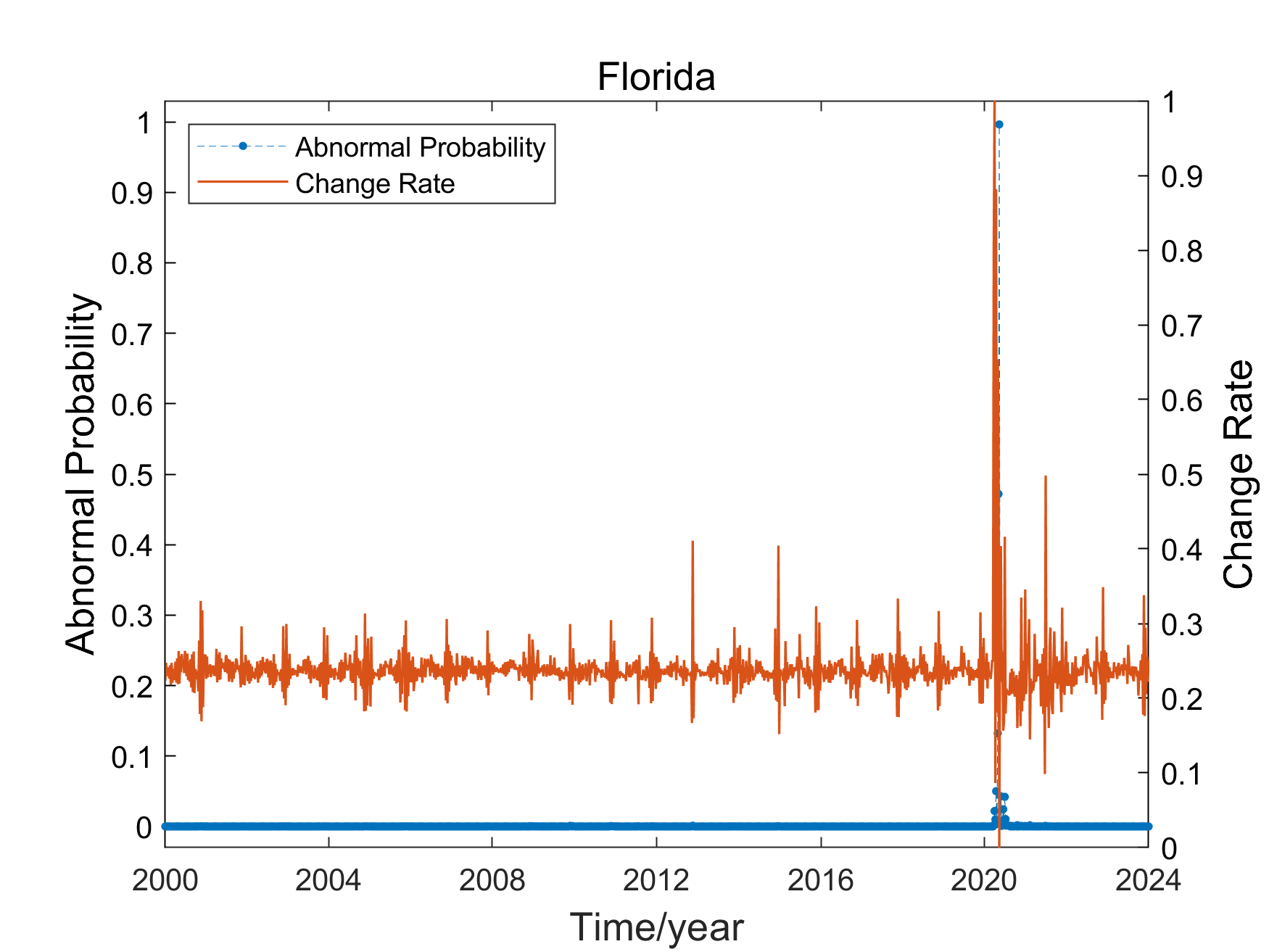}
			\end{minipage}
			\begin{minipage}{0.15\textwidth}
				\centering
				\includegraphics[scale=0.16]{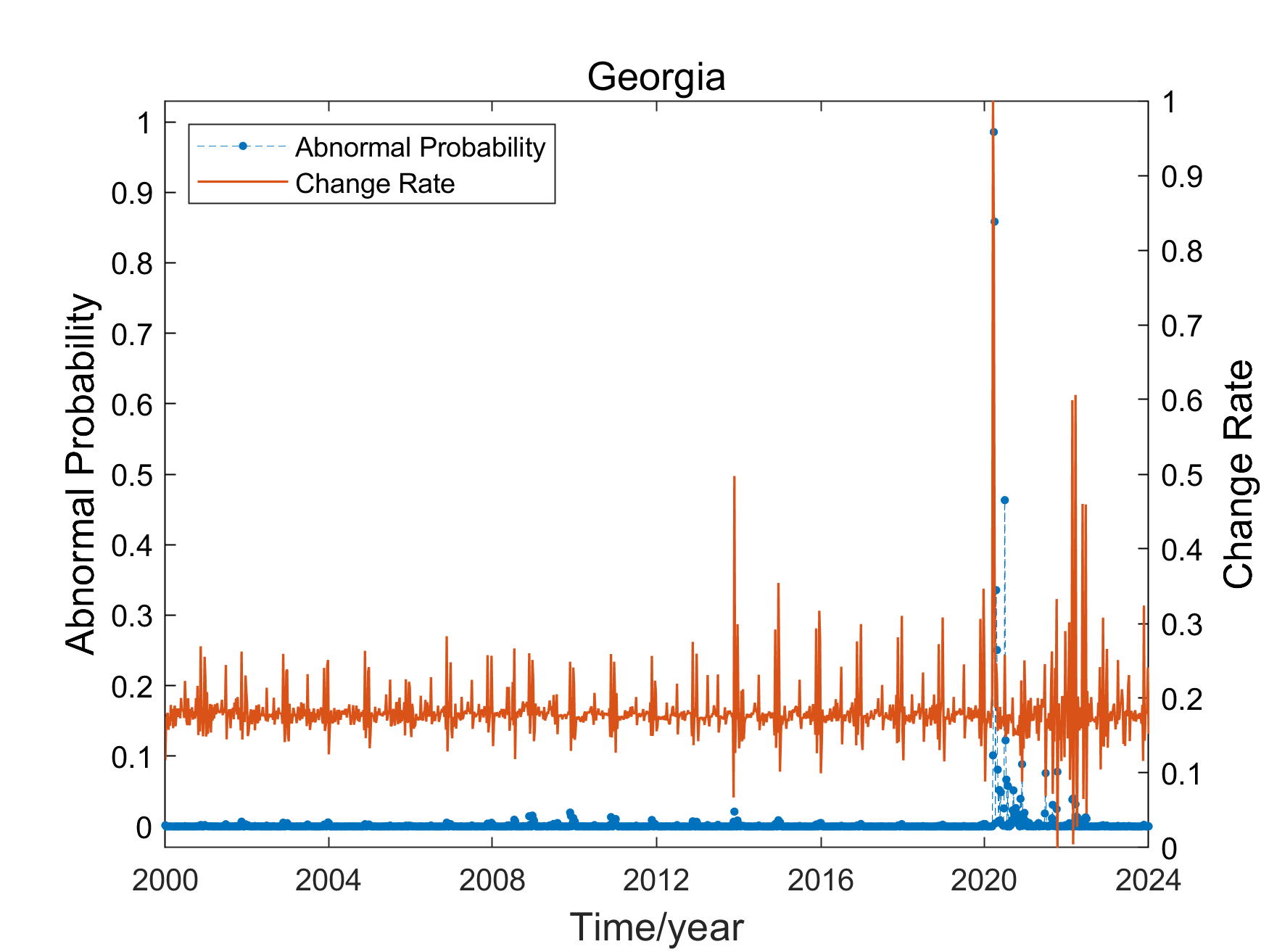}
			\end{minipage}
		}\\
		\subfigure{
			\begin{minipage}{0.15\textwidth}
				\centering
				\includegraphics[scale=0.16]{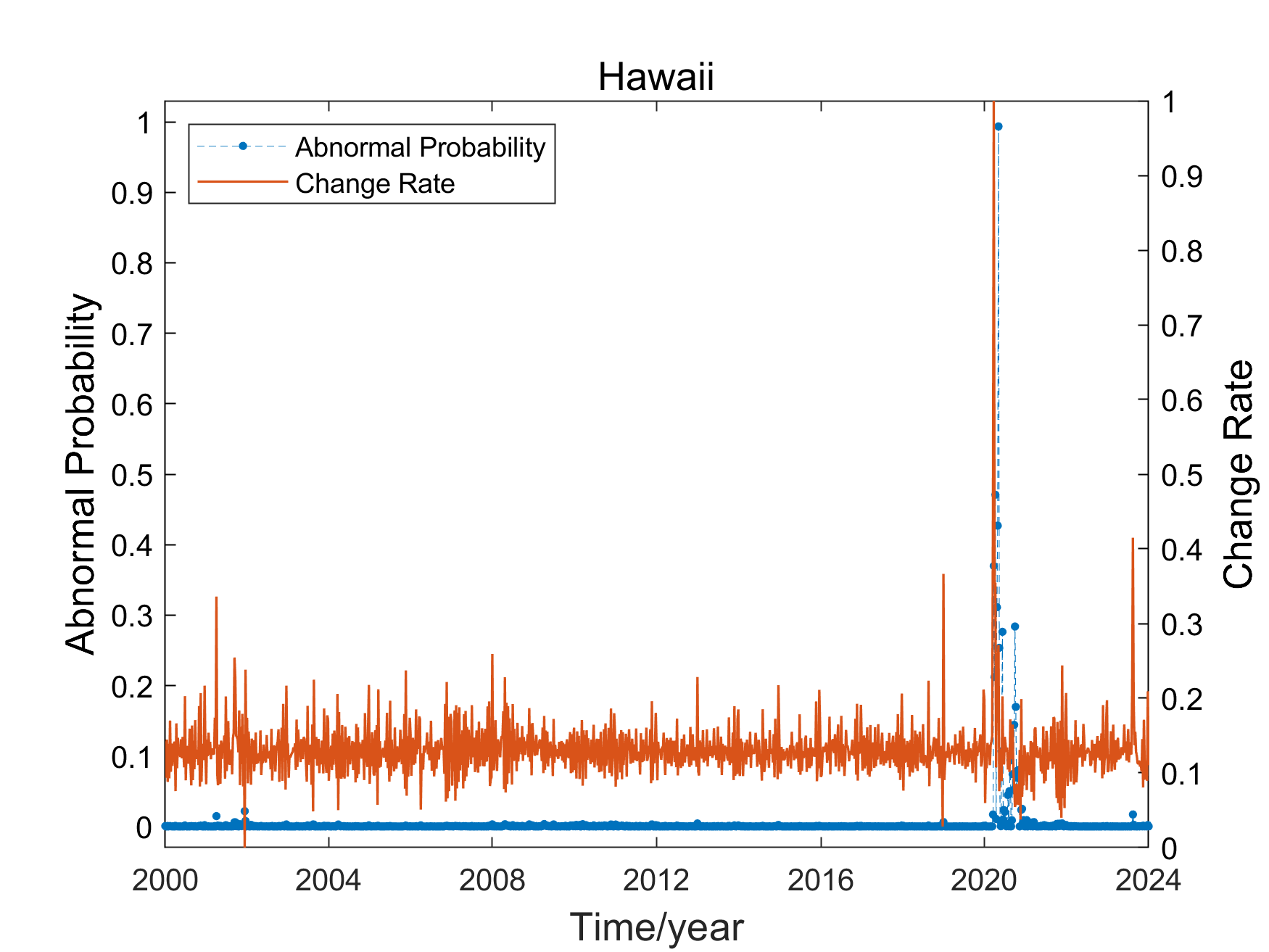}
			\end{minipage}
			\begin{minipage}{0.15\textwidth}
				\centering
				\includegraphics[scale=0.16]{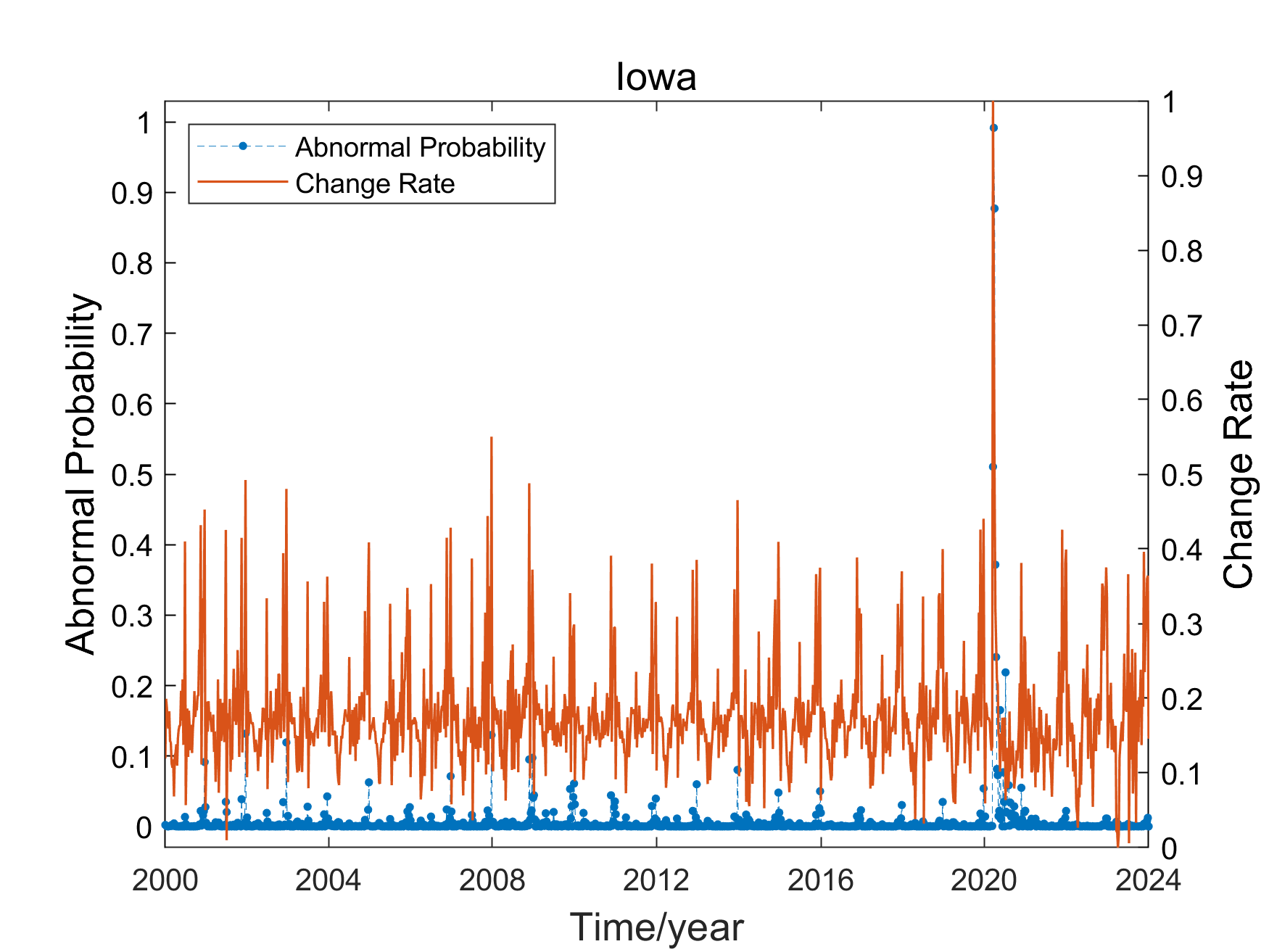}
			\end{minipage}
			\begin{minipage}{0.15\textwidth}
				\centering
				\includegraphics[scale=0.16]{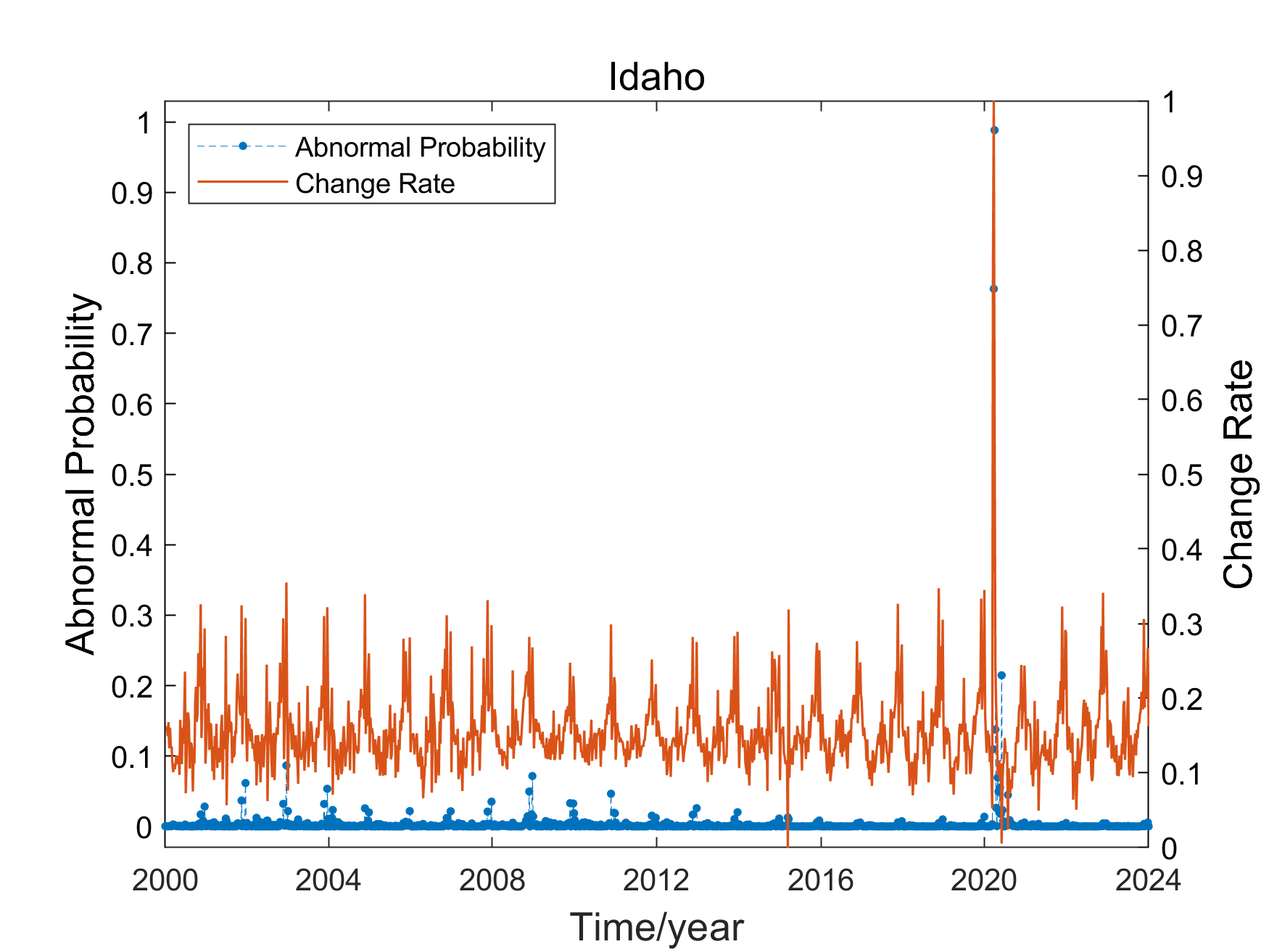}
			\end{minipage}
			\begin{minipage}{0.15\textwidth}
				\centering
				\includegraphics[scale=0.16]{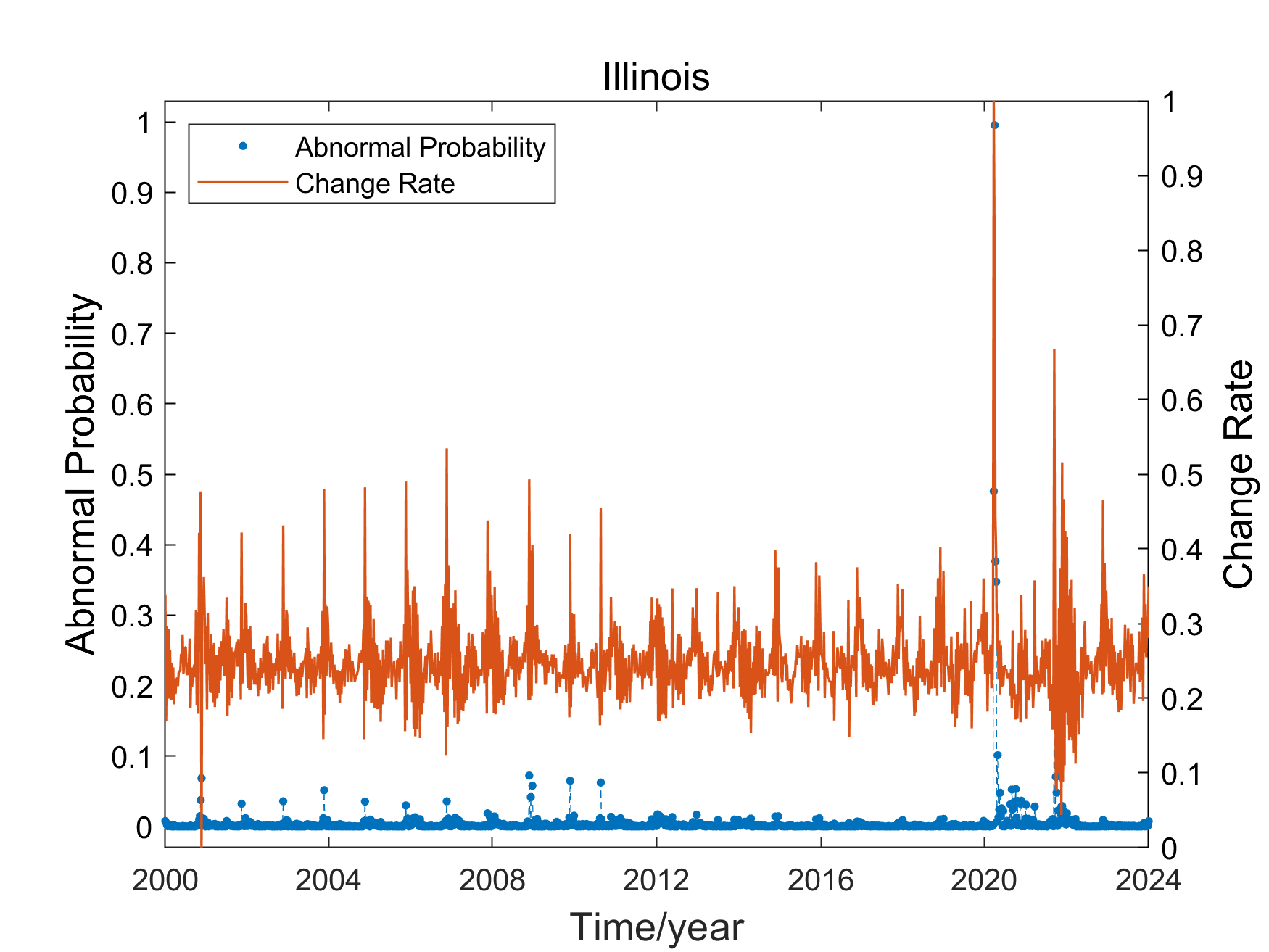}
			\end{minipage}
			\begin{minipage}{0.15\textwidth}
				\centering
				\includegraphics[scale=0.16]{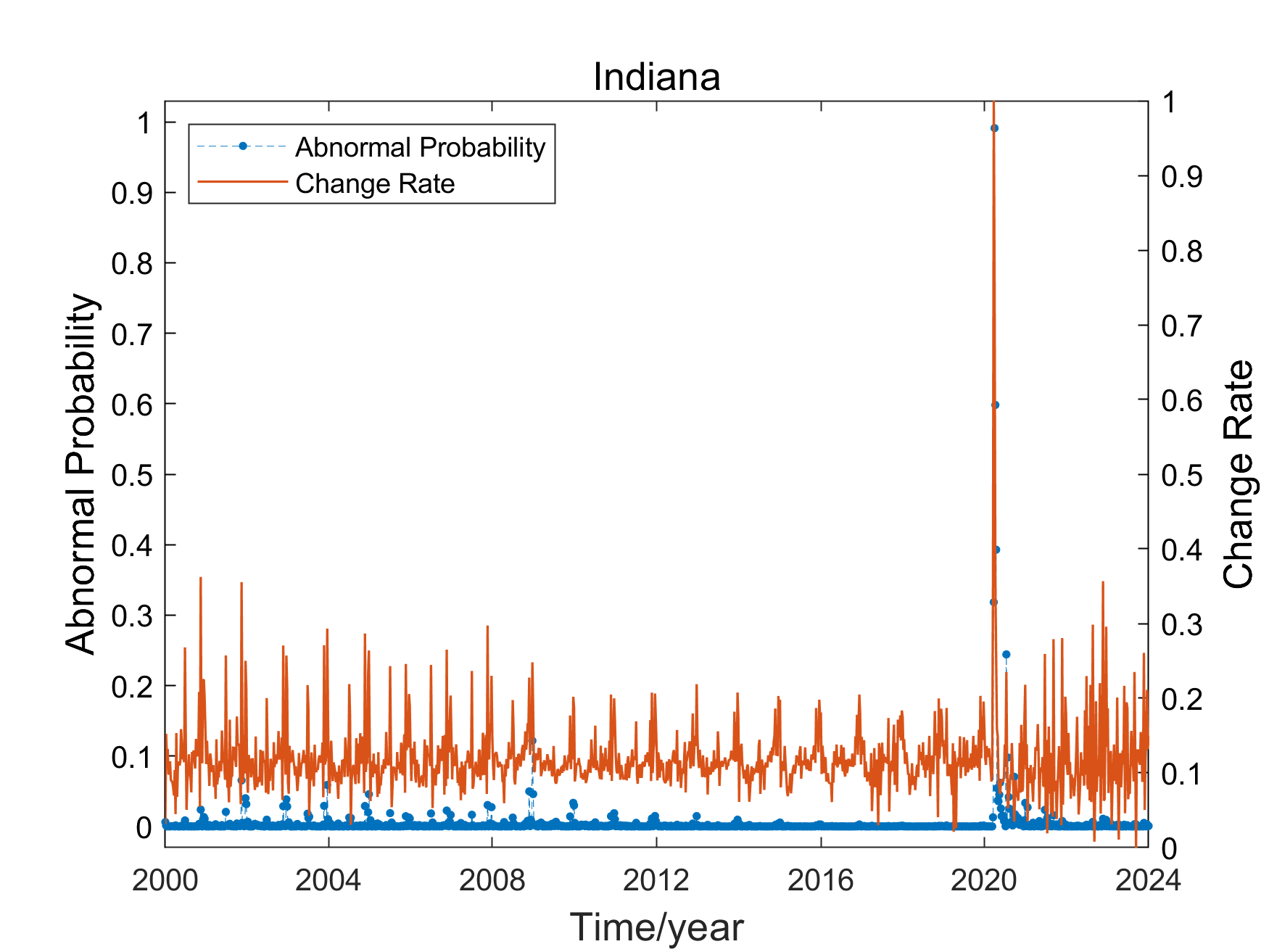}
			\end{minipage}
			
		}\\
		\subfigure{
			\begin{minipage}{0.15\textwidth}
				\centering
				\includegraphics[scale=0.16]{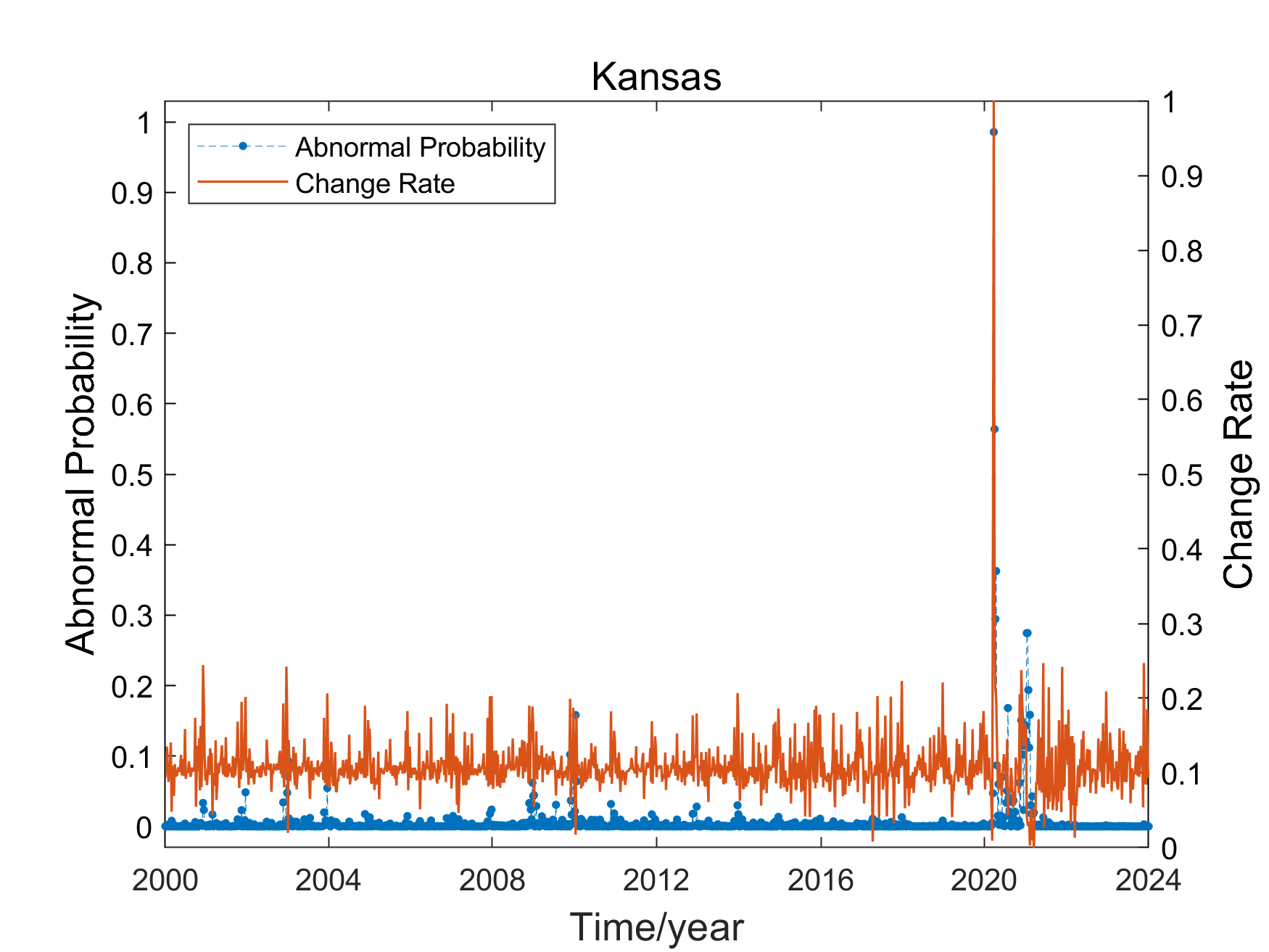}
			\end{minipage}
			\begin{minipage}{0.15\textwidth}
				\centering
				\includegraphics[scale=0.16]{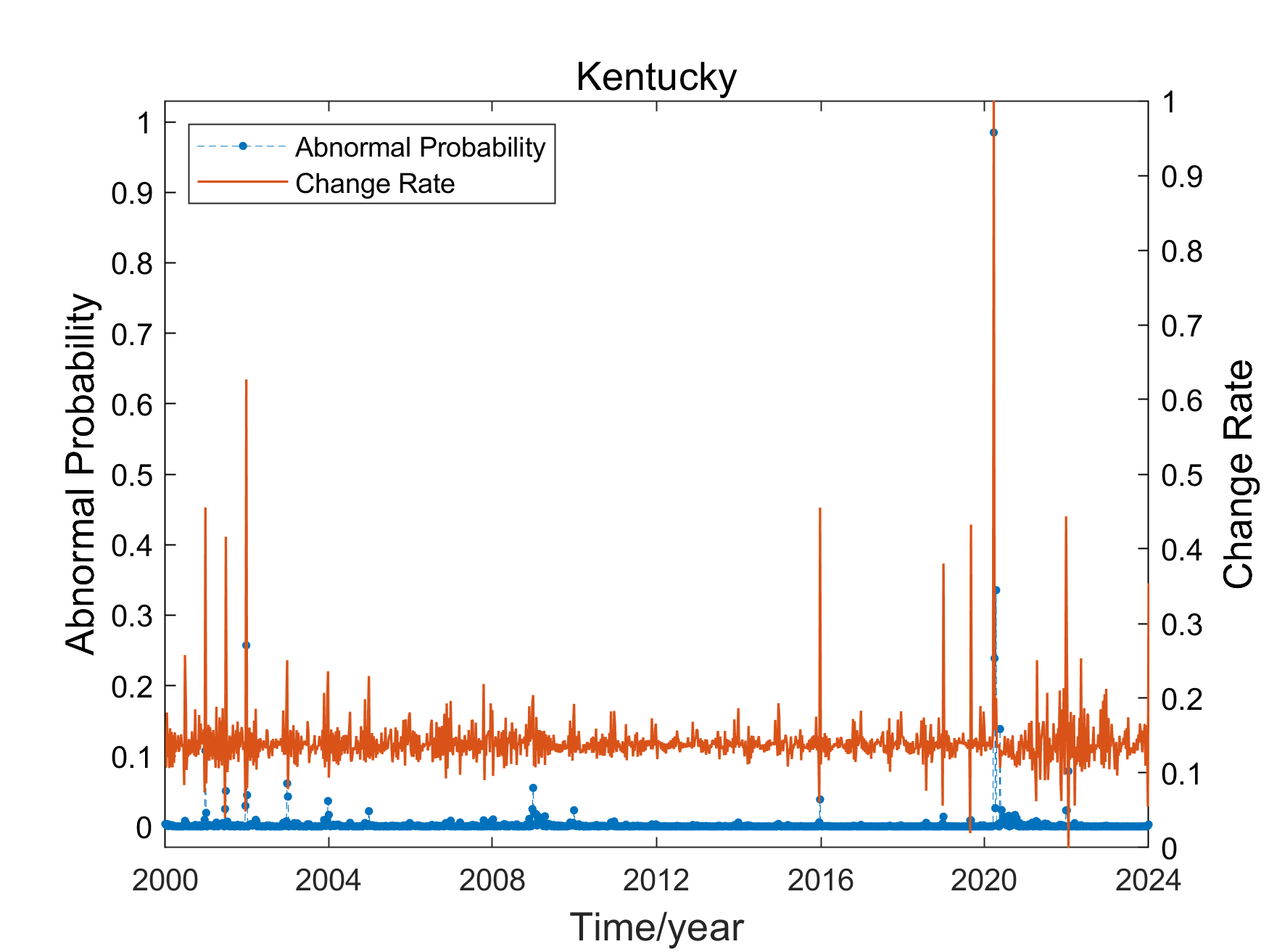}
			\end{minipage}
			\begin{minipage}{0.15\textwidth}
				\centering
				\includegraphics[scale=0.16]{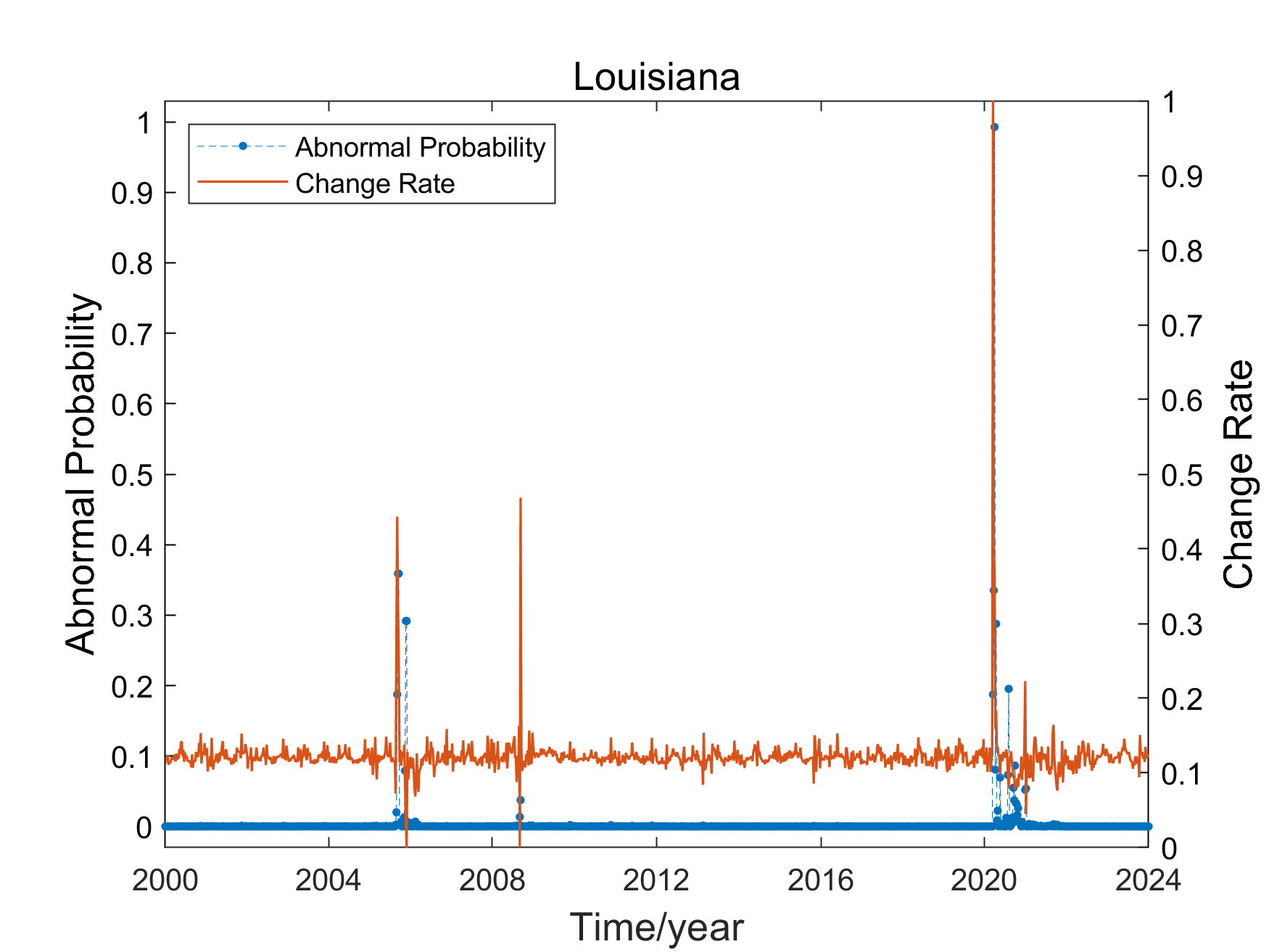}
			\end{minipage}
			\begin{minipage}{0.15\textwidth}
				\centering
				\includegraphics[scale=0.16]{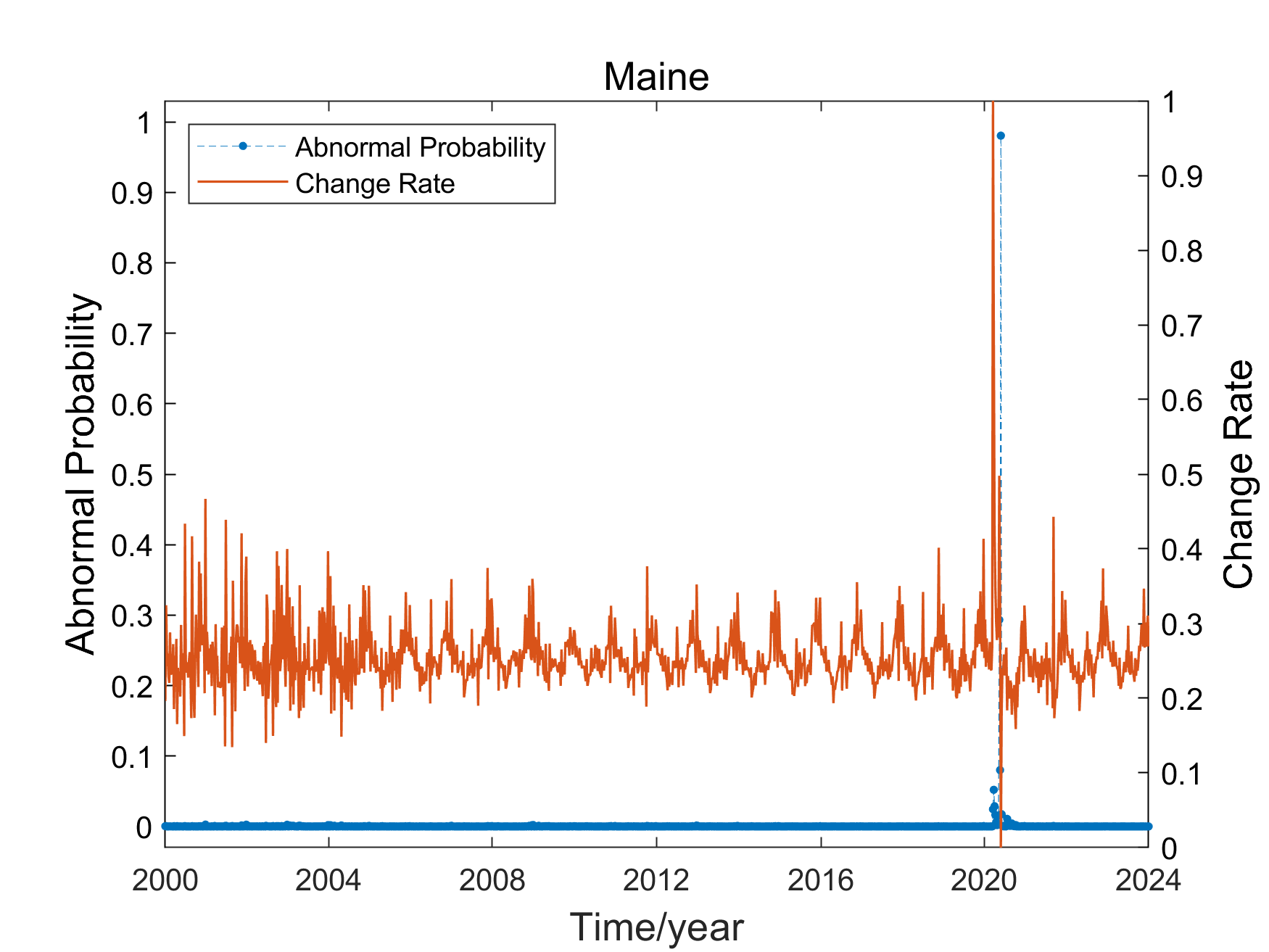}
			\end{minipage}
			\begin{minipage}{0.15\textwidth}
				\centering
				\includegraphics[scale=0.16]{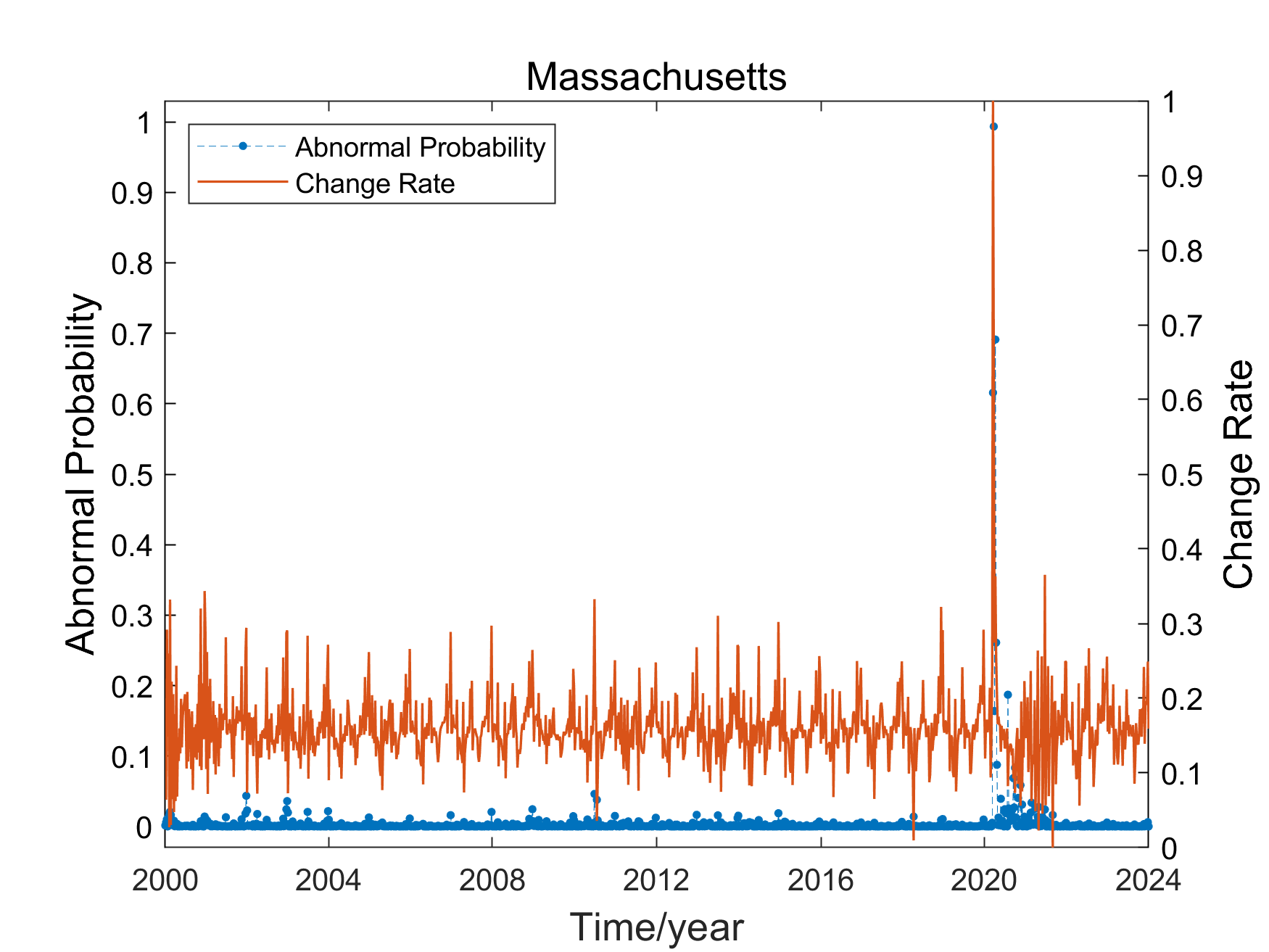}
			\end{minipage}
		}\\
		\subfigure{
			\begin{minipage}{0.15\textwidth}
				\centering
				\includegraphics[scale=0.16]{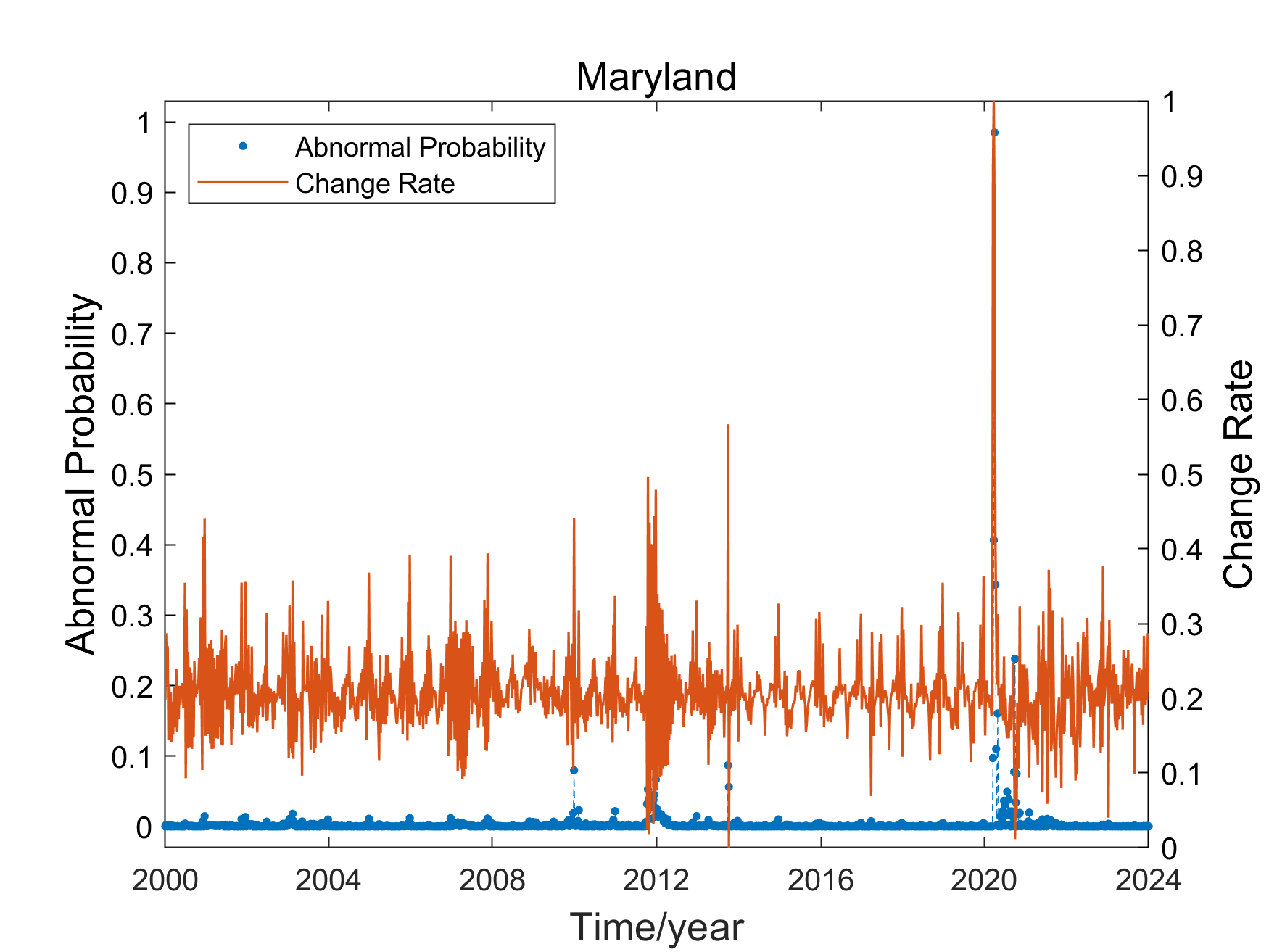}
			\end{minipage}
			\begin{minipage}{0.15\textwidth}
				\centering
				\includegraphics[scale=0.16]{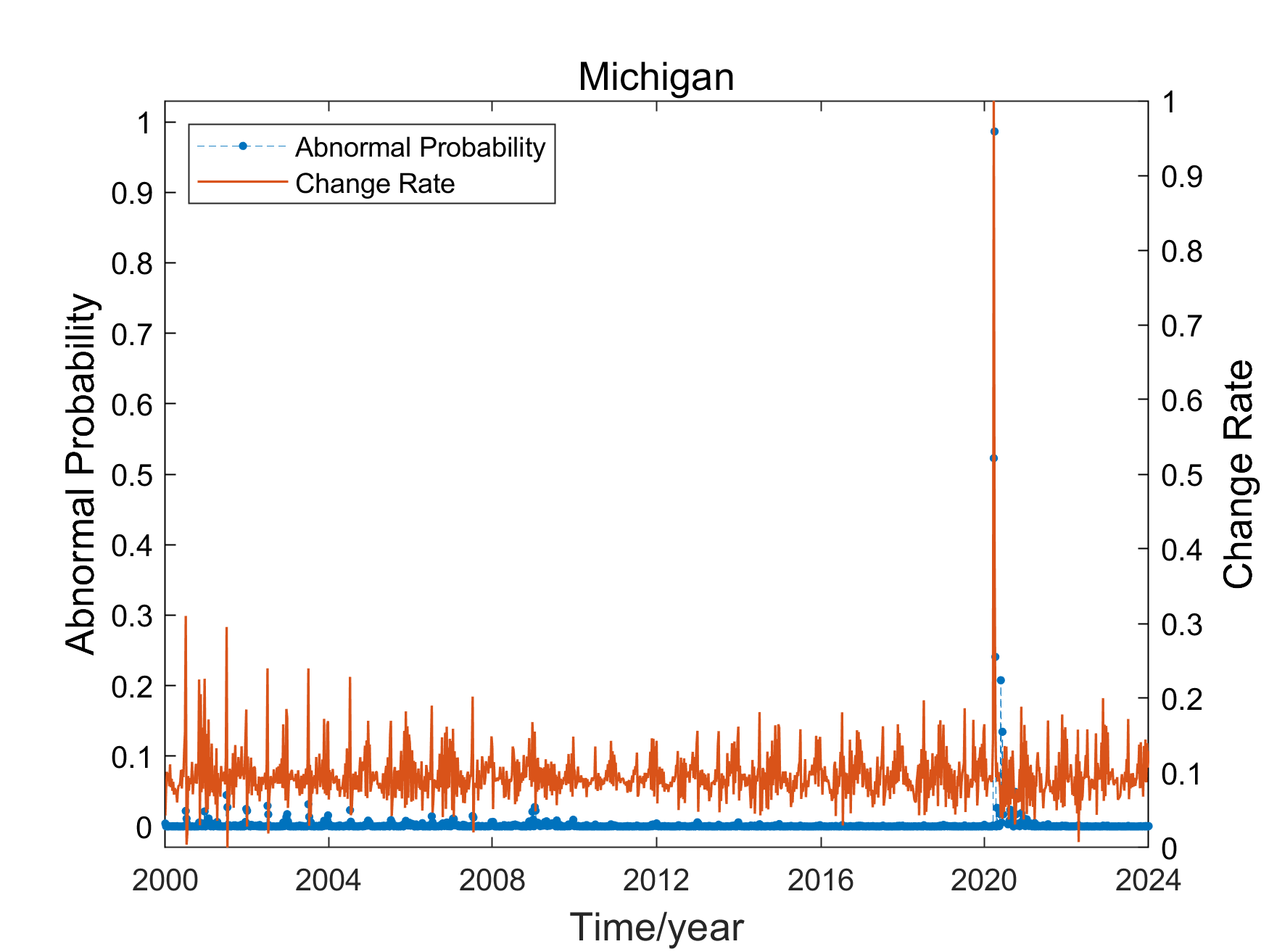}
			\end{minipage}
			\begin{minipage}{0.15\textwidth}
				\centering
				\includegraphics[scale=0.16]{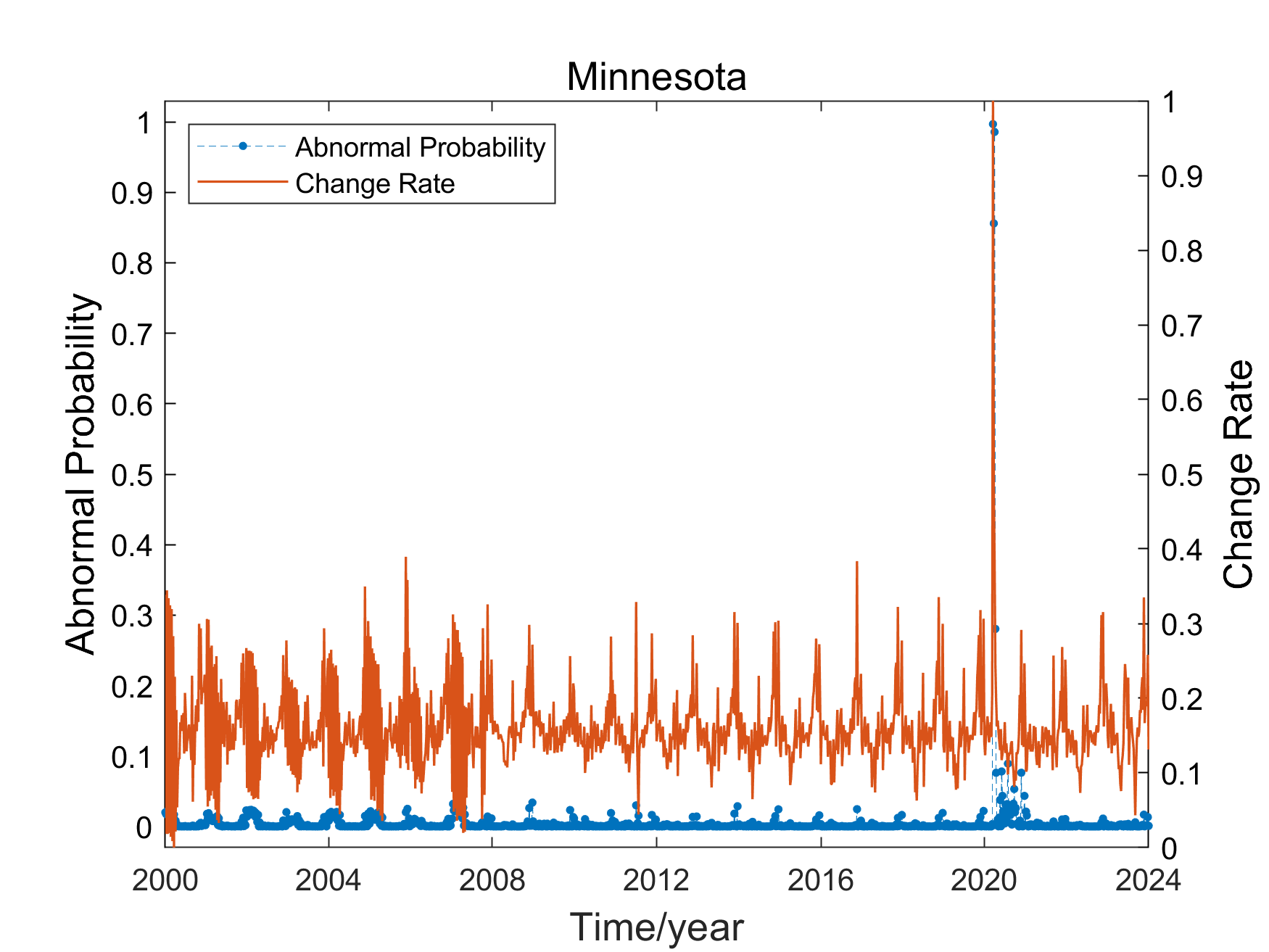}
			\end{minipage}
			\begin{minipage}{0.15\textwidth}
				\centering
				\includegraphics[scale=0.16]{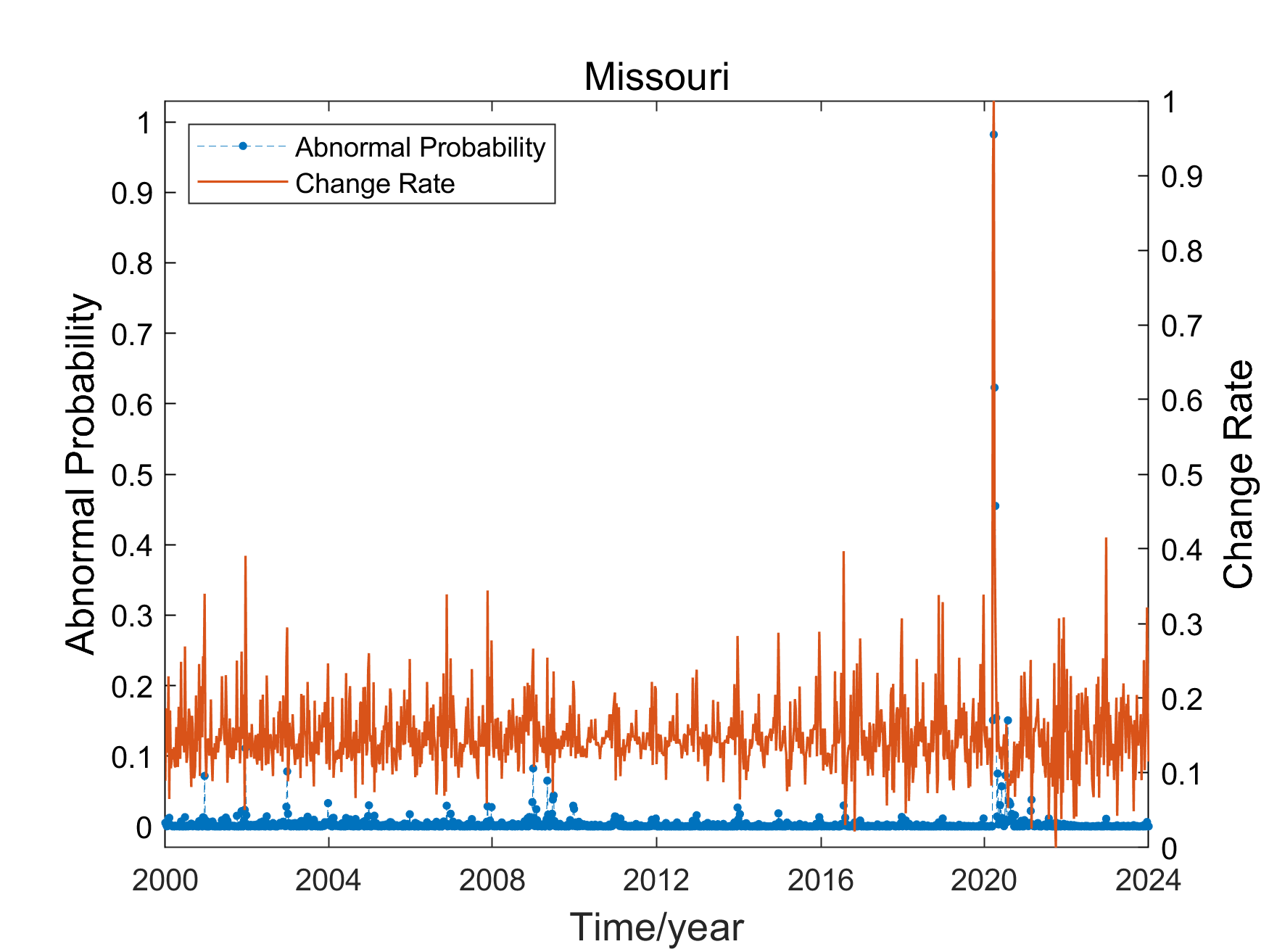}
			\end{minipage}
			\begin{minipage}{0.15\textwidth}
				\centering
				\includegraphics[scale=0.16]{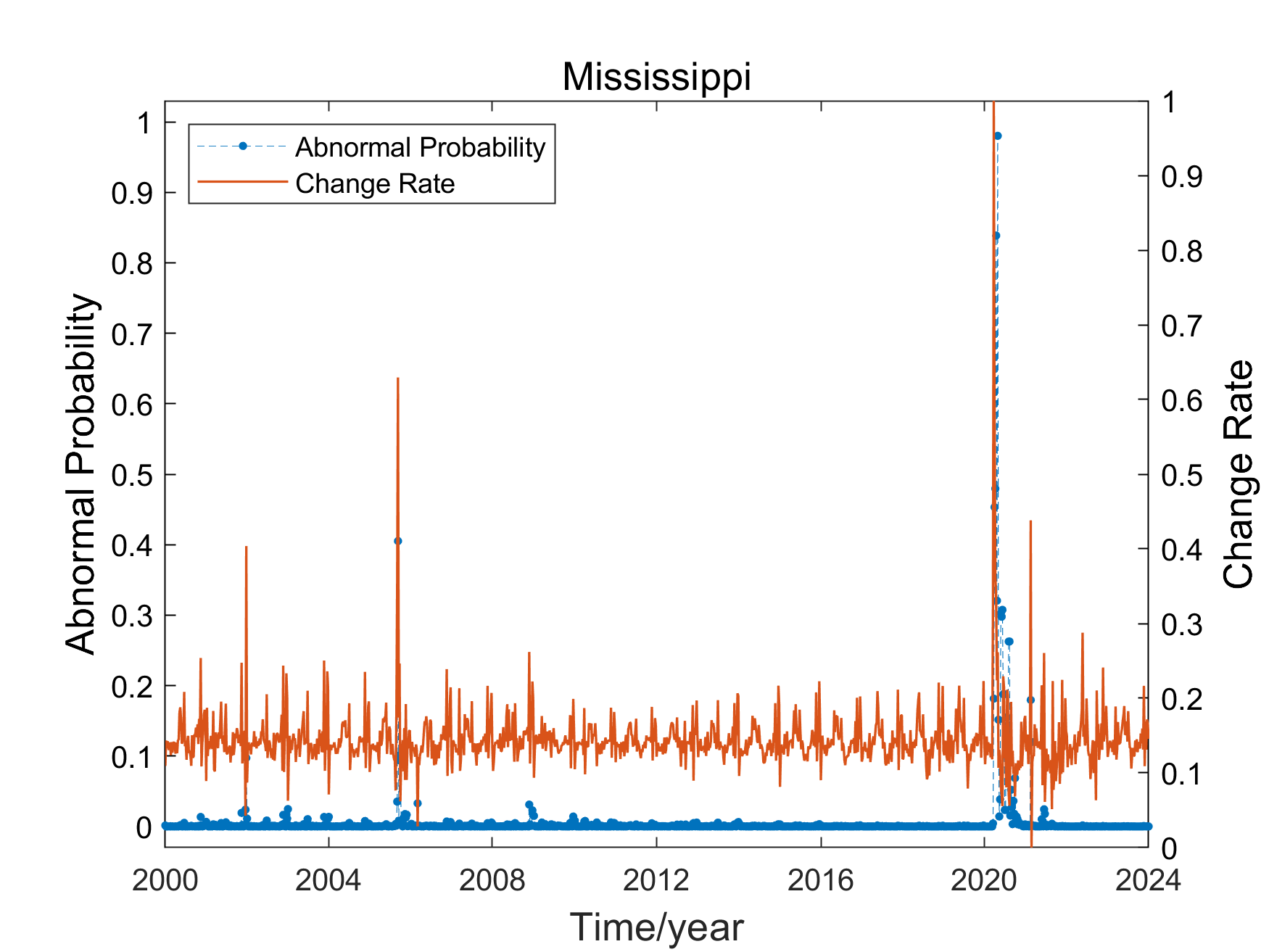}
			\end{minipage}
			
		}\\
		\subfigure{
			\begin{minipage}{0.15\textwidth}
				\centering
				\includegraphics[scale=0.16]{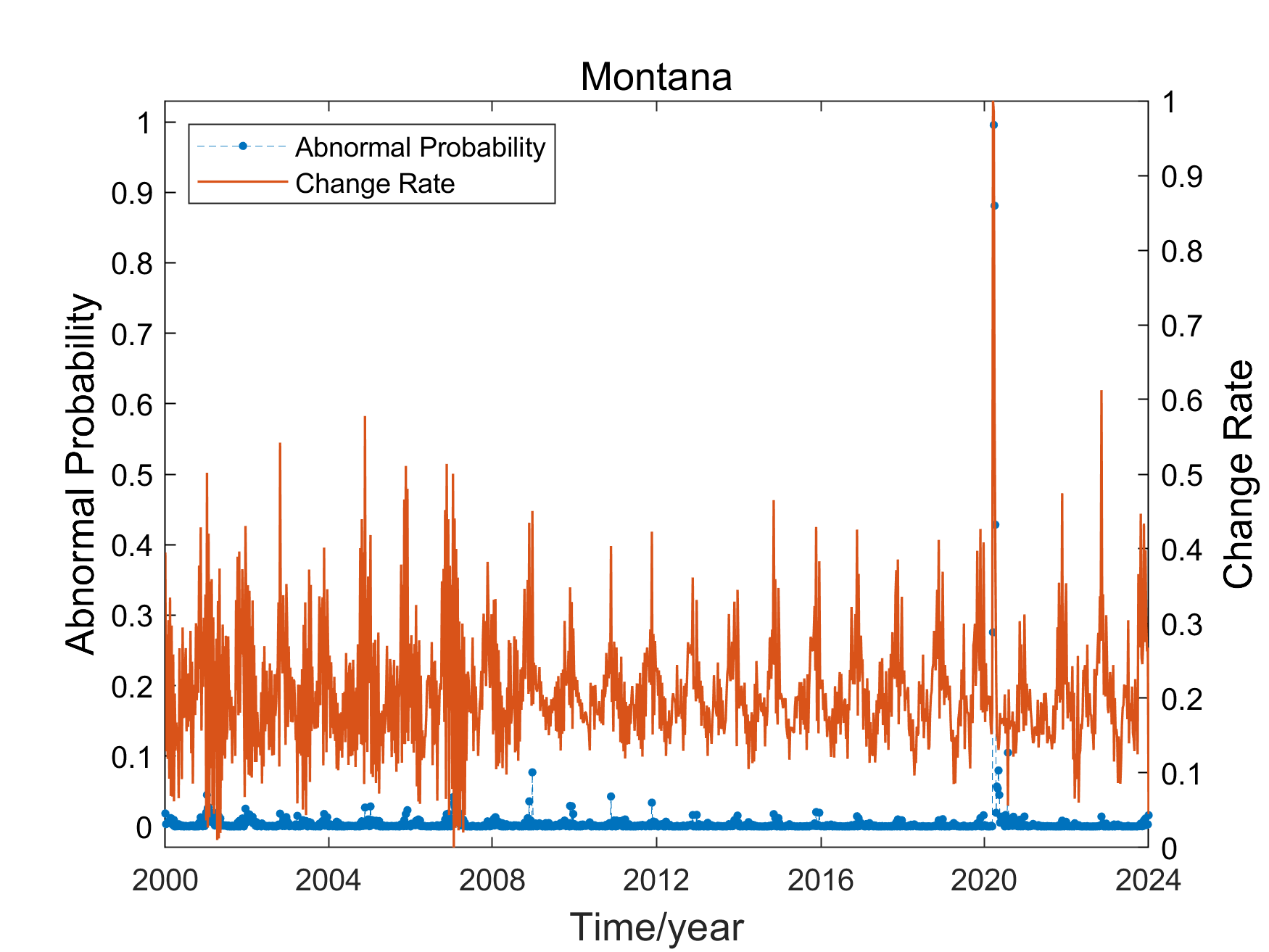}
			\end{minipage}
			\begin{minipage}{0.15\textwidth}
				\centering
				\includegraphics[scale=0.16]{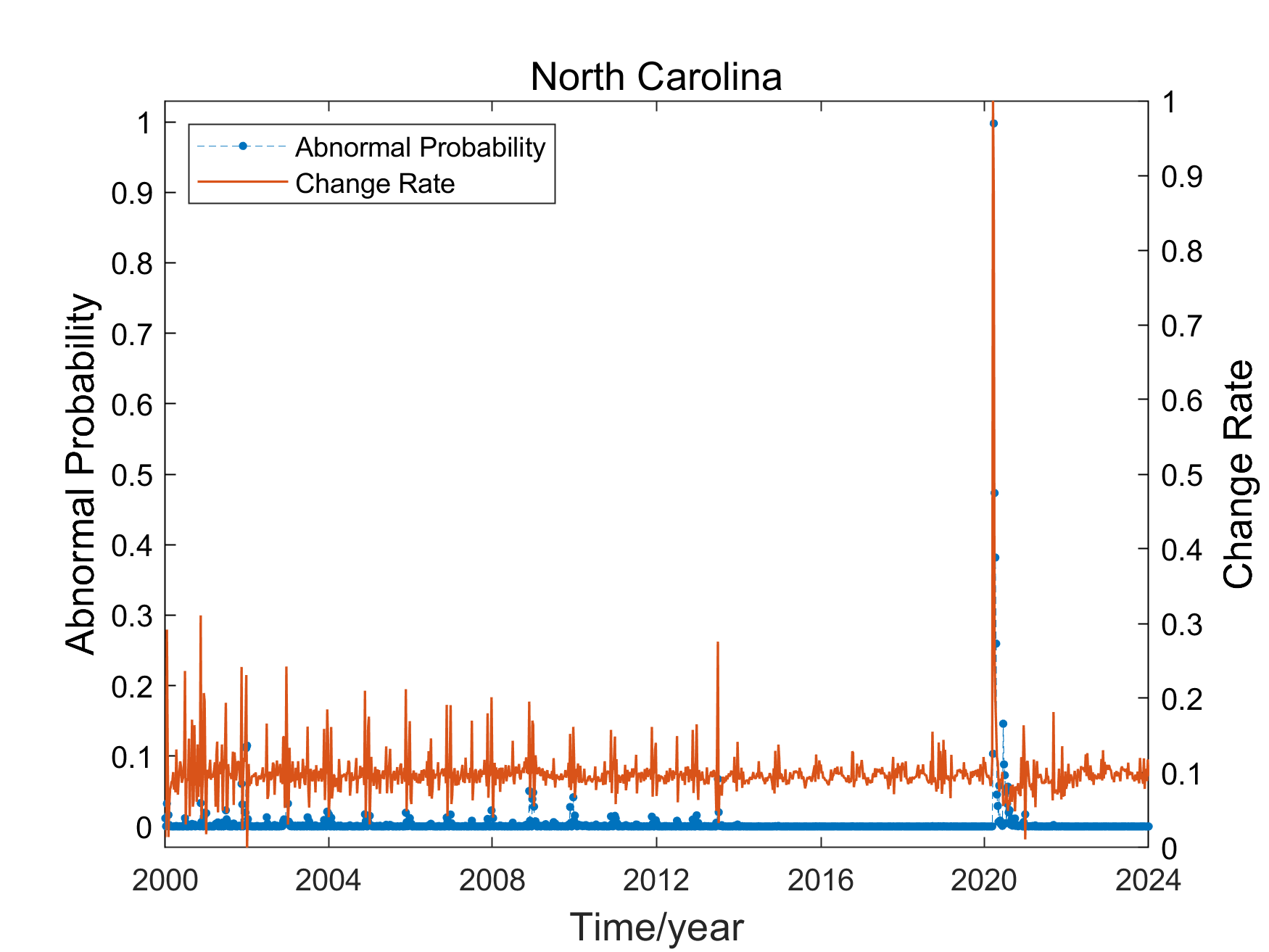}
			\end{minipage}
			\begin{minipage}{0.15\textwidth}
				\centering
				\includegraphics[scale=0.16]{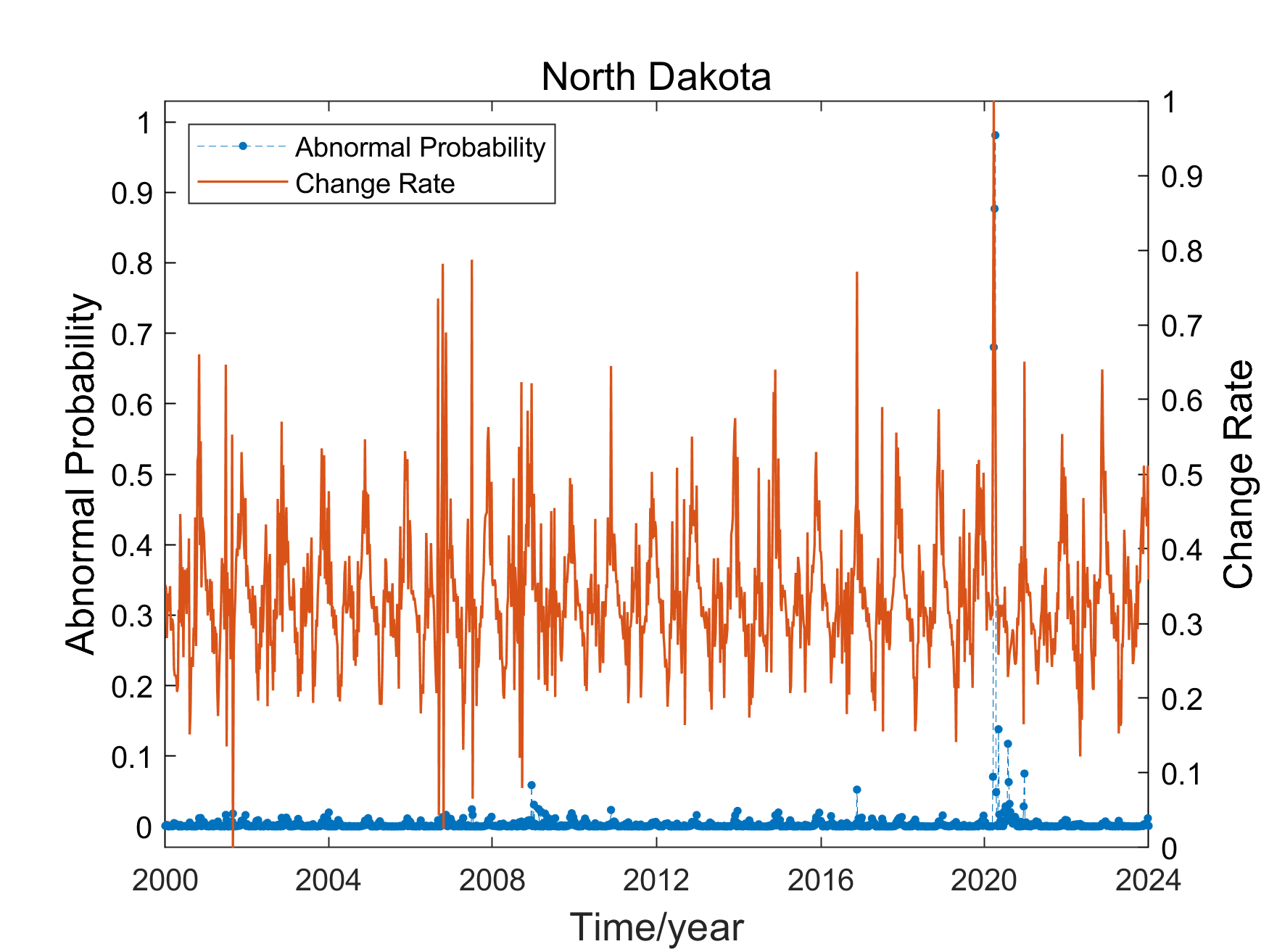}
			\end{minipage}
			\begin{minipage}{0.15\textwidth}
				\centering
				\includegraphics[scale=0.16]{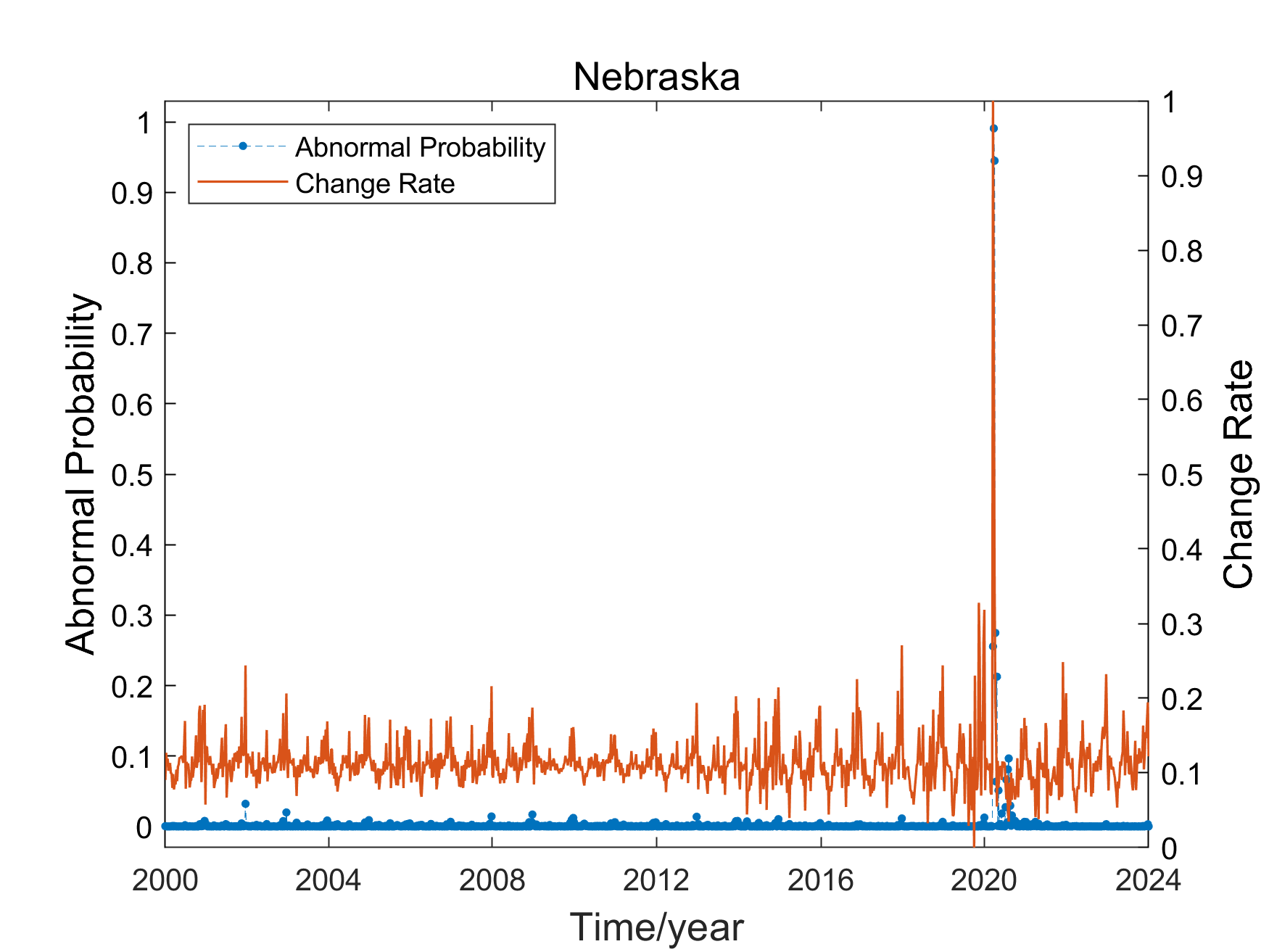}
			\end{minipage}
			\begin{minipage}{0.15\textwidth}
				\centering
				\includegraphics[scale=0.16]{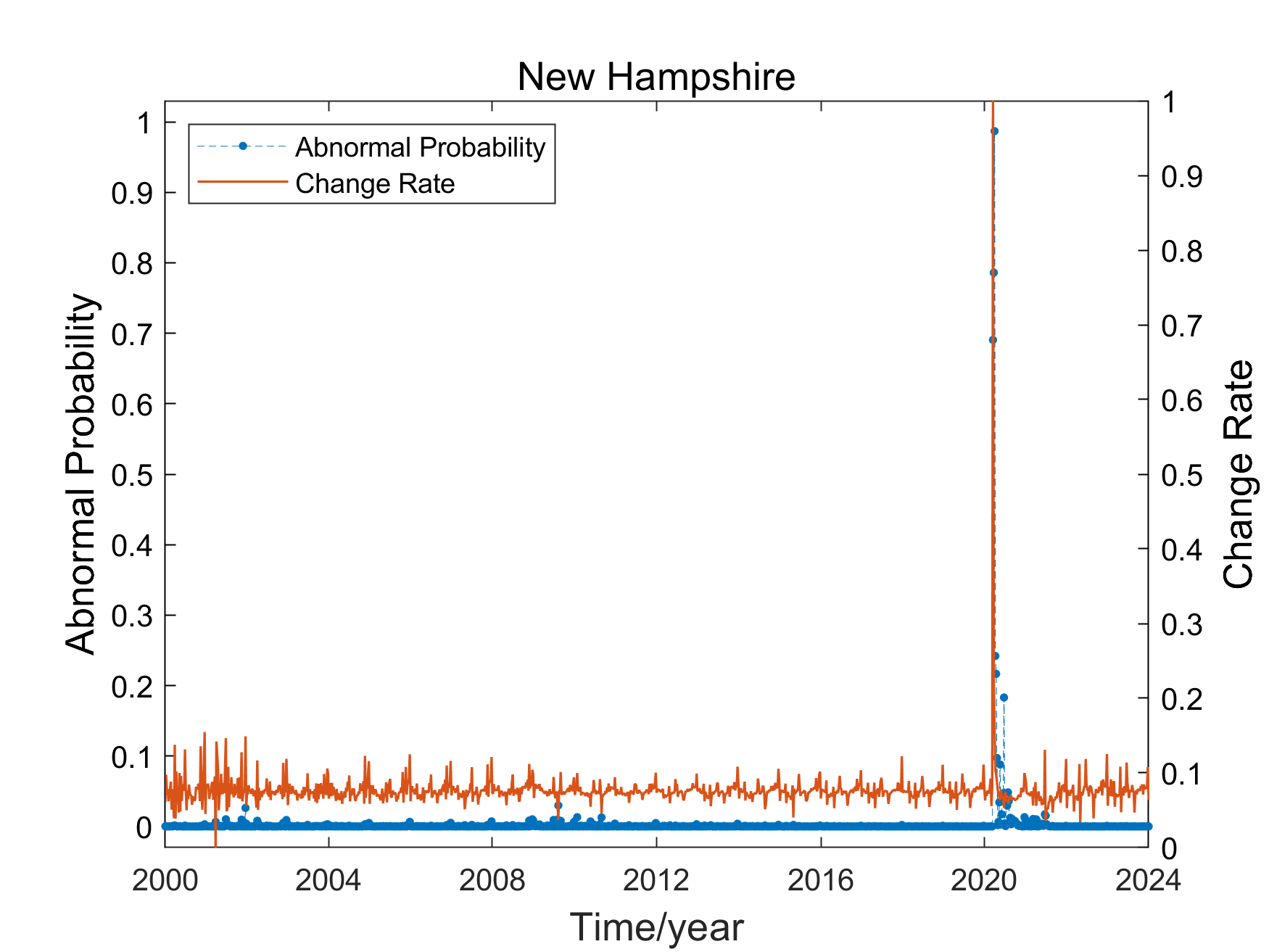}
			\end{minipage}
		}\\
		\subfigure{
			\begin{minipage}{0.15\textwidth}
				\centering
				\includegraphics[scale=0.16]{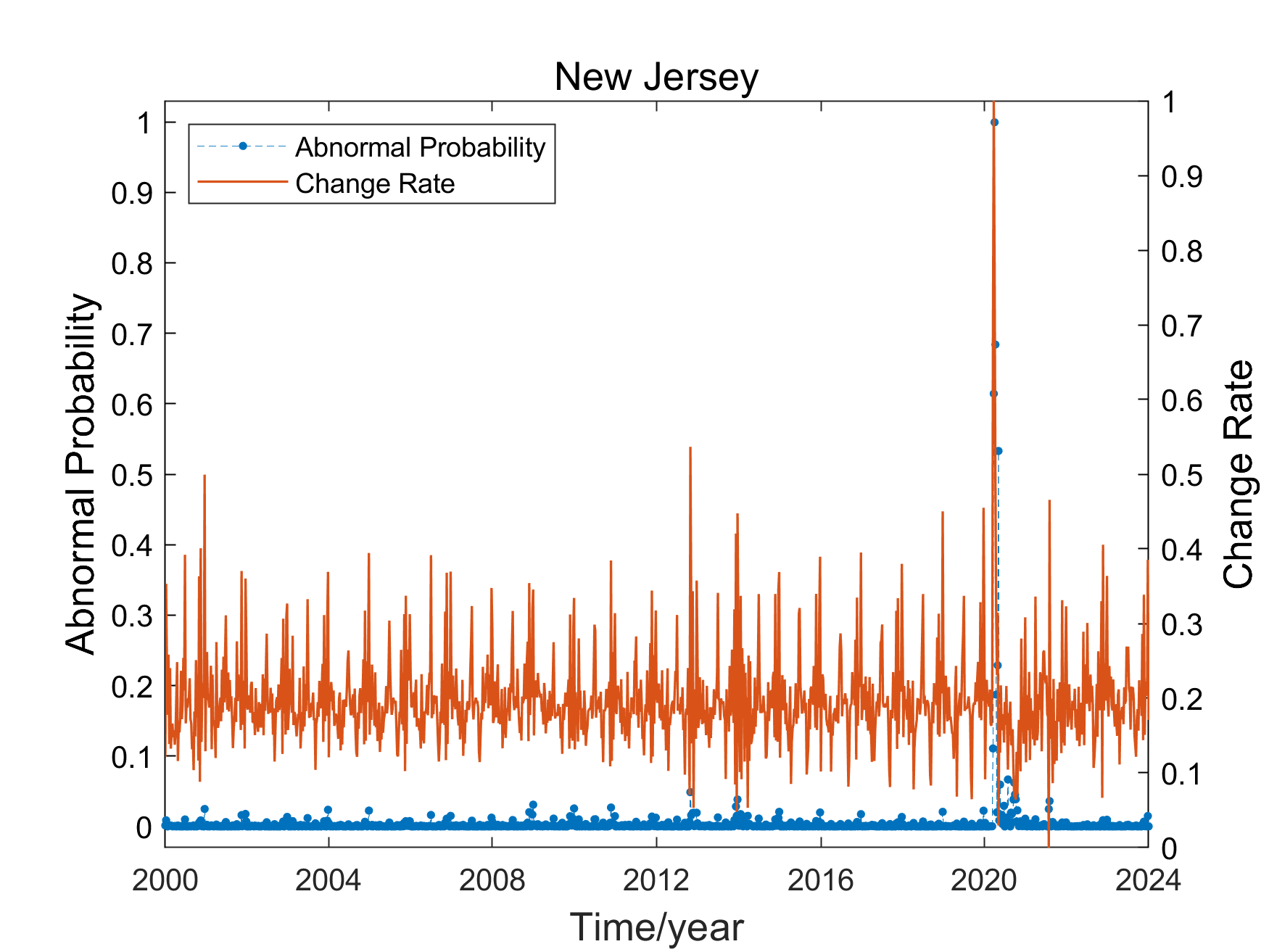}
			\end{minipage}
			\begin{minipage}{0.15\textwidth}
				\centering
				\includegraphics[scale=0.16]{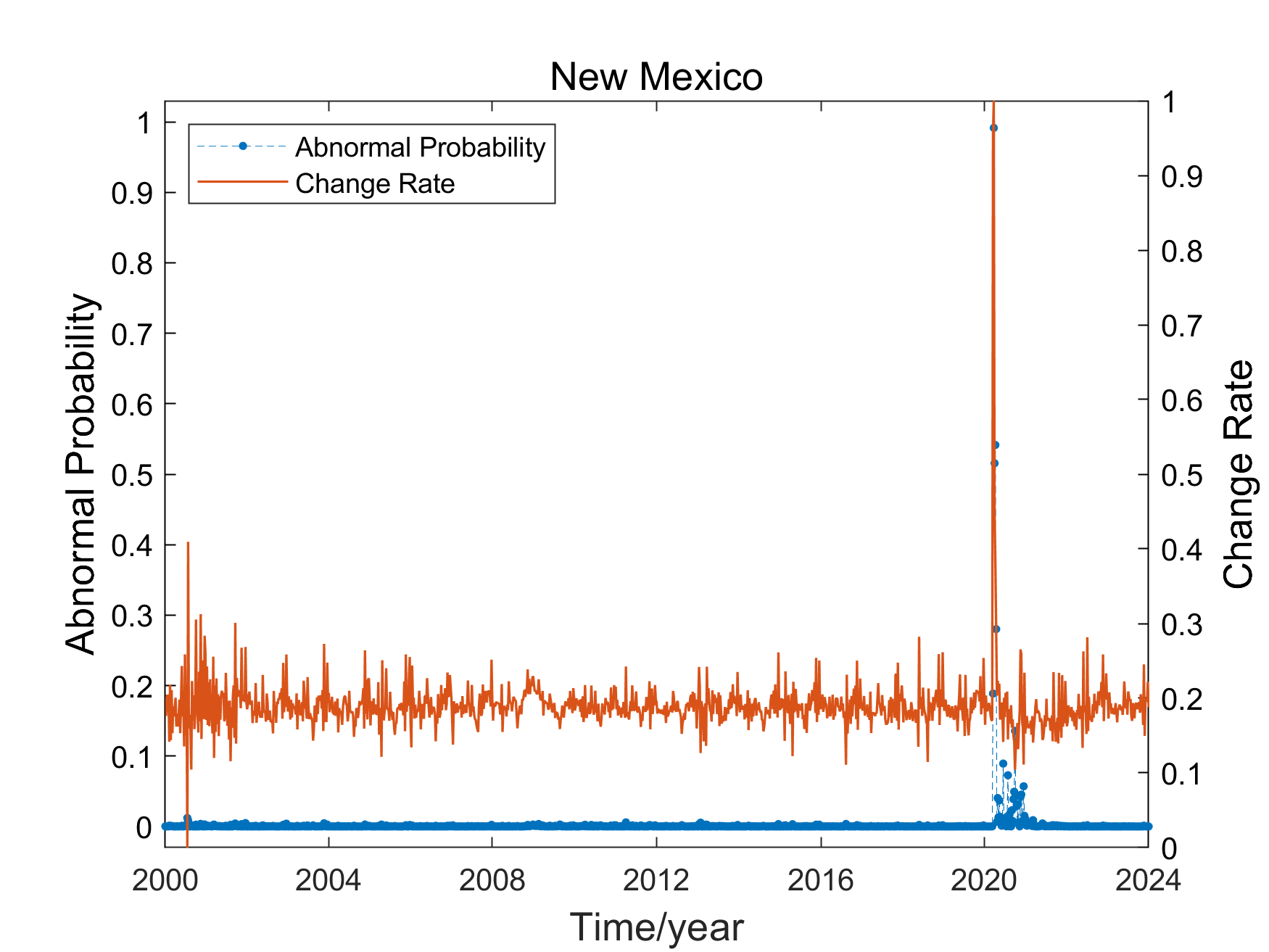}
			\end{minipage}
			\begin{minipage}{0.15\textwidth}
				\centering
				\includegraphics[scale=0.16]{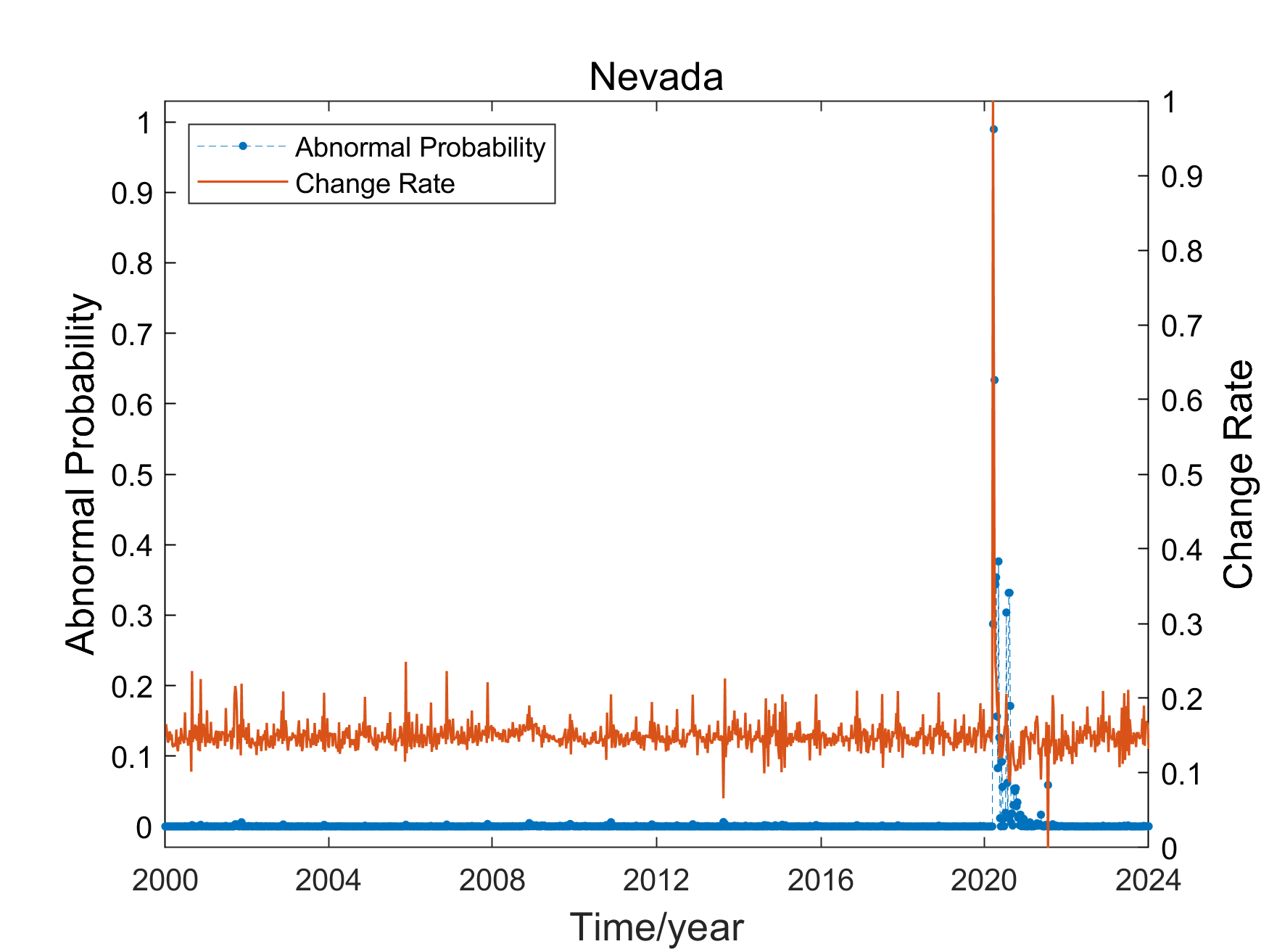}
			\end{minipage}
			\begin{minipage}{0.15\textwidth}
				\centering
				\includegraphics[scale=0.16]{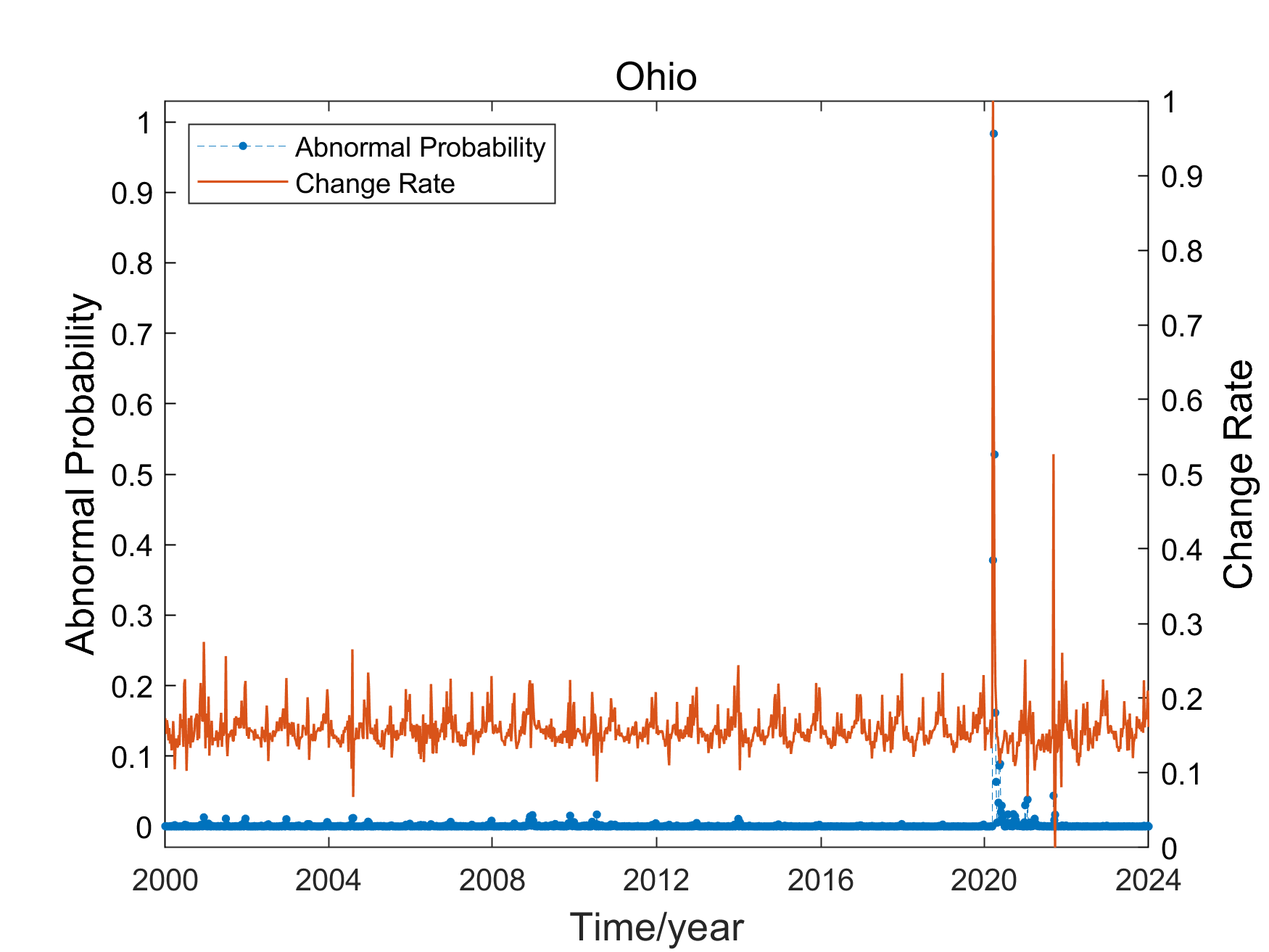}
			\end{minipage}
			\begin{minipage}{0.15\textwidth}
				\centering
				\includegraphics[scale=0.16]{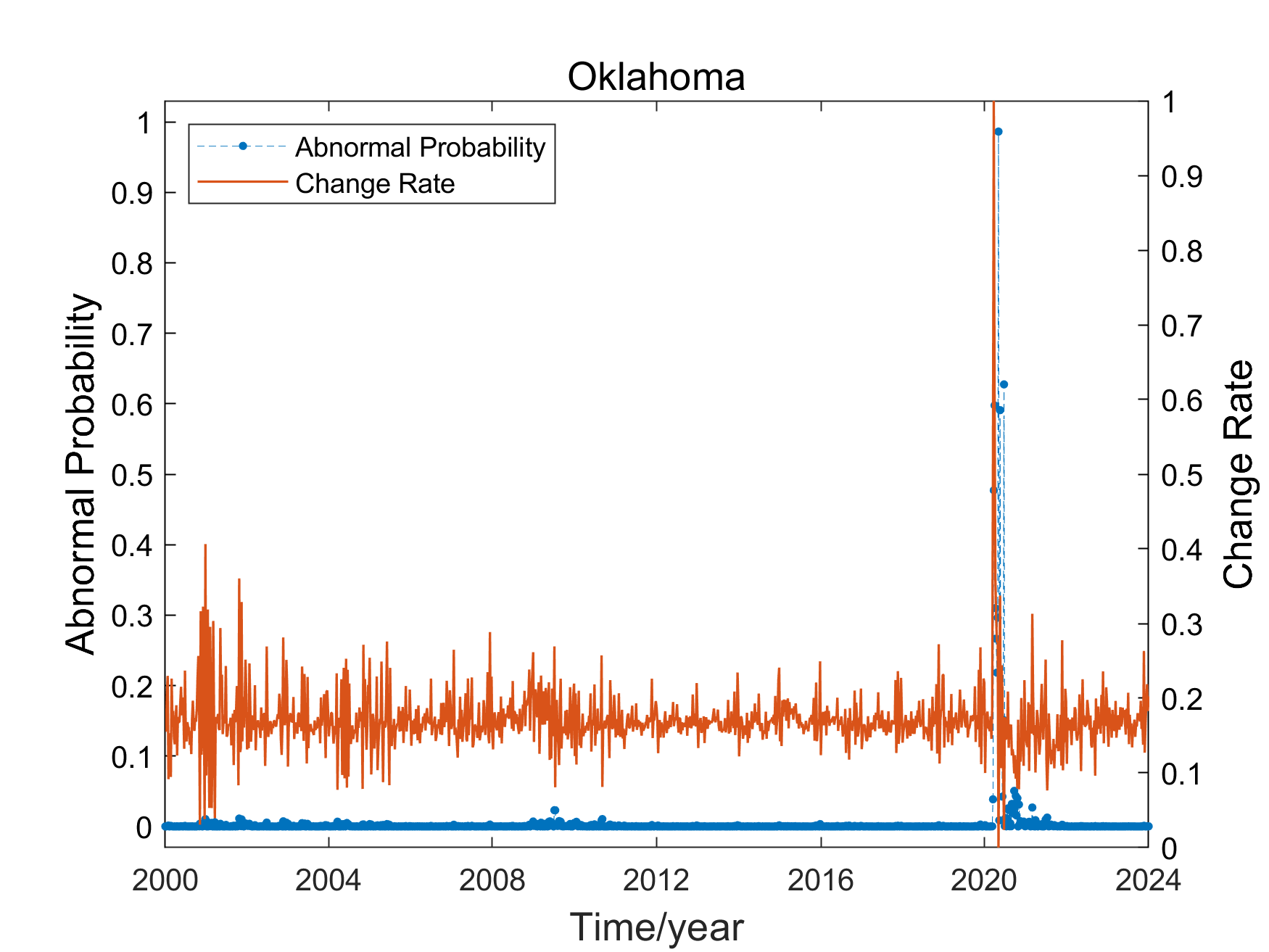}
			\end{minipage}
			
		}\\
		\subfigure{
			\begin{minipage}{0.15\textwidth}
				\centering
				\includegraphics[scale=0.16]{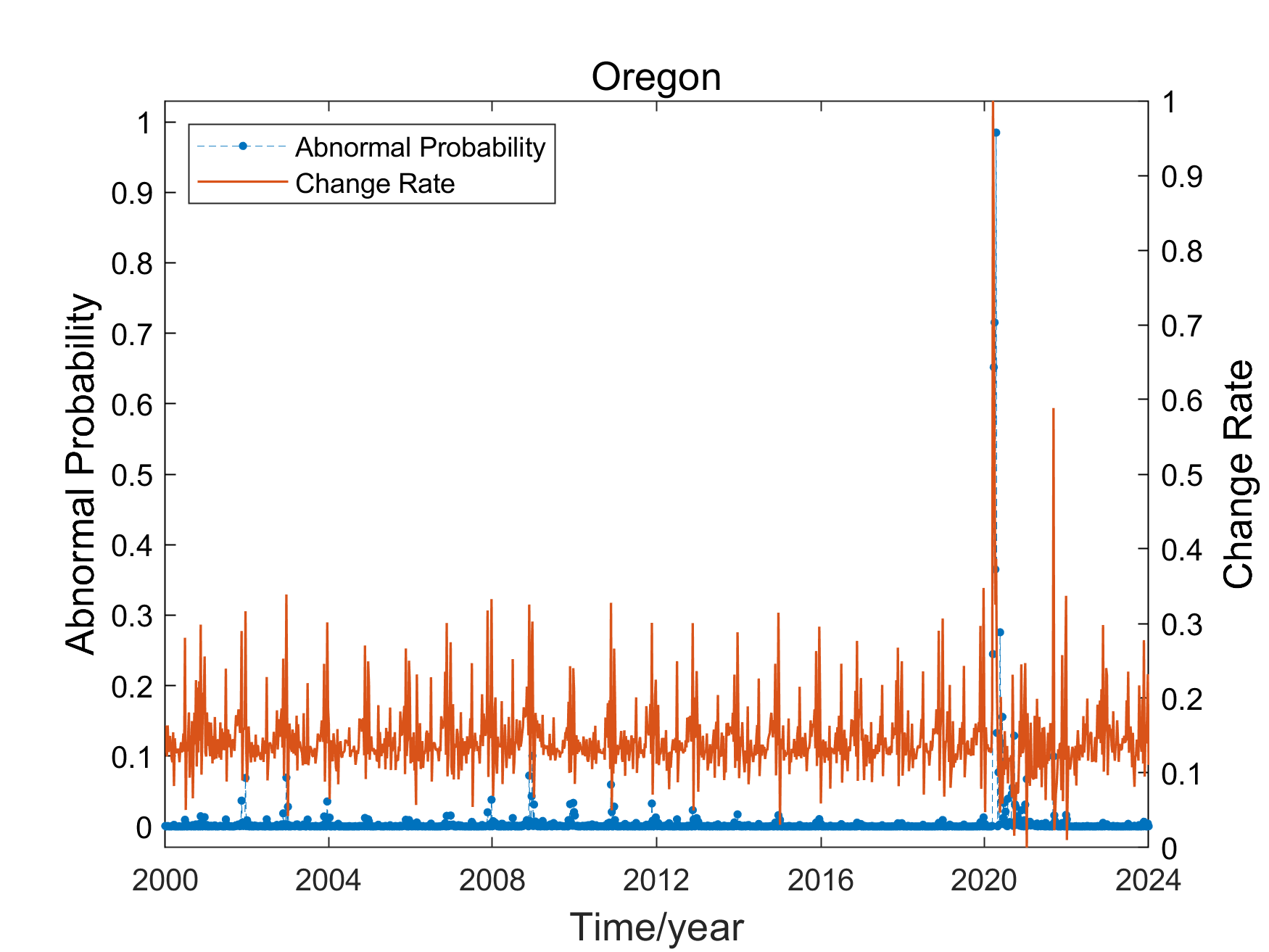}
			\end{minipage}
			\begin{minipage}{0.15\textwidth}
				\centering
				\includegraphics[scale=0.16]{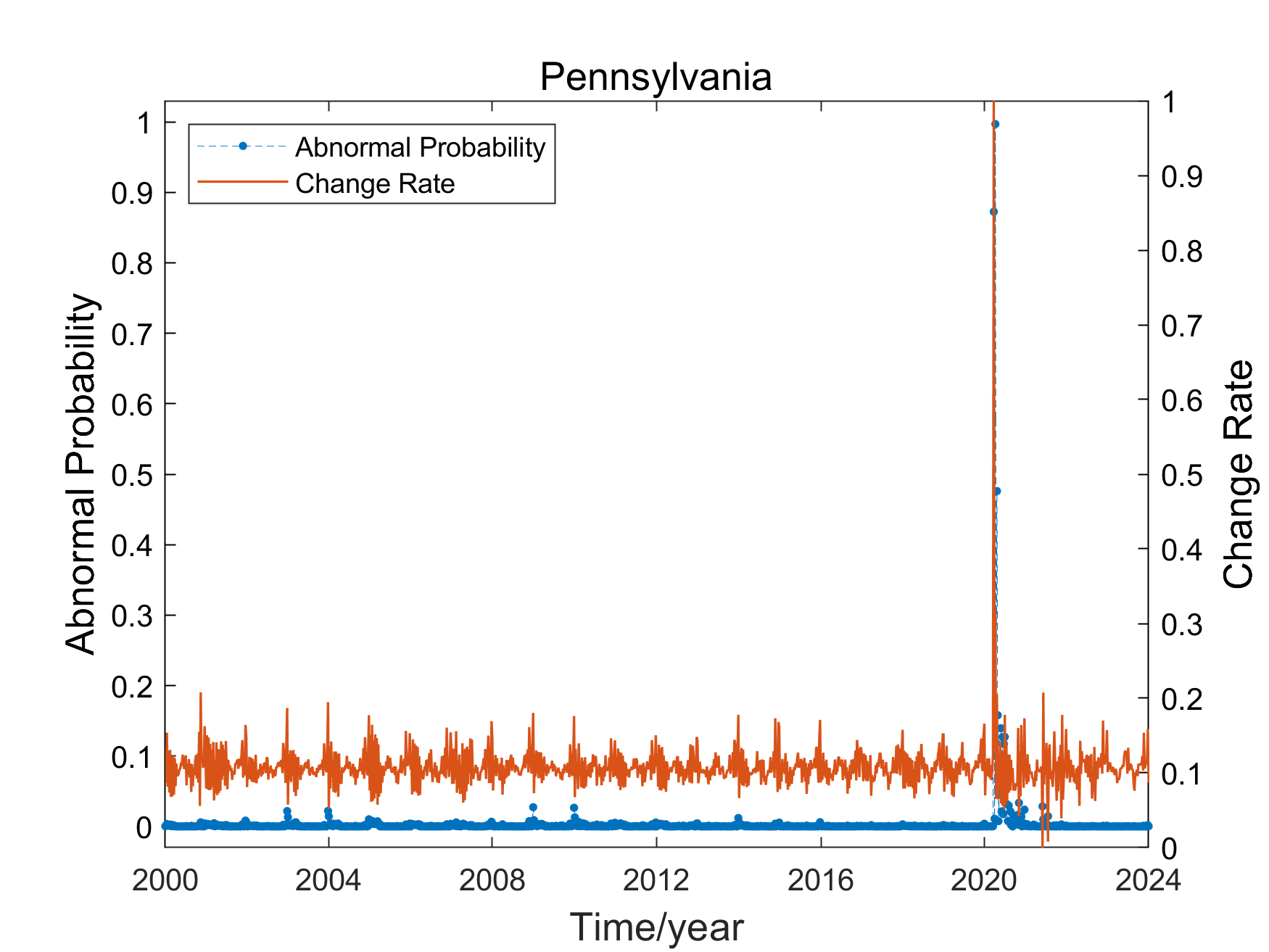}
			\end{minipage}
			\begin{minipage}{0.15\textwidth}
				\centering
				\includegraphics[scale=0.16]{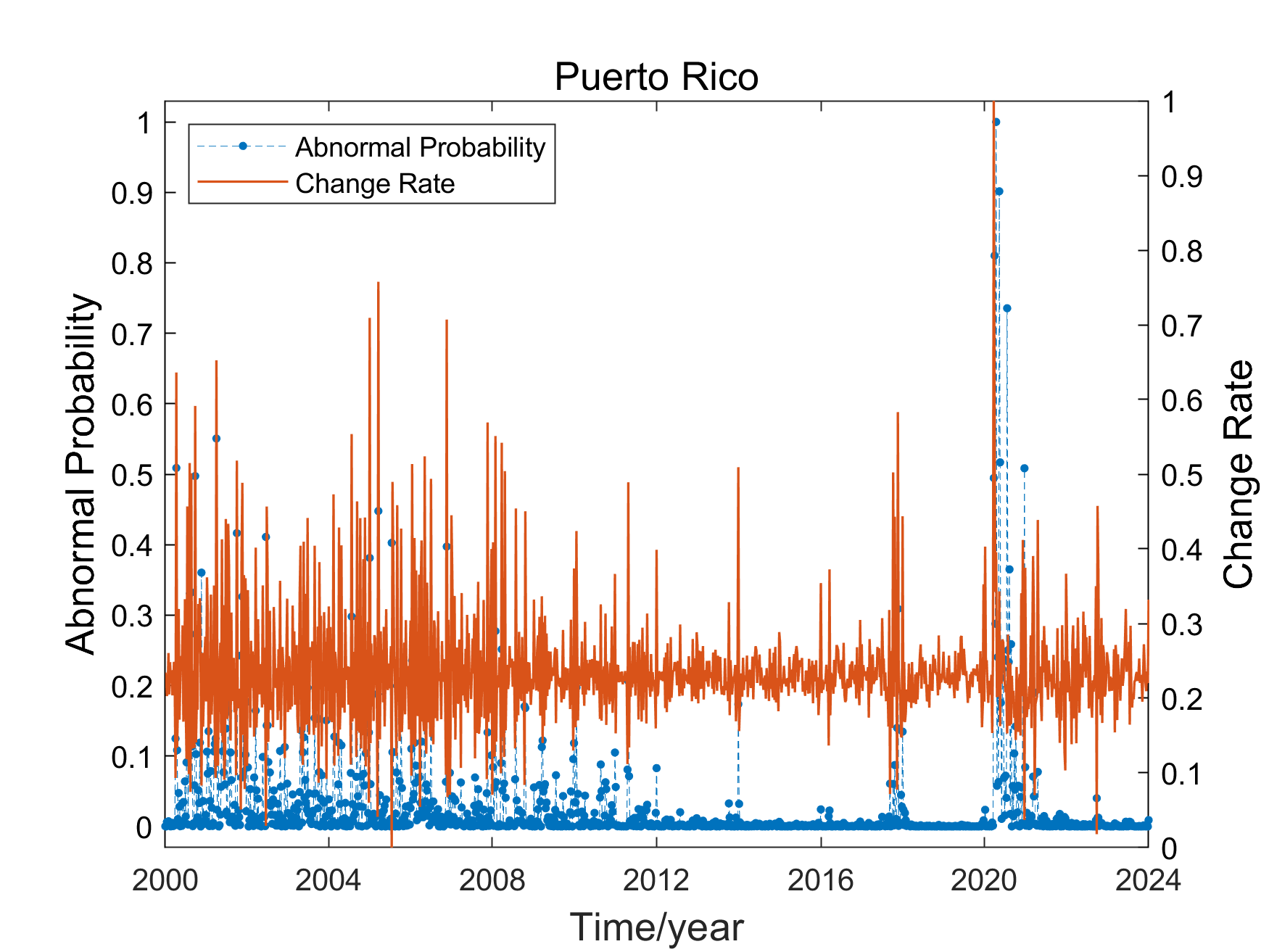}
			\end{minipage}
			\begin{minipage}{0.15\textwidth}
				\centering
				\includegraphics[scale=0.16]{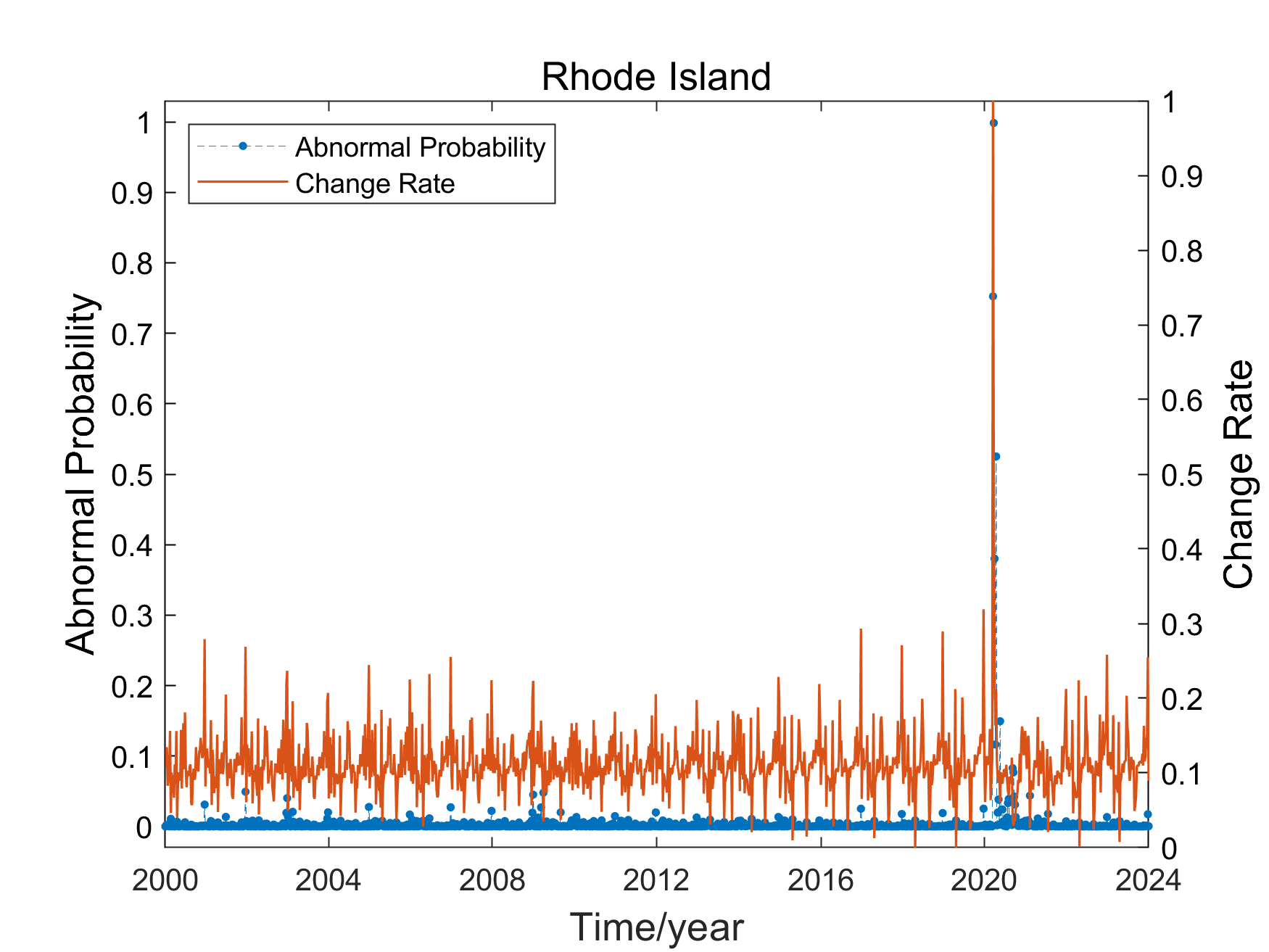}
			\end{minipage}
			\begin{minipage}{0.15\textwidth}
				\centering
				\includegraphics[scale=0.16]{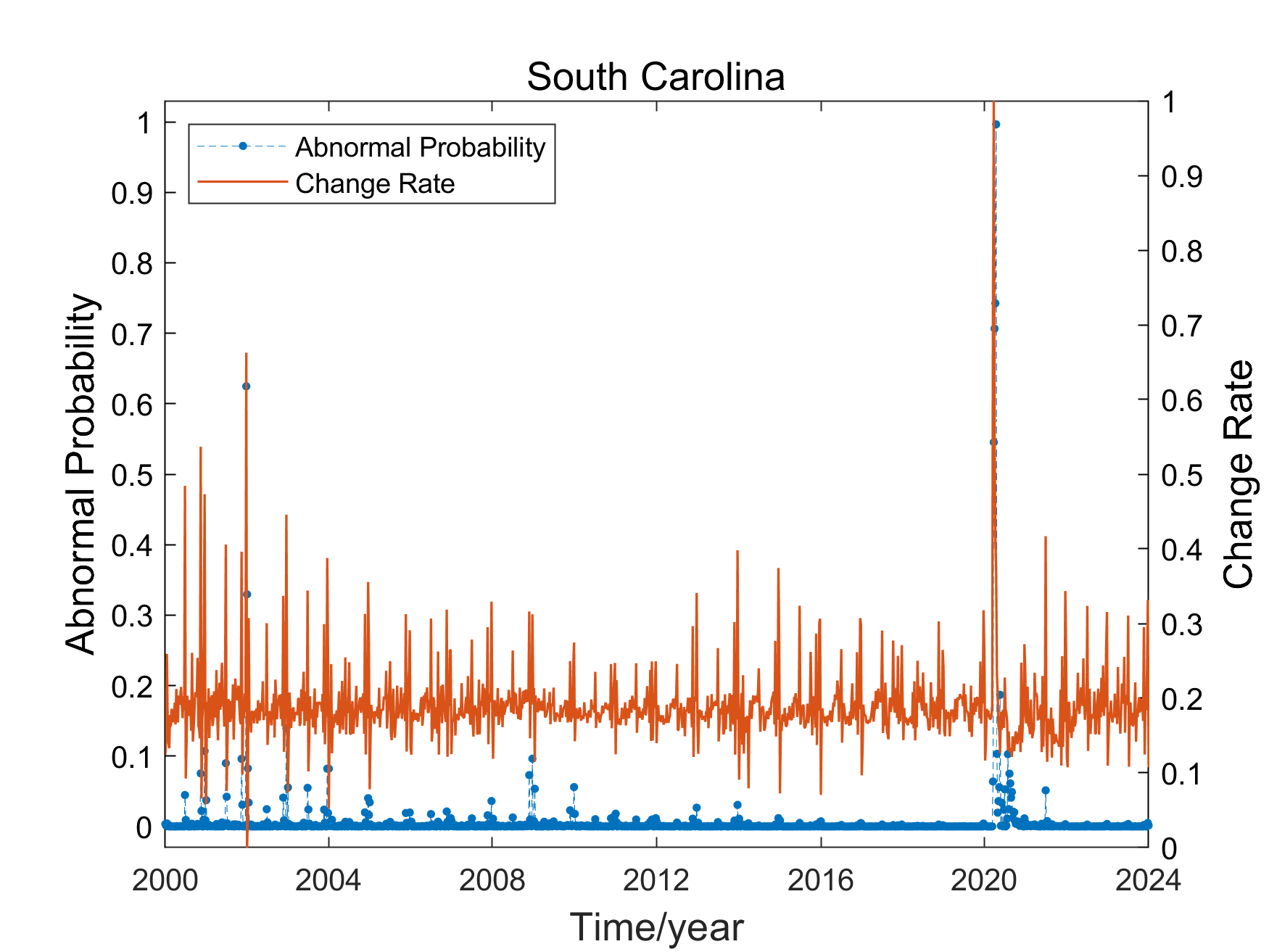}
			\end{minipage}
		}\\
		\subfigure{
			\begin{minipage}{0.15\textwidth}
				\centering
				\includegraphics[scale=0.16]{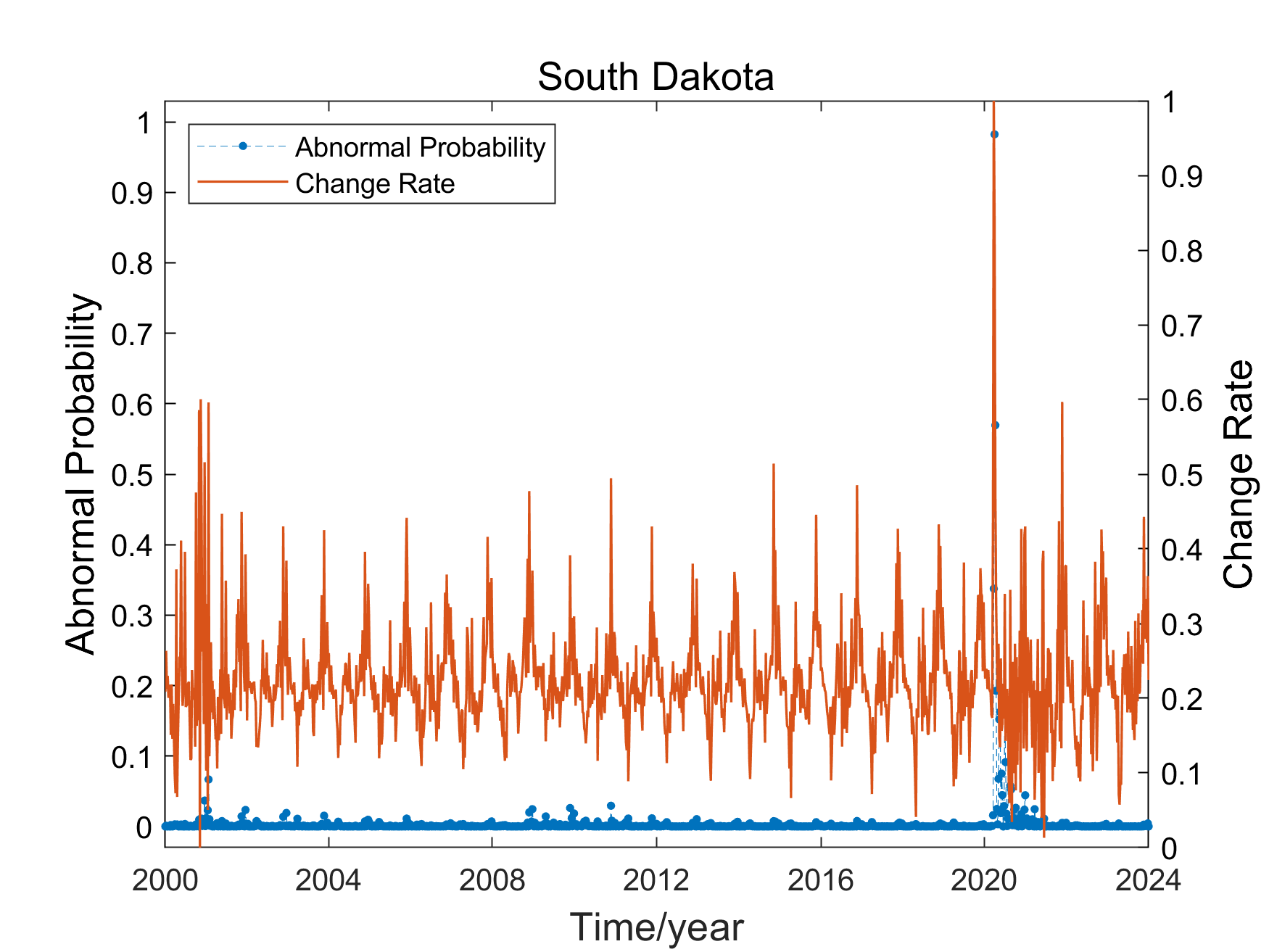}
			\end{minipage}
			\begin{minipage}{0.15\textwidth}
				\centering
				\includegraphics[scale=0.16]{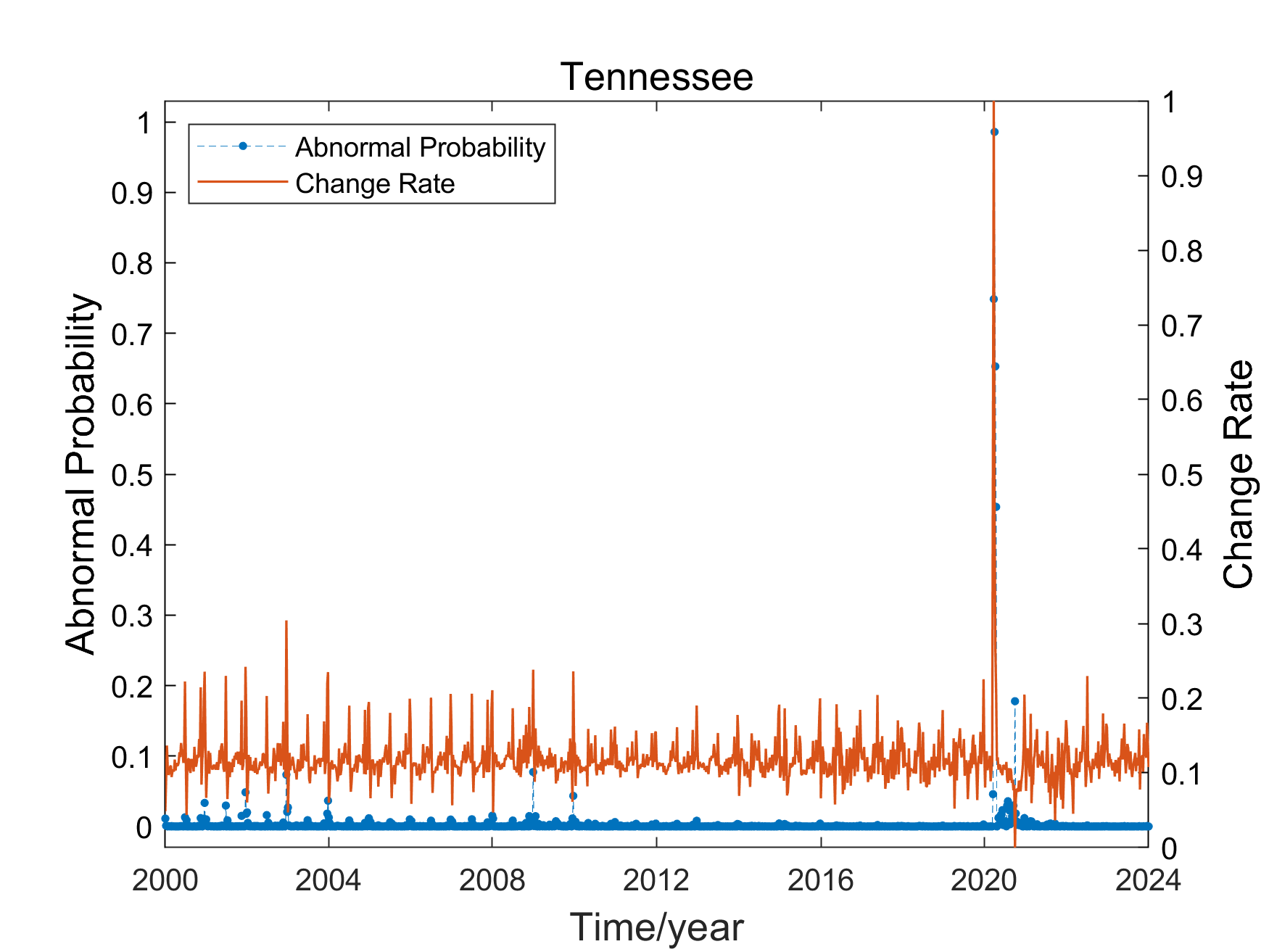}
			\end{minipage}
			\begin{minipage}{0.15\textwidth}
				\centering
				\includegraphics[scale=0.16]{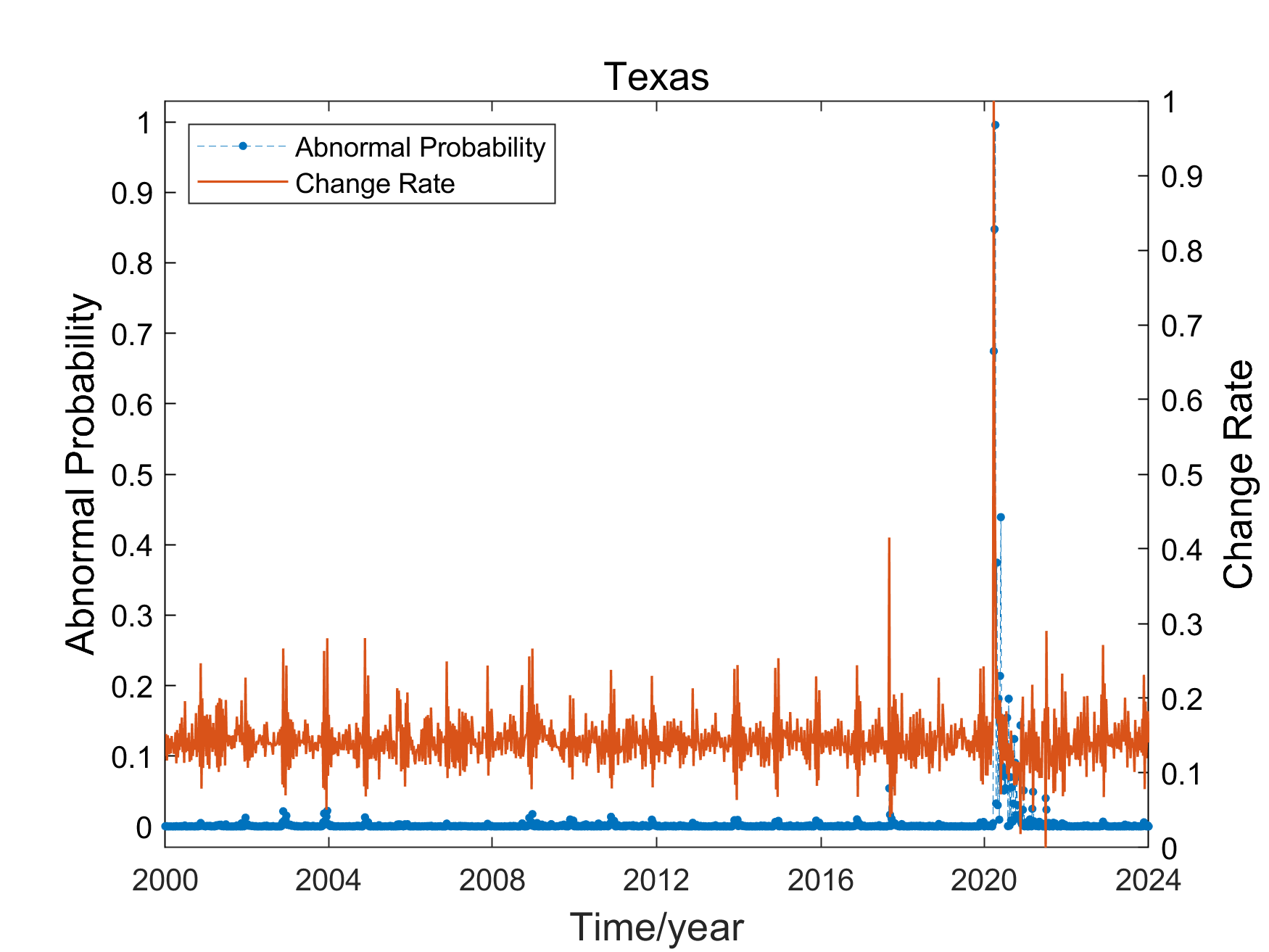}
			\end{minipage}
			\begin{minipage}{0.15\textwidth}
				\centering
				\includegraphics[scale=0.16]{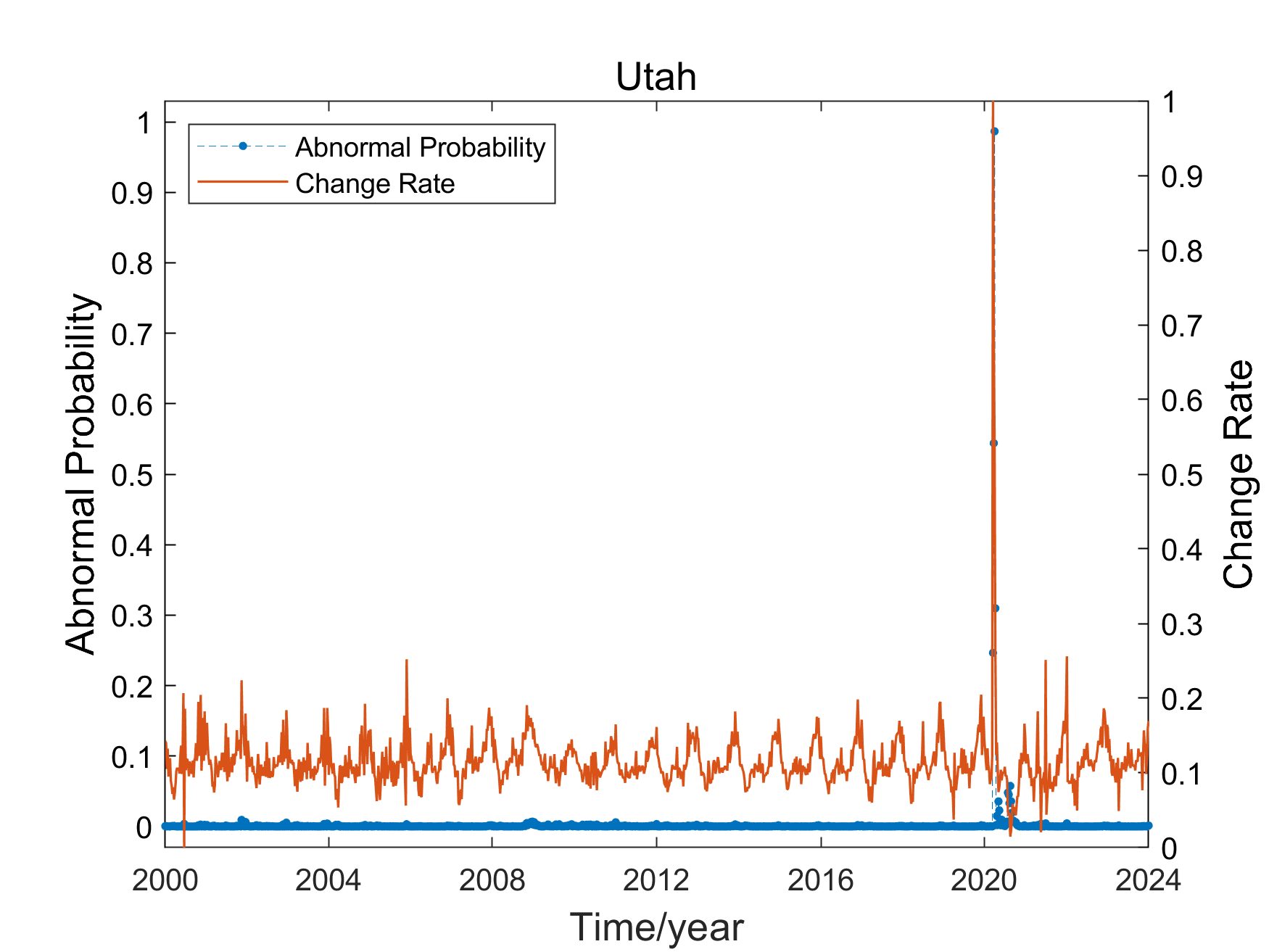}
			\end{minipage}
			\begin{minipage}{0.15\textwidth}
				\centering
				\includegraphics[scale=0.16]{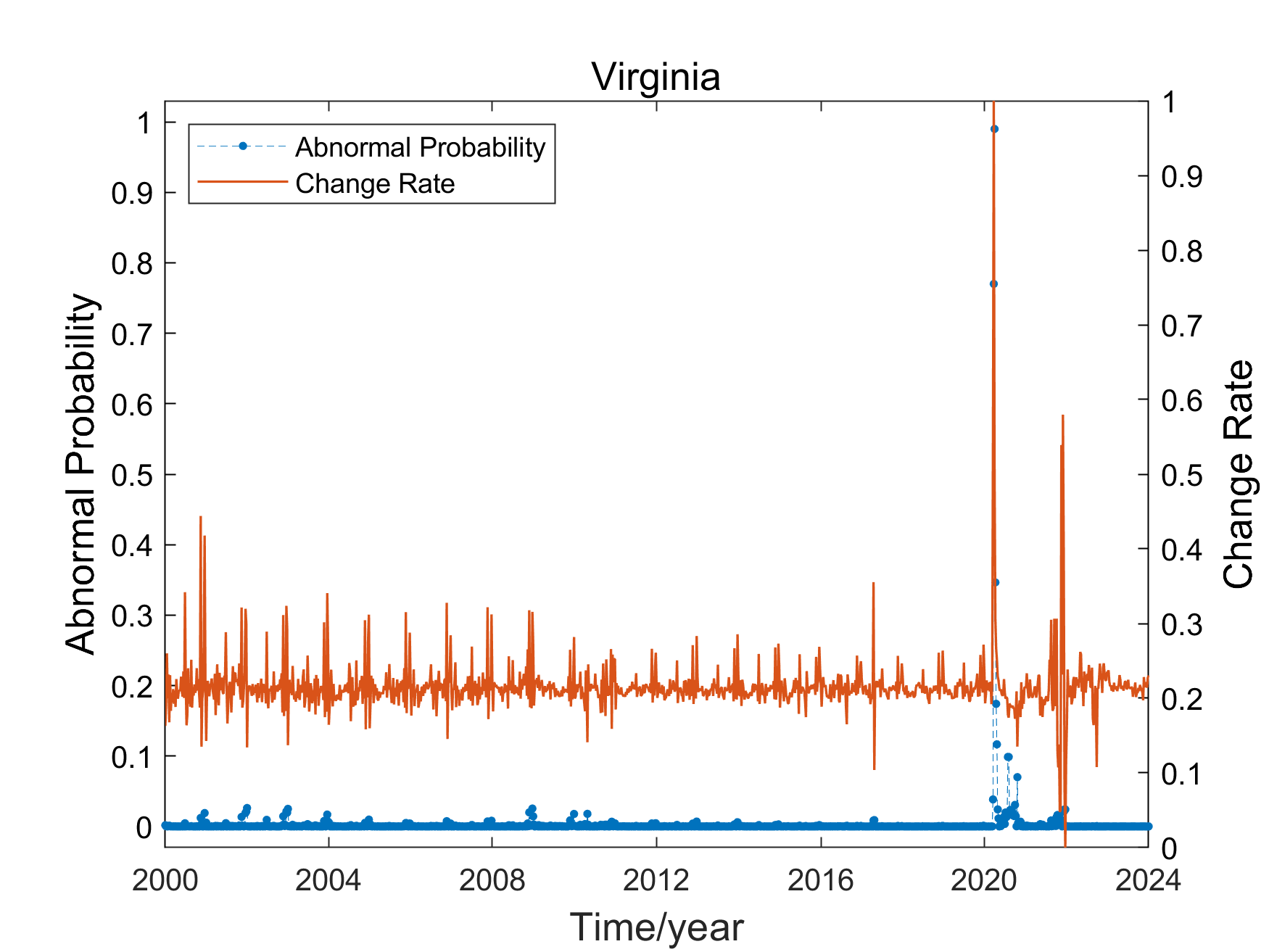}
			\end{minipage}
			
		}\\
		\subfigure{
			\begin{minipage}{0.15\textwidth}
				\centering
				\includegraphics[scale=0.16]{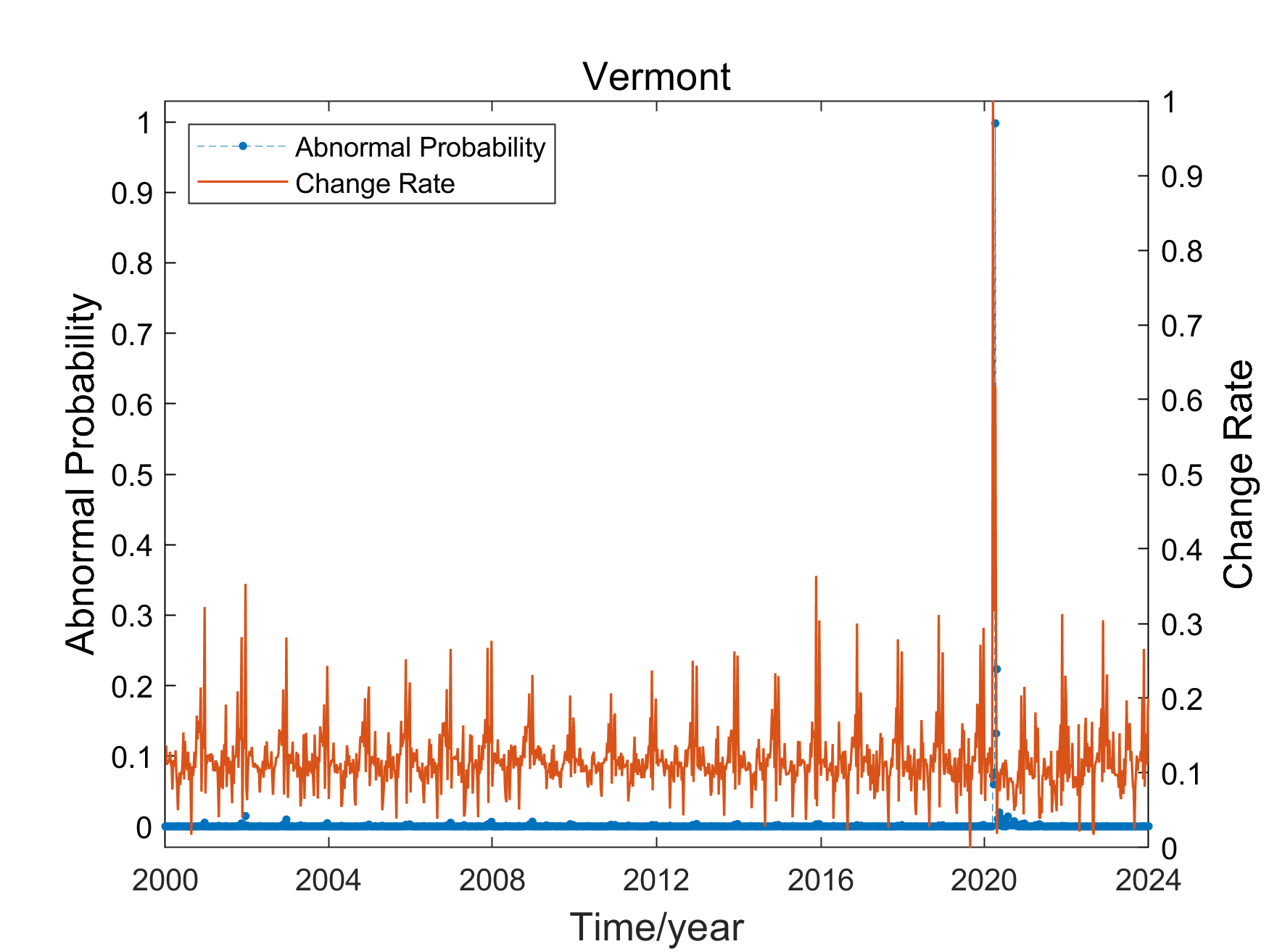}
			\end{minipage}
			\begin{minipage}{0.15\textwidth}
				\centering
				\includegraphics[scale=0.16]{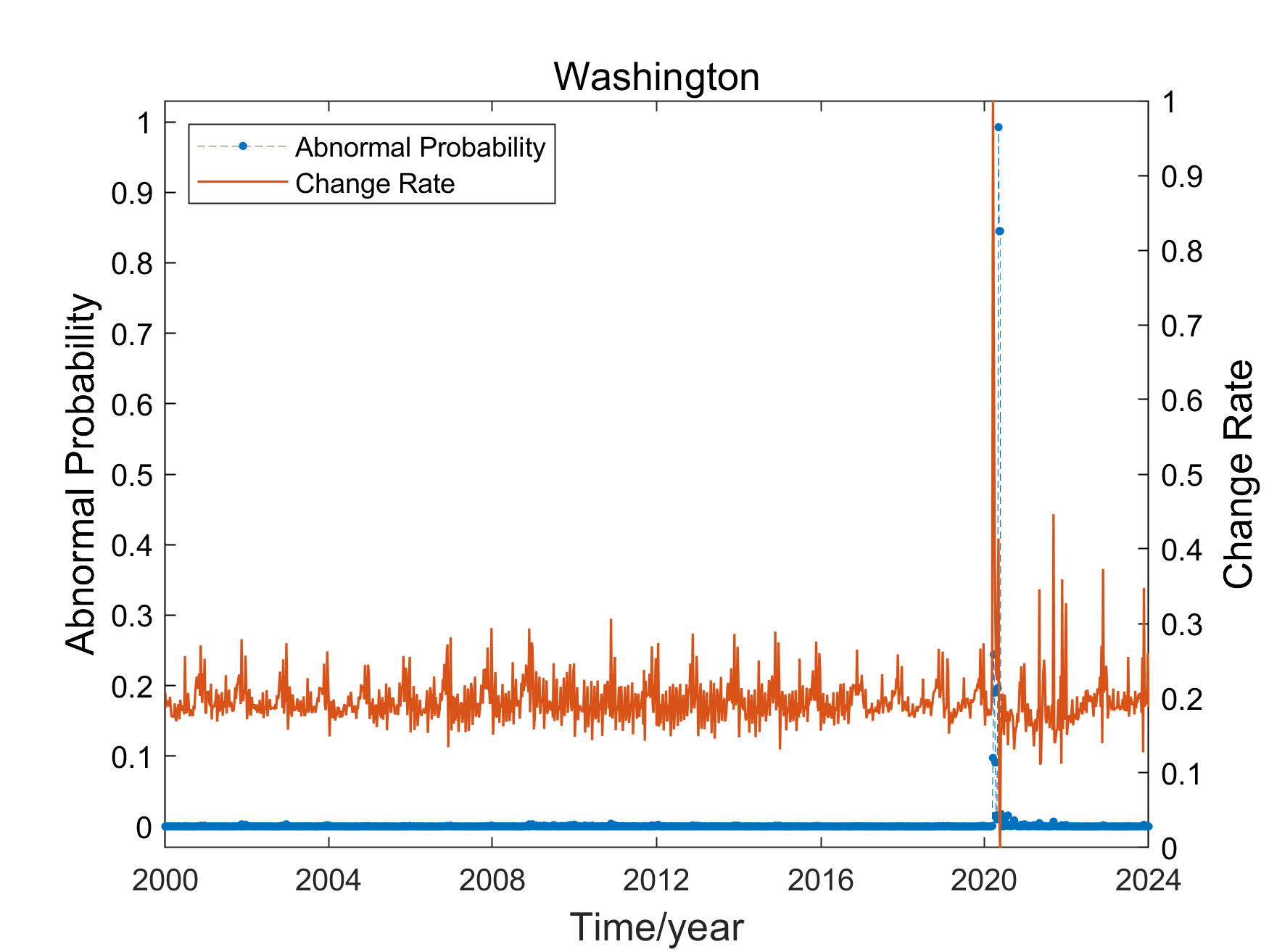}
			\end{minipage}
			\begin{minipage}{0.15\textwidth}
				\centering
				\includegraphics[scale=0.16]{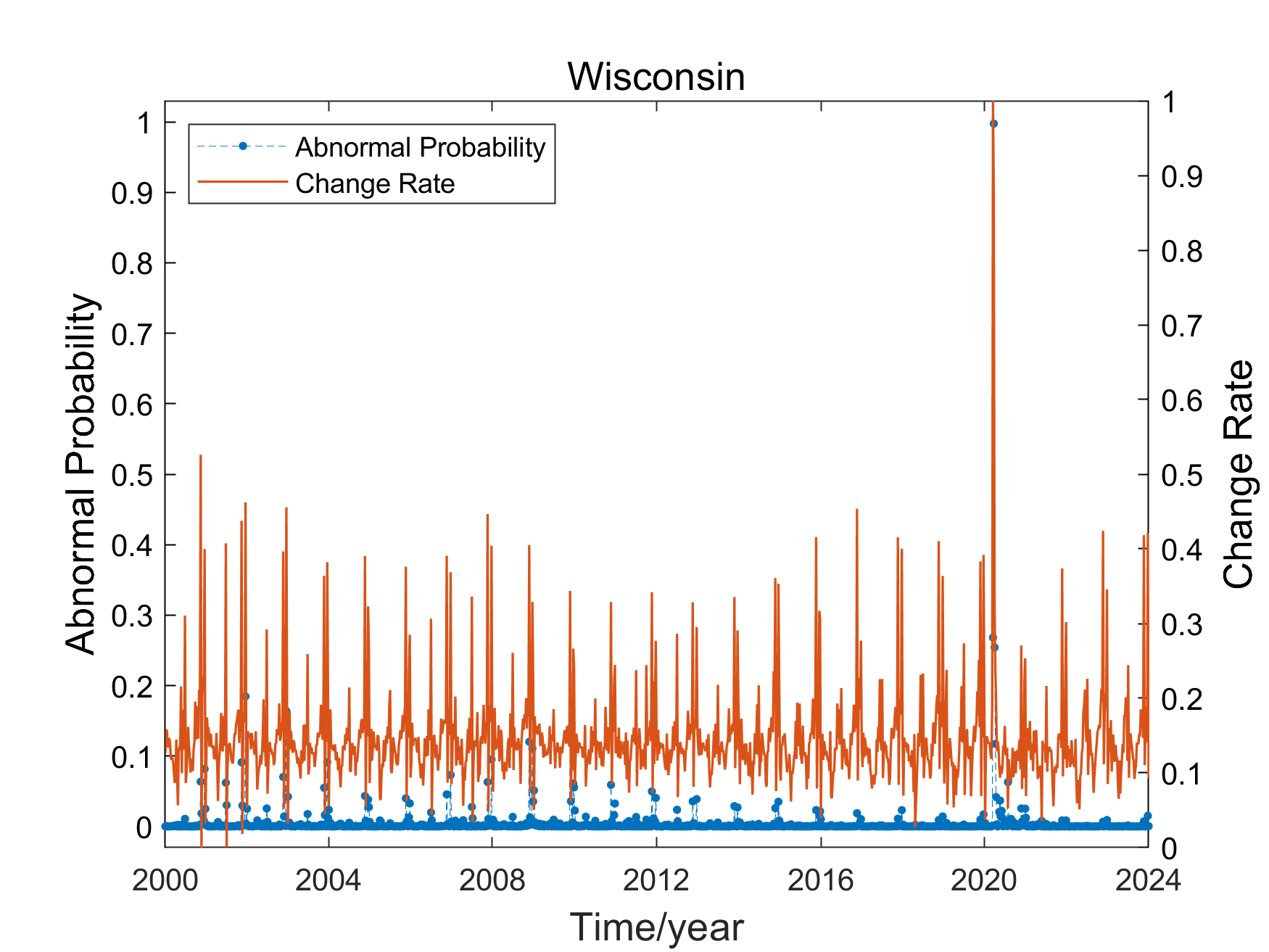}
			\end{minipage}
			\begin{minipage}{0.15\textwidth}
				\centering
				\includegraphics[scale=0.16]{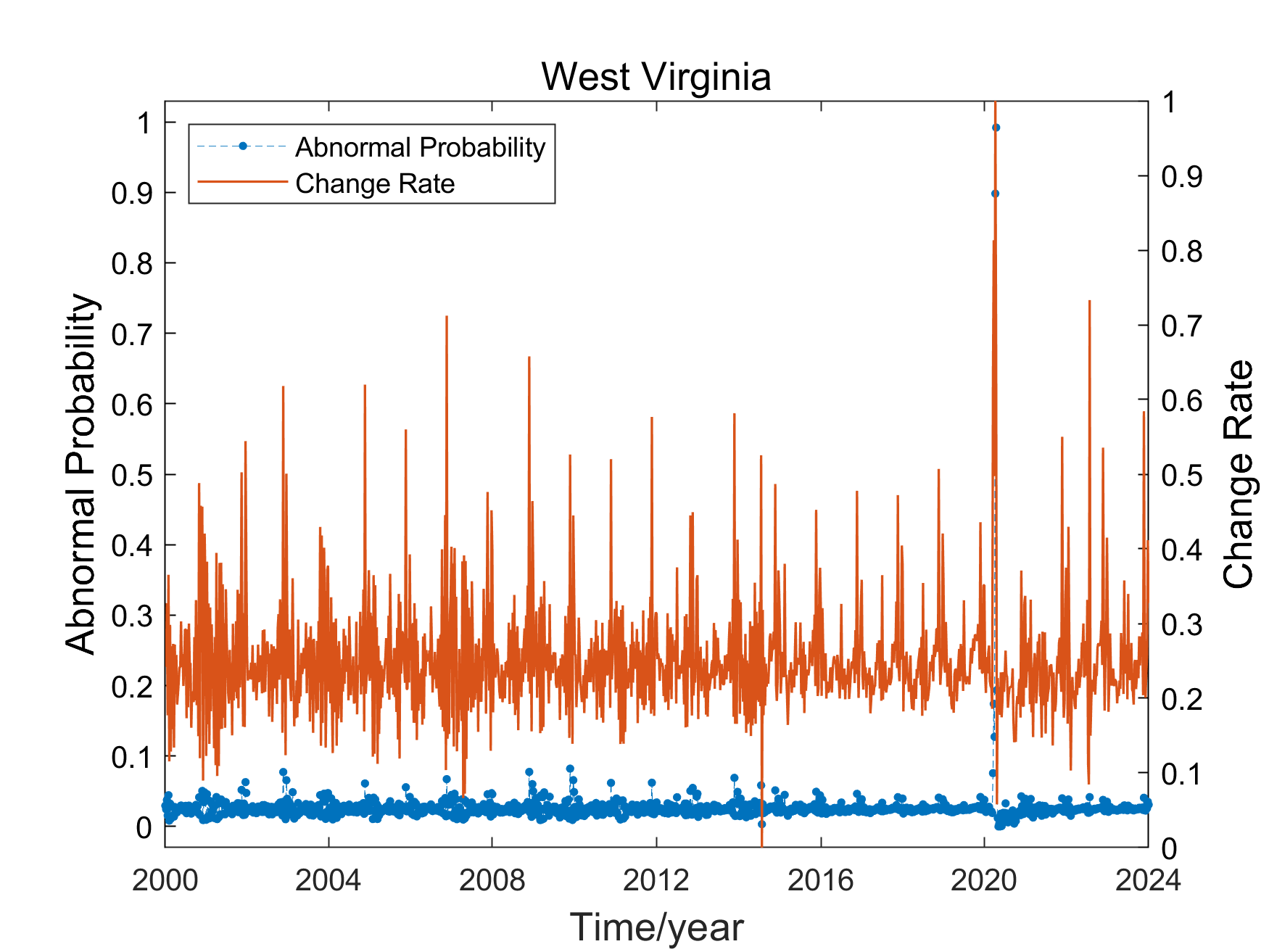}
			\end{minipage}
			\begin{minipage}{0.15\textwidth}
				\centering
				\includegraphics[scale=0.16]{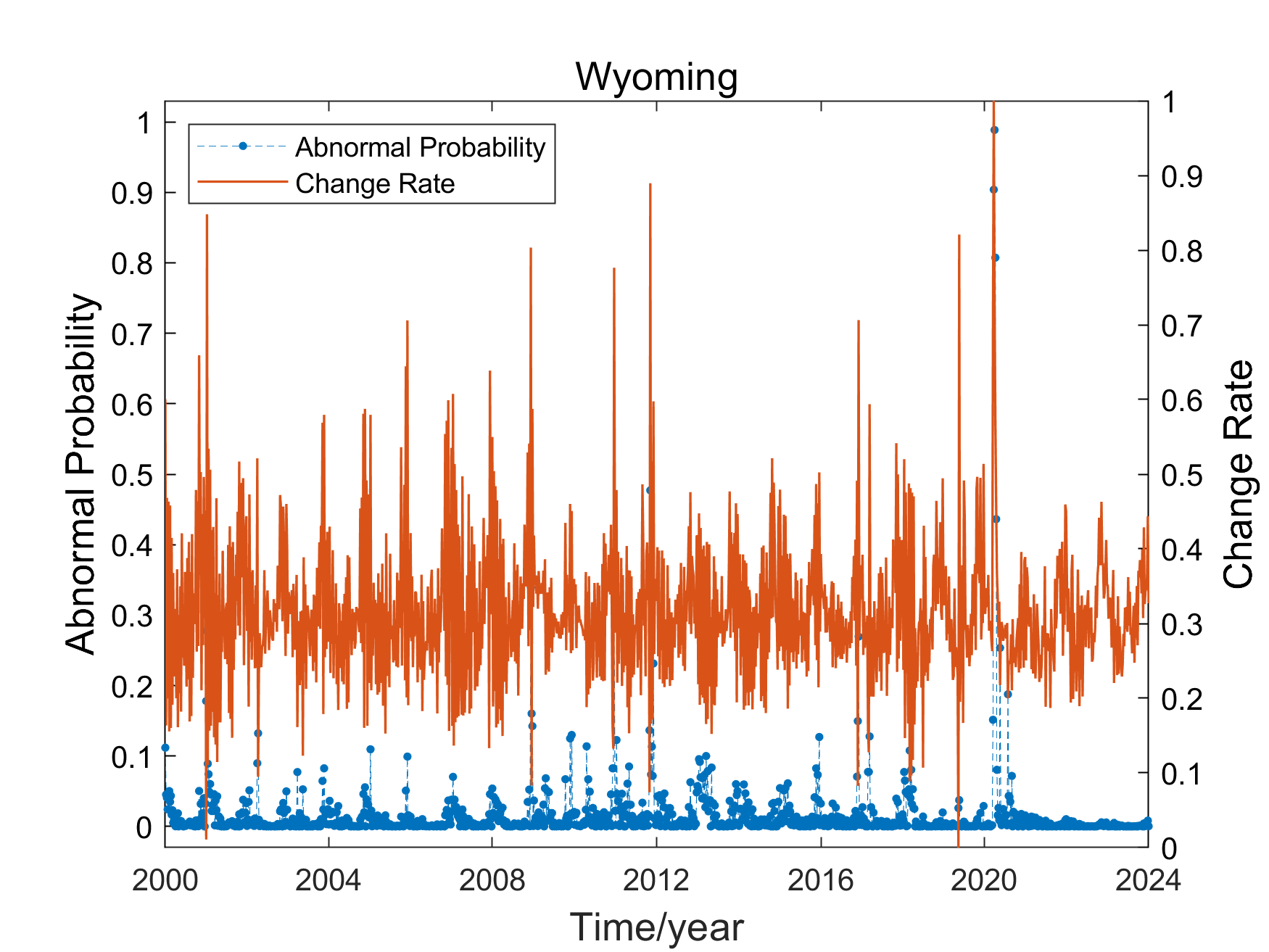}
			\end{minipage}
		}
		\caption{\label{us3}EMODM Results for Insured Unemployment Data in other regions.}
	\end{figure}

	After obtaining the statistical inference results for the overall data in the whole U.S. data, we applied the EMODM to the insured unemployment data for each of the 53 regions. For example, California is located on the West Coast of the United States. In contrast, New York is situated on the East Coast. They both have significant economic influence and large, diverse populations. In figure \ref{us2}, the left and middle figures, representing California and New York respectively, we observed that the system also entered an abnormal mode around March 21, 2020, due to the COVID-19 pandemic, consistent with the results for the total data in figure \ref{us1}. However, in the right figure, which represents the U.S. Virgin Islands, we get different results. The U.S. Virgin Islands, due to its smaller population, lower economic volume, and status as an overseas territory of the United States, shows data that is more susceptible to noise and delayed economic impacts. Consequently, the detection results for this region are less accurate. Since the EMODM is based on probabilistic models and statistical algorithms, incorporating larger sample sizes tends to yield more accurate results. The EMODM results for the other 50 regions are presented in figure \ref{us3}.  In summary, while the EMODM effectively detected the COVID-19 induced anomalies in the regions which have large economies and populations, its accuracy diminishes in smaller, more isolated regions due to the small sample data which is susceptible to noise and delayed effects on their economic markets. These results and figures collectively highlight the effectiveness of the EMODM in identifying and analyzing abnormal patterns in complex systems, particularly in the context of significant economic shocks such as the COVID-19 pandemic.

	\section{Comparison with Classical Methods}\label{Comparison}
	In this study, we proposed the EMODM, a novel fast online method for detecting abnormal patterns in the outputs of complex systems. Utilizing probabilistic models and statistical algorithms, the EMODM is based on a two-state Gaussian mixture model, enabling it to perform probability anomaly detection on real-time raw data without relying on special prior distribution information. We have chosen the numerical solution shown in figure \ref{result1} in section \ref{Third_sec} about the Sallen-Key low-pass filter model with single-component voltage input as the data used for the testing of each method. The result of comparing EMODM with other classical methods is presented in table \ref{tab.com}.
	
	Comparatively, various outlier detection algorithms each have unique strengths and weaknesses. The linear regression model(LRM) is simple and effective for trend analysis and continuous data, but it assumes a predefined model structure, making them less suitable for complex distributions\cite{wisnowski2001comparative}. The kernel Density Estimation(KDE) is non-parametric and does not assume a specific distribution, but it can be computationally intensive, especially in high dimensions\cite{latecki2007outlier}. The k-nearest neighbors(KNN) algorithm is intuitive and does not require assumptions about data distribution, yet its high computational complexity and sensitivity to the choice of the component number can be limitations\cite{chen2010neighborhood}. The local outlier factor(LOF) captures local data density variations effectively, making it suitable for data with distinct density variations, though its performance degrades with high-dimensional data\cite{breunig2000lof}. The K-means clustering is simple and effective for identifying cluster-based anomalies but requires the number of clusters to be predefined, and is less effective for non-spherical clusters\cite{chawla2013k}. The isolation forest(IF) is efficient for high-dimensional data and identifies anomalies based on fewer required cuts to isolate them\cite{liu2008isolation}. However, it assumes anomalies are few and significantly different from normal data. The recurrent neural networks(RNN) and the generative adversarial networks(GAN) offer powerful capabilities for sequential and high-dimensional data, respectively\cite{davari2021real, oh2019oversampling}. The RNN is particularly suited for time-series data with temporal dependencies but requires substantial training data and computational resources. The GAN is powerful for modeling complex, high-dimensional data distributions and generating synthetic data, but they both require extensive computational resources and hyperparameter tuning, along with complex model training processes.
	
	Each of these outlier detection algorithms is suited for different types of data and specific scenarios. The choice of algorithm should be guided by the specific characteristics and requirements of the data and application at hand. The EMODM, with its effectiveness in real-time anomaly detection in complex systems with Gaussian noise, proves to be a robust tool in the suite of outlier detection methodologies, particularly for large-scale and noisy datasets. Future work could explore further optimization of the EMODM for smaller datasets and its application to a broader range of complex systems, potentially integrating it with other advanced machine learning techniques to enhance its detection capabilities and robustness.    
	
	\begin{table}[H]
		\centering
		\begin{tabular}{c|ccc}
			\hline
			\hline
			Method & Anomalies Detected & Abnormal Probability & Computation Time/$s$  \\
			\hline
			Real Setup & 30 & 4.76\,\% & $\backslash$ \\
			$\bf{EMODM}$ & $\mathbf{22}$ & $\mathbf{4.66\,\%}$ & $\mathbf{12}$ \\ 
			LRM & 14 & 3.89\,\% & 7 \\ 
			KDE & 16 & 4.12\,\% & 19 \\ 
			KNN & 21 & 4.51\,\% & 27 \\ 
			K-means & 20 & 4.46\,\% & 15 \\ 
			IF & 12 & 3.75\,\% & 6 \\ 
			RNN & 24 & 4.71\,\% & 52 \\ 
			GAN & 26 & 4.80\,\% & 136 \\ 
			\hline
			\hline
		\end{tabular}
		\caption{\label{tab.com} Comparing EMODM with other classical methods}
	\end{table}

	\section{Conclusion and Discussion}\label{fifth_sec}
	In this study, we proposed the Exception Maximization Outlier Detection Method(EMODM), a novel fast online methodology for detecting abnormal patterns in the outputs of complex systems. Utilizing probabilistic models and statistical algorithms, EMODM is based on a two-state Gaussian mixture model, enabling it to perform probability anomaly detection on real-time raw data without relying on special prior distribution information. The efficacy of EMODM was confirmed through synthetic data from two numerical cases as Sallen-Key low-pass filter model and the LLG equation. In two real-world applications, the EMODM successfully detected short circuit patterns in a three-phase inverter system by analyzing current and voltage outputs. The EMODM identified the abnormal period during the COVID-19 pandemic in the insured unemployment data across 53 regions in the United States from 2000 to 2024. These applications demonstrated the method's effectiveness and accuracy in both synthetic and real-life datasets.
	
	Our results highlight that EMODM is capable of providing reliable outlier detection in complex systems affected by noise. This capability is crucial for maintaining system stability and reliability, especially in scenarios where traditional methods may fall short due to the need for prior knowledge or linear assumptions. Future work could explore further optimization of the EMODM for smaller datasets and its application to a broader range of complex systems. Additionally, integrating EMODM with other advanced machine learning techniques may enhance its detection capabilities and robustness, paving the way for more comprehensive anomaly detection solutions in various fields.
	
	\section*{Acknowledgements}
	This work was supported by the NSFC grant 12371198 and the fundamental research funds for the central universities under grand YCJJ20242224.
	
	\bibliographystyle{elsarticle-num}
	\bibliography{main}

\end{document}